\documentclass{sn-jnl}

\usepackage{graphicx}%
\usepackage{multirow}%
\usepackage{amsmath,amssymb,amsfonts}%
\usepackage{amsthm}%
\usepackage{mathrsfs}%
\usepackage[title]{appendix}%
\usepackage{xcolor}%
\usepackage{textcomp}%
\usepackage{manyfoot}%
\usepackage{booktabs}%
\usepackage{algorithm}%
\usepackage{algorithmicx}%
\usepackage{algpseudocode}%
\usepackage{listings}%

\theoremstyle{thmstyleone}%
\theoremstyle{thmstyletwo}%

\theoremstyle{thmstylethree}%

\usepackage[utf8]{inputenc}
\usepackage[T1]{fontenc}
\usepackage{lmodern}

\usepackage{amsmath, amssymb, mathrsfs}

\usepackage{graphicx}

\usepackage{subfig}

\usepackage{placeins}  

\usepackage{array, booktabs, multirow, makecell, tabularx}
\usepackage{siunitx}
\usepackage{rotating}
\usepackage{pdflscape}

\usepackage{algorithm}
\usepackage{algpseudocode}

\usepackage{pgfplots}
\pgfplotsset{compat=1.17}
\usepackage{tikz}
\usetikzlibrary{patterns, positioning}

\usepackage[table]{xcolor}
\definecolor{lightgray}{gray}{0.93}

\usepackage{hyperref}
\hypersetup{
	colorlinks=true,
	linkcolor=blue,
	citecolor=blue,
	urlcolor=blue,
	pdftitle={DADIR: Density-Aware Data-level Imbalanced Regression Framework},
	pdfauthor={Shermin Shahbazi, Hossein Mohammadi, Mohsen Afsharchi}
}

\begin{document}

	\title[DADIR: Density-Aware Data-level Imbalanced Regression Framework]{DADIR: Density-Aware Data-level Imbalanced Regression Framework}
	
	
	\author*[1]{\fnm{Shermin} \sur{Shahbazi}}\email{sh.shahbazi@znu.ac.ir}
	
	\author*[1]{\fnm{Hossein} \sur{Mohammadi}}\email{hosm@znu.ac.ir}
	
	\author[1]{\fnm{Mohsen} \sur{Afsharchi}}\email{afsharchi@znu.ac.ir}
	
	\affil[1]{\orgdiv{Department of Computer Engineering}, \orgname{University of Zanjan},
			\state{Zanjan}, 
			\country{Iran}}

	
	\abstract{Imbalanced learning addresses predictive modeling problems with underrepresented regions of the data distribution. Although extensively studied in classification, imbalanced regression remains challenging due to continuous target variables and heterogeneous density patterns. Existing data-level approaches often rely on fixed partitioning or synthetic sample generation without jointly considering density variations and local feature-space structure. To address these limitations, we propose \emph{DADIR}, a \emph{Density-Aware Data-level Imbalanced Regression} framework that leverages density information throughout the entire balancing process. DADIR consists of three complementary phases. First, \emph{Density-Aware Adaptive Partitioning (DAAP)} recursively partitions the target space according to intrinsic density variations, enabling more reliable identification of minority and majority regions, than fixed-bin strategies. Second, \emph{Density-Regularized Conditional Variational Autoencoder (DR-CVAE)} learns latent representations that preserve sparse regions while maintaining global data structure. Third, \textit{data-level balancing via latent-space modeling} integrates feature-level clustering with latent-space oversampling to generate synthetic samples that respect local structural patterns and reduce sampling bias within underrepresented regions. Beyond their integration within a unified framework, the proposed DAAP, DR-CVAE, and latent-space balancing modules each represent independent methodological contributions that address complementary challenges in identifying minority regions, preserving sparse-region representations, and generating structurally consistent synthetic samples. The resulting balanced dataset can be directly employed by downstream regressors without modifying their architecture or learning objective. Experiments on diverse imbalanced regression datasets show that DADIR consistently improves predictive performance, particularly in underrepresented regions, while enhancing overall accuracy.}
	

	\keywords{imbalanced learning in regression, data-level imbalanced learning, density-aware imbalanced Learning, adaptive bin partitioning}
	
	
	
	\maketitle

	\section{Introduction}
	\label{sec:Introduction}
	\emph{Imbalanced learning} concerns predictive modeling scenarios in which some target values are represented by substantially fewer observations than others \cite{Chen2024, REZVANI2023}. Although this issue has been widely investigated in classification, where imbalance typically appears as an unequal distribution of class labels, its interpretation is more complex in regression because the target variable is continuous. In regression settings, imbalance is usually reflected by uneven sample density across the target space, such that particular intervals, especially rare or extreme target ranges, contain only a limited number of observations \cite{BELHAOUARI2024, Ribeiro2020}. As a result, important regions of the target space may be poorly covered in the training set, even when they correspond to critical phenomena in real-world applications \cite{DOLAR2025tuba, ZHU2019140}.
	
	The uneven nature of such distributions poses specific challenges for regression algorithms. Conventional regression models, which are typically optimized by minimizing global prediction error, tend to favor the densely populated regions of the target space. Consequently, their predictive accuracy often deteriorates in sparsely represented, low-density \emph{minority regions} \cite{Kaur2019Har}. This limitation is particularly critical in applications where rare samples carry substantial practical importance \cite{Ribeiro2020}, such as extreme weather forecasting \cite{ZHEN2025120952, Scheepens2023}, environmental and earth sciences \cite{LI2025118265, TANG2025126386}, rare-event prediction in medicine \cite{Miao2017ieee, YEW2025min}, and anomaly detection \cite{2017Candelieri, Hajjami2020},  and financial forecasting \cite{2020Lucas, LI2025Financial}. In such domains, accurately predicting rare or extreme outcomes is often more important than achieving uniformly low average error across the entire target range. Therefore, the limited generalization ability of standard models in these critical regions highlights the need for specialized regression methods capable of handling imbalance in continuous targets.
	
	To address the challenges associated with imbalanced regression, imbalanced learning has evolved as an important research direction since early studies on imbalanced learning in the late 1990s \cite{Kubat1997}. Initial studies mainly focused on classification problems, where imbalance appears as skewed class distributions \cite{Chen2024, Goswami2024, REZVANI2023}; however, increasing attention has recently been directed toward regression tasks, in which imbalance occurs over a continuous target space and therefore requires different methodological treatment. Existing approaches are commonly grouped into two main families: \emph{data-level methods}, which modify the training distribution through resampling, synthetic sample generation, or dataset balancing, and \emph{algorithm-level} or \emph{cost-sensitive methods}, which adjust the learning objective, typically through cost-sensitive losses or sample-weighting mechanisms \cite{Aguiar2024, Chen2024, Kaur2019Har}. Data-level methods are especially attractive because they can improve the representation of sparse target regions while remaining largely independent of the downstream regressor. Prominent data-level methods include SMOGN \cite{SMOGN2017}, which generates synthetic samples through interpolation, WERCS \cite{Ribeiro2020}, which employs weighted relevance-based sampling, and SmoteR \cite{SmoteR2013}, which adapts the SMOTE algorithm for continuous targets. However, many existing data-level approaches do not explicitly integrate the density structure of the continuous target space, the local organization of the feature space, and structured representation learning within a unified balancing pipeline. Consequently, synthetic samples may be generated in target-sparse regions without sufficiently accounting for feature-space heterogeneity, increasing the risk of producing samples in incompatible or poorly supported regions. To address this limitation, this study proposes \emph{DADIR}, a \emph{Density-Aware Data-level Imbalanced Regression framework} composed of three complementary phases: 
		
	\begin{itemize}
		\item \emph{Phase 1. Density-Aware Adaptive Partitioning (DAAP):} to form locally coherent intervals in the continuous target space, DAAP employs a recursive partitioning scheme driven by density valleys estimated through Kernel Density Estimation (KDE). In this context, density-consistent bins refer to intervals separated by pronounced low-density regions, so that high-density target modes are preserved rather than arbitrarily divided. Unlike fixed-width, quantile-based, or relevance-threshold binning, DAAP derives its boundaries from the intrinsic density structure of the target distribution. Consequently, the resulting partitions better preserve natural target concentrations and isolate sparse transition zones, providing a more meaningful basis for localized balancing in subsequent phases.
				
		\item \emph{Phase 2. Density-Regularized Conditional Variational Autoencoder (DR-CVAE):} DR-CVAE adopts a conditional VAE formulation to learn a smooth, compact, and target-aware latent representation suitable for synthetic sample generation. The latent space is conditioned on the continuous target value, allowing feature dependencies to be organized according to target-related variations. In addition, density regularization derived from the marginal target density estimated in Phase 1 is imposed through sample-wise reconstruction weights, thereby emphasizing observations from low-density regions. This mitigates the dominance of majority regions during representation learning and improves latent-space quality where reliable modeling is most critical.

		\item \emph{Phase 3. Data-level balancing via latent-space modeling:} To perform the final balancing step, DADIR augments underrepresented regions in the learned latent space through feature-level clustering and density-guided sample generation. This phase is designed to balance underrepresented target regions while preserving local feature-space structures, thereby mitigating the risk of generating synthetic samples across incompatible regions.

	\end{itemize}
	
	The main contribution of this study is the development of three complementary components, DAAP, DR-CVAE, and latent-space data balancing, which are integrated within the DADIR framework to address imbalanced learning in regression from a data-level perspective. The resulting data-level framework produces a balanced training dataset that can be directly used by downstream regressors without requiring changes to the model architecture or learning objective. Accordingly, this study sets out to investigate the following research questions:

	\begin{itemize}
		\item \textbf{RQ1:} How does the proposed data-level imbalanced learning method perform in imbalanced regression tasks compared to (i) standard regression baselines without imbalance handling and (ii) state-of-the-art imbalanced regression methods, both in terms of overall predictive performance and performance within minority target regions?
		
		\item \textbf{RQ2:} How do the degree and characteristics of target variable skewness across datasets affect the effectiveness and robustness of the proposed data-level imbalanced learning method?

		\item \textbf{RQ3:} What are the limitations of the proposed data-level imbalanced learning method?
		
		\item \textbf{RQ4:} How does each phase of the proposed framework (Density-Aware Adaptive Partitioning (DAAP), Density-Regularized Conditional Variational Autoencoder (DR-CVAE), Data-Level Balancing via Latent-Space Modeling) contribute to the overall performance in imbalanced regression tasks?

	\end{itemize}

	The remainder of this paper is organized as follows. \hyperref[sec:literature-review]{Literature Review} investigates the related literature, with particular emphasis on data-level imbalanced learning for regression. \hyperref[sec:methods]{Methods} presents the proposed three-phase DADIR architecture and explains its main methodological components. \hyperref[sec:resultAndDiscussion]{Results and Discussion} first describes the evaluation metrics, hyperparameter settings, baseline models, and experimental protocol, and then reports the empirical results, answers the research questions, and provides a comparative analysis against the selected baselines. Finally, \hyperref[sec:conclusion]{Conclusion} concludes the paper and outlines possible directions for future research.

	\section{Literature Review}
	\label{sec:literature-review}
	Imbalanced learning has been extensively investigated in classification problems, where discrete class labels enable a clear distinction between majority and minority classes. This has led to the development of a broad spectrum of solutions, including resampling techniques, cost-sensitive learning, ensemble methods, and hybrid strategies \cite{REZVANI2023, Chen2024, Goswami2024, ArafIm2024}. In contrast, imbalanced regression, has received considerably less attention despite its importance in numerous real-world applications involving rare but critical target values.
	
	A comprehensive review of the literature reveals that existing imbalanced regression approaches can generally be categorized into two main groups \cite{Chen2024, Aguiar2024, Kaur2019Har}: the first group, \emph{data-level methods}, aims to alleviate imbalance by modifying the training data distribution through \emph{oversampling}, or \emph{undersampling}, thereby directly altering the composition of the input dataset prior to model training. The second group, \emph{algorithm-level methods}, addresses imbalance by adapting the learning process itself, rather than directly altering the input dataset, typically through specialized loss functions, sample-weighting mechanisms, architectural modifications, or optimization strategies that place greater emphasis on rare target values \cite{ArafIm2024, fernandez2018learning}. While such approaches can improve the model's sensitivity to underrepresented target regions, they do not directly modify the underlying data distribution and may therefore remain constrained by the limited availability of informative samples in sparse regions \cite{Aguiar2024, SHAHBAZI2026}.
	
	In addition to these two main groups, some studies have combined data-level and algorithm-level methods to form \emph{hybrid approaches}, in which both types of strategies are integrated within a unified framework \cite{SHAHBAZI2026,Chen2024,fernandez2018learning,KHAN2024122778,Dablain2024}. By combining improvements in data representation with modifications to the learning process, hybrid methods seek to exploit the complementary strengths of both paradigms and have demonstrated promising performance in several imbalanced regression studies. Nevertheless, their increased methodological complexity, computational overhead, and dependence on multiple interacting components may limit their practicality and generalizability across different application domains.
	
	Given that the proposed DADIR framework belongs to the data-level family, the remainder of this section focuses on data-level imbalanced regression approaches and their associated challenges.

	\subsection{Data-Level Imbalanced Learning}
	\label{subsec:LitRew_Data_level}
	Data-level methods, commonly referred to as \emph{preprocessing techniques}, address imbalance by modifying the training data distribution prior to model construction. Unlike algorithm-level approaches, which alter the learning objective or optimization procedure, data-level methods operate directly on the available observations and can therefore be applied independently of the underlying regression algorithm \cite{Aguiar2024, Chen2024}. This \emph{model-agnostic} nature has made them one of the most widely adopted strategies for imbalanced learning, as they can be readily integrated into existing machine learning pipelines without requiring modifications to the predictive model itself.
	
	The importance of data-level balancing is particularly evident in imbalanced regression tasks, where prediction accuracy often deteriorates in sparsely populated target regions due to the limited availability of representative training samples \cite{BELHAOUARI2024, SHAHBAZI2026, gao2026}. This continuity of the target space makes rare-region identification, importance estimation, and representative sample generation inherently more difficult than in discrete-label settings \cite{gao2026, BELHAOUARI2024}. Consequently, identifying rare regions, quantifying their importance, and generating representative samples become considerably more challenging \cite{gao2026, BELHAOUARI2024}. Furthermore, the gradual transition between dense and sparse target regions (and the consequent variation in local target density) increases the risk of producing synthetic observations that fail to preserve the underlying data characteristics \cite{Steininger2021, Shahbazi2025}.
	
	To address the scarcity of representative observations in rare target regions, data-level methods seek to balance the training distribution while preserving the underlying characteristics of the original dataset. Accordingly, data-level balancing has become a central paradigm in imbalanced regression, although its effectiveness depends on enriching minority regions without distorting the original feature--target distribution \cite{Avelino2024, GONG2023Yoidu, Carvalho2025}. Nevertheless, achieving an appropriate balance between minority-region enrichment and preservation of the original data distribution remains a central challenge.
	
	Existing data-level balancing strategies can generally be categorized into two major groups: \emph{oversampling} and \emph{undersampling}. Oversampling methods increase the representation of minority regions through sample generation, interpolation, or duplication, whereas undersampling methods alleviate imbalance by selectively reducing the influence of majority regions. The following subsections review these two categories and their representative developments in the imbalanced regression literature.

	\subsubsection{Oversampling}
	\label{subsec:LitRew_Oversampling}
	\emph{Oversampling} constitutes the most extensively studied family of data-level balancing techniques for imbalanced regression \cite{BELHAOUARI2024, Carvalho2025, DOUZAS20181}. Rather than removing majority-region observations, oversampling enriches underrepresented regions through duplication, interpolation, perturbation, or generative sample synthesis, making it particularly attractive when rare observations are limited.
		
	Existing oversampling methods differ primarily in the criteria used to identify important regions and generate additional samples. Based on their underlying balancing mechanisms, the literature can be broadly organized into three categories: \emph{relevance-based methods}, which rely on user-defined or data-driven relevance functions to identify rare target regions; \emph{density-aware methods}, which explicitly incorporate local density information to guide the balancing process; and \emph{deep generative methods}, which employ representation learning and generative models to synthesize new observations. The following subsections review these three directions and discuss their respective developments in the imbalanced regression literature.

	\paragraph{A. Relevance-Based Methods}
	In imbalanced regression, \emph{relevance} quantifies the importance of target values according to their rarity, utility, or domain-specific significance \cite{Ribeiro2020, BRANCO201976}. Since regression problems lack explicit minority labels, relevance-based methods employ a \emph{relevance function} to map continuous target values onto an importance scale. Observations whose relevance exceeds a predefined threshold are considered rare or critical and are subsequently oversampled to improve their representation in the training data.
	
	The central premise of these methods is that not all target values contribute equally to the learning objective. By explicitly prioritizing high-relevance regions, relevance-based oversampling directs the learning process toward underrepresented yet important observations, making it particularly suitable for applications in which extreme or uncommon target values are of greater practical interest than densely populated regions.
	
	A key advantage of this family of methods is its simplicity and interpretability. The relevance function provides a transparent mechanism for identifying regions that require additional attention, which has established relevance-based oversampling as one of the foundational paradigms in data-level imbalanced regression.

	Early relevance-based studies established the utility-oriented view of imbalanced regression by arguing that rare continuous target values should receive greater attention than frequent ones. In this direction, Torgo et al. introduced SMOTER \cite{SmoteR2013}, one of the first adaptations of SMOTE \cite{chawla2002smote} to regression, where rare cases are identified through a relevance function and synthetic samples are generated around them to reduce the imbalance between rare and frequent target regions. The same line of work was later extended through broader resampling strategies for regression, including both oversampling and undersampling mechanisms, emphasizing the model-independent nature of preprocessing-based solutions \cite{Torgo2015}. Branco et al. \cite{branco2016ubl} further systematized utility-based learning through the UBL package, which provided practical implementations of relevance-based resampling methods for both classification and regression tasks. Building on these foundations, SMOGN \cite{SMOGN2017} combined SMOTER-style interpolation with Gaussian-noise-based generation, allowing the sampling strategy to vary according to neighborhood characteristics and thereby improving flexibility in rare-region augmentation. More recent extensions have sought to improve the quality and placement of synthetic samples. For example, G-SMOTE for regression expands the synthetic generation region from a simple line segment to a geometric space around selected observations, increasing sample diversity while preserving the preprocessing nature of the method. Similarly, WSMOTER \cite{Camacho2024} introduces a weighting mechanism into SMOTE-based regression oversampling, aiming to generate synthetic samples in a more importance-aware manner. Complementary to these sampling methods, studies on imbalanced regression and extreme-value prediction formalized relevance estimation and evaluation through non-parametric relevance functions and metrics such as SERA, highlighting the importance of aligning model assessment with rare-value prediction \cite{Ribeiro2020}.

	Despite their importance, this family of methods also exhibits several limitations. Their performance is highly dependent on the definition of the relevance function and the threshold used to separate rare and normal target regions. In many cases, these choices are either user-defined or derived from simple target-distribution heuristics, which may not fully reflect the structural complexity of the data. Moreover, relevance is typically estimated mainly from the target space, while feature-space geometry, local density variation, and neighborhood consistency may receive limited attention. As a result, synthetic samples generated around relevant observations may increase minority region representation but still fail to preserve the local feature--target relationships required for reliable regression modeling.

	\paragraph{B. Density-Aware Methods:}in imbalanced regression, \emph{density} refers to the concentration of observations within a particular region of the target space, feature space, or joint feature--target space. Density-aware oversampling methods use this information to guide the balancing process, with the aim of identifying not only rare target values but also locally underrepresented regions where the available samples may be insufficient to characterize the underlying data distribution \cite{Steininger2021, SHAHBAZI2026}. Rather than relying solely on a fixed relevance threshold, these methods typically estimate distributional structure through mechanisms such as kernel density estimation (KDE), neighborhood statistics, clustering, or local distribution modeling, and then generate or emphasize samples according to the estimated scarcity of different regions.
	
	The main advantage of density-aware methods is that they provide a more data-adaptive view of imbalance. By considering how observations are distributed across continuous regions, they can reduce the dependence on manually defined rare-region boundaries and better account for gradual changes in target density. This is particularly important in regression, where sparsity is often local, continuous, and heterogeneous rather than confined to clearly separated minority regions. As a result, density-aware oversampling can potentially improve the fidelity of synthetic samples and produce a more balanced representation of both rare and frequent target ranges.
	
	Density-aware thinking in imbalanced regression has first appeared mainly through weighting and distribution-adjustment strategies. DenseWeight and DenseLoss estimate target-value rarity using kernel density estimation and increase the training influence of rare observations through density-based weights, demonstrating the usefulness of density information for improving rare-region prediction, although the method modifies the learning objective rather than the data distribution \cite{Steininger2021}. Similarly, distribution-smoothing approaches such as label distribution smoothing (LDS) and feature distribution smoothing (FDS) calibrate label and feature distributions by exploiting the continuity of nearby target values \cite{yang2021delving}, while Balanced MSE reformulates the regression loss to account for imbalanced label distributions \cite{Ren2022Je}; however, these methods are primarily algorithm-level rather than oversampling-based. Within data-level learning, GOLIATH introduced a KDE-based augmentation framework that unifies perturbation- and interpolation-based synthetic generation and applies it to imbalanced regression through density-guided sample generation \cite{stocksieker2024}. More recently, SDE-OS proposed a stacking-based density estimation mechanism to better capture complex target distributions and synthesize rare instances in a more fine-grained manner \cite{Xin2025Stacking}. LDAO further advanced this direction by avoiding hard rare/frequent categorization and decomposing the data into local distributions, from which samples are generated to obtain a more balanced representation across the target range while preserving local statistical structure \cite{alahyari2025}. These studies show a clear movement from fixed minority region definitions toward adaptive density- or distribution-guided balancing, but they differ in whether density is used for weighting, loss correction, distribution smoothing, or actual data generation.
	
	Although density-aware methods provide a more adaptive treatment of imbalance than fixed-threshold strategies, several challenges remain. Many existing approaches estimate density mainly in the target space, which may be insufficient when samples with similar target values exhibit heterogeneous feature-space structures. Conversely, methods that model joint or local distributions often require additional design choices, such as bandwidth selection, clustering parameters, neighborhood size, or assumptions about local distributional form. These choices can strongly affect the quality and location of generated samples. Moreover, when density estimation is performed in the original feature space, high dimensionality and noisy attributes may reduce the reliability of local density estimates. These limitations suggest that density-aware oversampling still requires mechanisms that can jointly preserve target-space imbalance information and the structural organization of the feature space.

	\paragraph{C. Deep Generative Methods:}deep generative oversampling methods address imbalanced regression by learning an underlying data-generating mechanism and using it to synthesize additional observations for underrepresented target regions. Unlike interpolation-based methods, which typically construct new samples from local neighborhoods, generative approaches aim to model more complex and nonlinear feature--target relationships through architectures such as generative adversarial networks (GANs), variational autoencoders (VAEs), and conditional generative models \cite{DOUZAS2018, Engelmann2021, Andrei2021}. In this setting, synthetic samples are produced either directly in the original data space or through a learned latent representation that captures the structure of the input distribution.
	
	The main appeal of deep generative methods lies in their capacity to generate more diverse and distributionally plausible samples, particularly when the relationship between predictors and the continuous target variable is nonlinear or high-dimensional. By learning latent representations or adversarially refining synthetic observations, these methods can potentially overcome some of the rigidity of conventional oversampling strategies and improve sample quality in sparse regions. This makes them especially relevant to recent imbalanced regression studies that seek to move beyond simple interpolation or noise injection.
	
	Recent studies have explored several forms of generative augmentation for regression and imbalanced regression. Early VAE-based augmentation work showed that autoencoder-style generative models can support regression under limited data by learning a latent representation and generating additional samples to improve downstream generalization \cite{Ohno2020}. Building on this idea in the imbalanced-regression setting, Huang et al. \cite{HUANG2022} proposed a conditional VAE-based boosting resampling method, in which rare and normal regions are identified through relevance functions and synthetic samples are generated to improve rare-value prediction. More recent work has moved toward more explicit latent-space and adversarial generation. Stocksieker et al. \cite{Stocksieker2025} developed a disentangled VAE combined with smoothed bootstrap sampling in the latent space, aiming to produce more structured synthetic tabular samples for imbalanced regression. Similarly, SMOGAN uses an initial oversampler followed by a distribution-aware GAN refinement stage, where adversarial learning and distribution alignment are used to improve the consistency of synthetic samples with the joint feature--target distribution \cite{ALAHYARI2026}. Conditional GANs have also been investigated for imbalanced and incomplete time-series regression, showing that generative models can improve extrapolation and generalization when properly constrained, although vanilla GANs may suffer from mode collapse \cite{Hssayeni2024}. In a related direction, IRDA introduced implicit data augmentation for deep imbalanced regression, combining augmentation-aware training with a regression loss designed for non-uniform label and feature distributions \cite{IRDA2024}. Beyond regression, GAN- and VAE-based oversampling has also been explored in imbalanced classification domains such as fraud detection, medical diagnosis, hyperspectral imaging, and low-resource text classification \cite{Liu2025Yu, ZHI2025100557, ALZAHRANI2025, Chaoqian2024, KAUSHIK2026, MANGA2026}, but these studies are mainly relevant as methodological background rather than direct solutions for imbalanced regression.

	Despite their flexibility, deep generative oversampling methods also face important limitations in imbalanced regression. First, training generative models in sparse target regions is inherently difficult, because the very regions that require augmentation often contain too few samples to support reliable distribution learning. Second, generative models may produce visually or statistically plausible observations that do not preserve the precise feature--target consistency required for regression. Third, these approaches usually introduce additional architectural choices, training instability, and computational cost, which may reduce their practicality compared with simpler preprocessing methods. Consequently, while deep generative oversampling offers a promising direction, its effectiveness depends strongly on the quality of the learned representation and its ability to respect local target-dependent structure.

	\subsubsection{Undersampling}
	\label{subsec:LitRew_Undersampling}
	\emph{Undersampling} addresses imbalance by reducing the dominance of highly populated target regions, typically by removing or down-selecting observations from frequent or low-relevance regions \cite{aleksic2025}. Its main appeal is computational efficiency, particularly for large datasets, since it can reduce training size while increasing the relative visibility of rare target ranges.
	
	Existing undersampling methods can generally be divided into \emph{random undersampling} and \emph{informed undersampling}. Random undersampling removes observations from overrepresented regions without explicitly considering their structural role in the data distribution. In contrast, informed undersampling relies on additional criteria, such as neighborhood structure, redundancy, clustering, density, or relevance, to decide which observations can be removed with lower risk of information loss. This distinction reflects the central trade-off in undersampling: reducing majority-region dominance while preserving the samples needed to characterize the original feature-target relationship. 
	
	Random undersampling has been widely used as a simple baseline strategy, particularly in classification and clinical prediction studies, where majority samples are removed without an explicit structural criterion \cite{YangCy2024, Komsrimorakot2025}. Its main advantage is computational simplicity, but its direct transfer to imbalanced regression is limited because continuous target values do not provide clear majority-class boundaries and randomly discarded samples may still contain useful information about local feature--target behavior. To reduce this risk, recent work has increasingly focused on informed undersampling. In imbalanced regression, Aleksić et al. \cite{aleksic2025} proposed Selective Undersampling (SUS) and its iterative variant SUSiter, which remove majority-region observations more carefully by considering relevance- and density-based criteria, showing improved performance over random undersampling and several resampling baselines, especially with neural-network learners and high-dimensional data. Similarly, Chen et al. \cite{CHEN2026Dong} introduced Error Distribution Smoothing (EDS), a pruning strategy guided by the mismatch between local function complexity and data density; by removing redundant samples from dense, low-complexity regions, EDS improves robustness and computational efficiency in low-dimensional imbalanced regression. Beyond regression, informed undersampling has also been explored through inverse random undersampling, radial-based pruning, and spatial-distribution-based selection, where majority samples are retained or removed according to decision-boundary diversity, mutual class potential, or local distributional structure \cite{TAHIR20123738, KOZIARSKI2, yan2023Yuan}. Although these methods are mainly classification-oriented, they reinforce the general principle that undersampling should preserve representative structural information rather than remove majority observations indiscriminately.

	Although undersampling can reduce majority-region dominance and improve computational efficiency, its main limitation is the potential loss of informative observations. This issue is particularly important in regression, where target values vary continuously and samples from dense regions may still contain useful information about local trends, transitions, or boundary behavior. Random removal strategies are especially vulnerable to this problem, while informed methods attempt to reduce it by selecting samples more carefully. Nevertheless, their effectiveness depends strongly on the reliability of the selection criterion, and excessive removal may distort the original data distribution. For this reason, undersampling is often less suitable when the preservation of global distributional structure is critical.

	\subsection{Research Gap}
	\label{subsec:research_gap}
	The reviewed literature indicates that existing data-level methods address imbalanced regression from different but still partial perspectives. Relevance-based oversampling methods provide an interpretable mechanism for identifying rare target regions, yet they often rely mainly on target rarity or predefined relevance thresholds. Density-aware methods move beyond this limitation by incorporating distributional information, but many of them focus primarily on target-space density or estimate density directly in the original feature space, where high dimensionality and heterogeneous local structures may reduce reliability. Deep generative methods improve sample diversity through learned representations, but they do not necessarily ensure that synthetic samples follow the density structure of rare regions. Similarly, undersampling can reduce majority-region dominance but may discard informative observations needed to preserve continuous feature--target relationships. Therefore, few existing approaches jointly consider target-space density, feature-space structure, and representation quality within a unified data-level framework. To address this gap, DADIR integrates density-aware adaptive target partitioning, density-regularized representation learning, and latent-space balancing to provide a model-independent framework for imbalanced regression.

	\section{Methods}
	\label{sec:methods}

	\subsection{Datasets}
\begin{figure}[!htbp]
		\centering
		\includegraphics[width=1.0\textwidth]{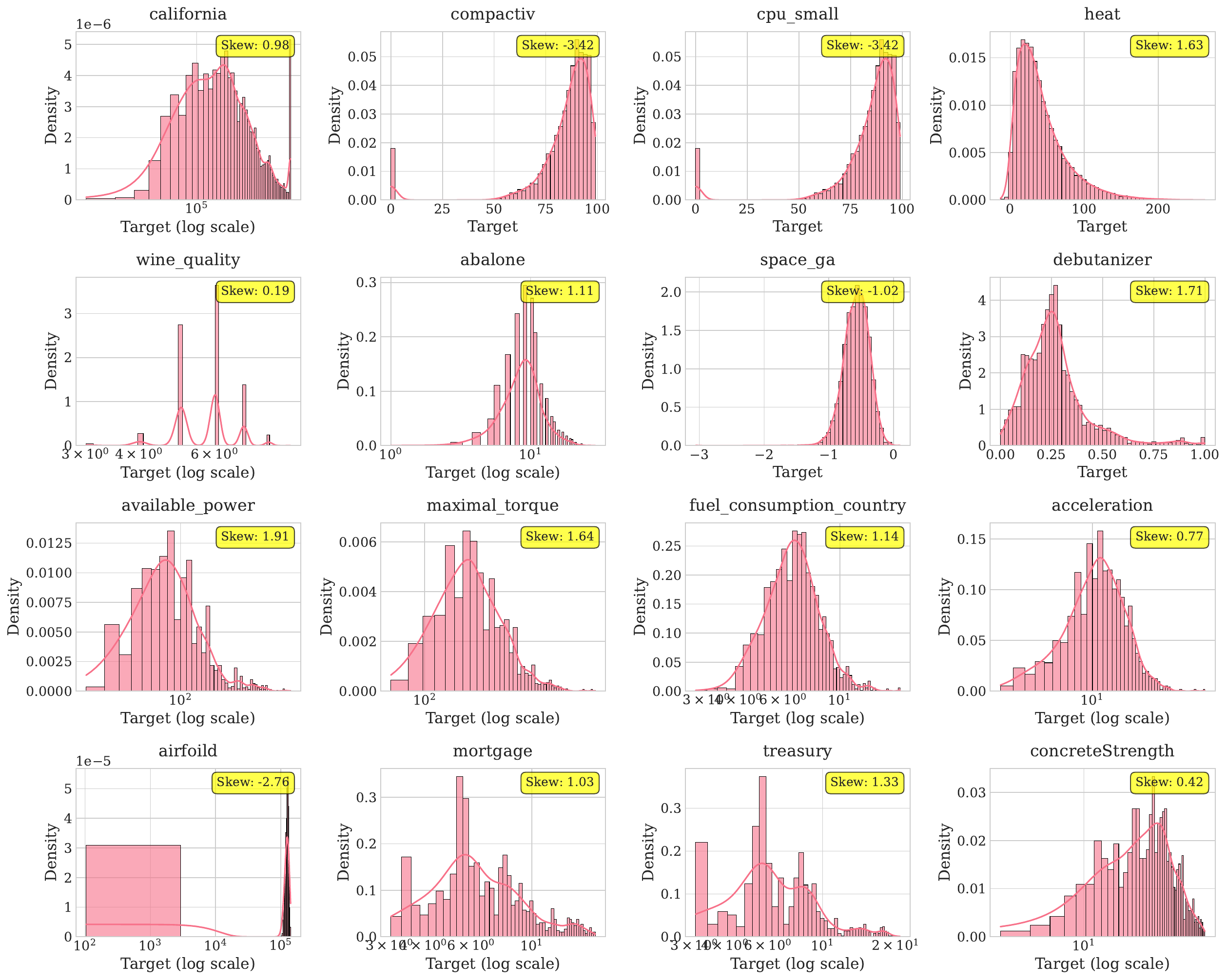} 
		\caption{Target variable distributions for all considered datasets are depicted. Histograms reflect empirical density, and the overlaid curve corresponds to a Kernel Density Estimate (KDE), providing a smooth approximation of the underlying distribution and highlighting skewness. For datasets exhibiting strong positive skewness or large scale variation, a logarithmic x-axis is employed to better visualize tail behavior and facilitate the assessment of regression imbalance.}
		\label{fig:target_distributions_appendix}
	\end{figure}

	To validate the effectiveness of the proposed framework, experiments are conducted on a collection of widely recognized benchmark regression datasets frequently used in imbalanced learning research. A summary of these datasets is presented in Table \ref{tab:datasets}. The selected datasets cover a broad range of application domains, including housing, energy systems, material science, and finance. Importantly, they exhibit varying degrees of target imbalance and skewed distributions, which makes them well-suited for evaluating the performance of imbalanced regression approaches. Fig.~\ref{fig:target_distributions_appendix} illustrates the target variable distributions for all datasets, highlighting noticeable skewness and heavy-tailed behavior, key indicators of imbalance in regression problems.

	\begin{table} [!htbp]
		\tiny
		\centering
		\caption{Benchmark regression datasets applied in this study along with their key characteristics.}
		\label{tab:datasets}
		\begin{tabular}{lcc}
			\hline
			\textbf{Dataset} & \textbf{Number of Samples} & \textbf{Number of Features} \\
			\hline
			california & 20,640 & 8 \\
			compactive & 8,192 & 21 \\
			cpu\_small & 8,192 & 12 \\
			heat & 7,400 & 11 \\
			wine\_quality & 4,898 & 11 \\
			abalone & 4,177 & 8 \\
			space\_ga & 3,107 & 6 \\
			debutanizer & 2,394 & 7 \\
			available\_power & 1,802 & 15 \\
			maximal\_torque & 1,802 & 32 \\
			fuel\_consumption\_country & 1,764 & 37 \\
			acceleration & 1,732 & 14 \\
			airfoild & 1,503 & 5 \\
			mortgage & 1,049 & 15 \\
			treasury & 1,049 & 15 \\
			concreteStrength & 1,030 & 8 \\
			\hline
		\end{tabular}
	\end{table}

	\subsection{System Architecture}
\begin{figure}[htbp]
		\centering
		\includegraphics[width=1.0\textwidth]{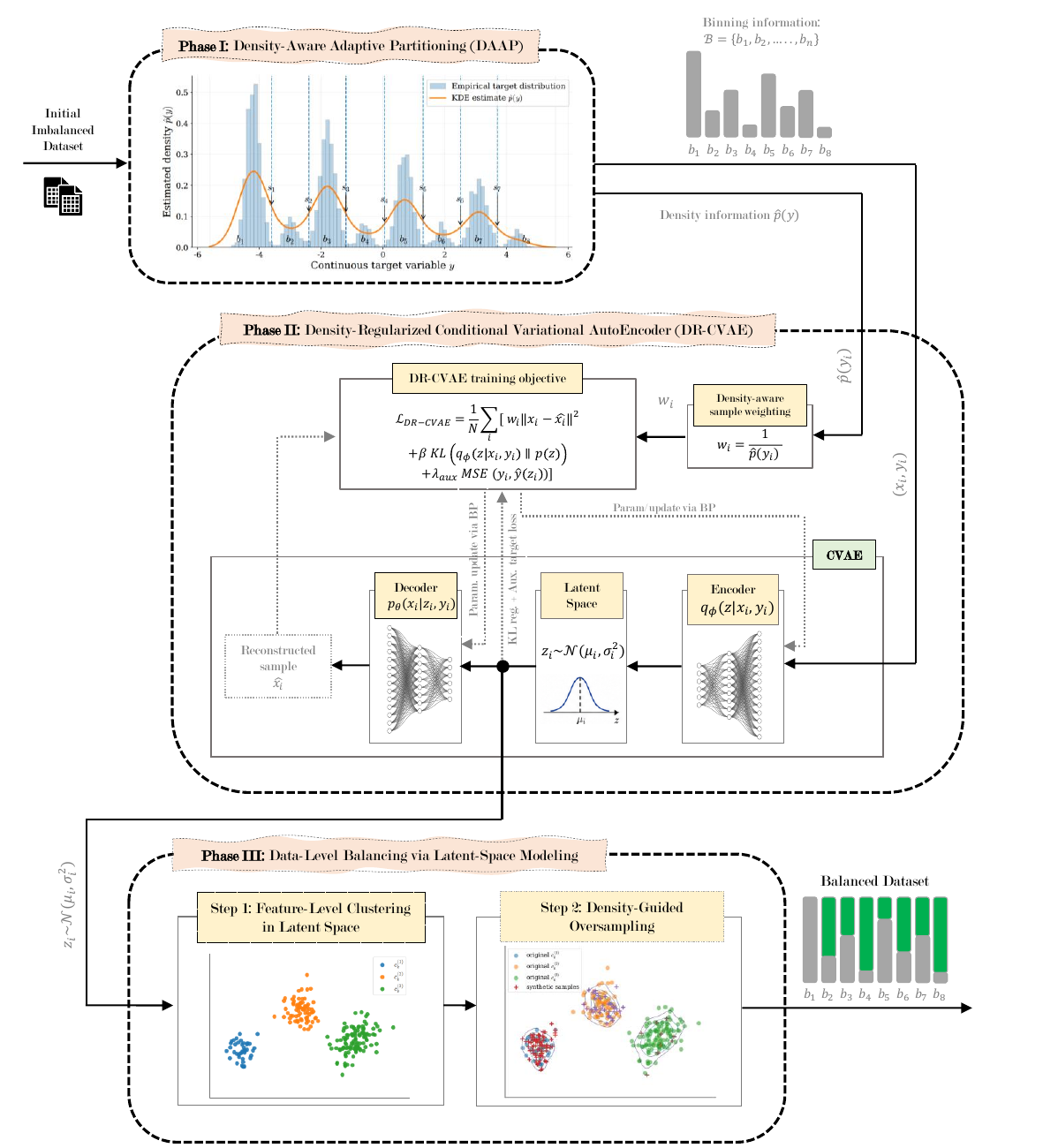} 
		\caption{The general architecture of the proposed system (DADIR)}
		\label{fig:general_Architecture}
	\end{figure}

	In response to the gaps identified in the \hyperref[subsec:research_gap]{Research Gap} section and the research questions presented in the \hyperref[sec:Introduction]{Introduction}, a three-phase architecture is proposed, as illustrated in Fig.~\ref{fig:general_Architecture}. This framework introduces a data-level balancing strategy for regression that adaptively partitions the target space into bins and augments the dataset by generating synthetic samples that jointly preserve feature-space geometry and align with the local target distribution.
	
	\subsection{Phase I: Density-Aware Adaptive Partitioning (DAAP)}
	\label{sec:phase0}
	In imbalanced regression, the target variable \(y \in \mathbb{R}\) is continuous and lacks inherent class boundaries, making the identification of sparse and structurally heterogeneous regions inherently challenging. This limitation propagates to subsequent balancing stages, where poorly represented or highly non-uniform regions can lead to unreliable synthetic sample generation. To address this issue, we introduce \emph{Density-Aware Adaptive Partitioning (DAAP)}, a fully data-driven strategy that recursively partitions the target space into density-consistent regions based on \emph{kernel density estimation (KDE)} and \emph{local density valleys}, resulting in bins that reflect the underlying structure of the target distribution.
	
	Let \(D = \{(x_i, y_i)\}_{i=1}^N\) denote the training dataset, where \(x_i \in \mathbb{R}^d\) and \(y_i \in \mathbb{R}\). The core idea of DAAP is to recursively partition the target space into intervals that are locally consistent in terms of density structure. Unlike conventional binning strategies such as fixed-width or quantile-based partitioning, DAAP adapts its bins to the intrinsic shape of the target distribution by leveraging kernel density estimation and density-valley-based splitting, thereby producing partitions that better reflect the underlying imbalance patterns.
	
	To enable density-aware and numerically stable partitioning of the continuous target space, we first introduce a uniform evaluation grid over each candidate interval. Specifically, for any interval $I = [a,b]$, we construct a finite grid $ \mathcal{G} = \{y_g\}_{g=1}^{G} \subset [a,b], $, which approximates the continuous support of the target space. This discretization allows efficient numerical evaluation of the kernel density function and provides a stable mechanism for identifying local density valleys, which are otherwise difficult to locate in closed form in continuous settings.
	
	The algorithm is initialized with a single interval covering the full target range, $ [\min(y), \max(y)]$. This interval represents the entire support of the target distribution and is iteratively refined through recursive partitioning. For each interval $I = [a,b]$, we estimate the local density using KDE. The estimated density over the interval is defined as Eq.~\ref{eq:kde}:
	
	\begin{equation}
		\label{eq:kde}
		\hat{p}_{S_I}(y) = \frac{1}{|S_I|h} \sum_{y_j \in S_I} K\!\left(\frac{y - y_j}{h}\right),
	\end{equation}
	where $S_I = \{y_i \mid a \le y_i \le b\}$ denotes the subset of target values within the interval, $K(\cdot)$ is a Gaussian kernel, and $h$ is the bandwidth parameter. To ensure a fully data-driven estimation without manual tuning, the bandwidth is selected using Silverman’s rule of thumb as Eq.~\ref{eq:bandwidth}:
	\begin{equation}
		\label{eq:bandwidth}
		h = 0.9 \cdot \min\left(\sigma_y, \frac{\mathrm{IQR}(y)}{1.34}\right) N^{-1/5},
	\end{equation}
	where $\sigma_y$ is the standard deviation of the target values and $\mathrm{IQR}(y)$ denotes the interquartile range. This choice provides a robust and scale-adaptive density estimate, particularly suitable for skewed or heavy-tailed target distributions.
	
	The discrete evaluation of the KDE is then performed on the grid points $\mathcal{G}$, yielding $\hat{p}_{S_I}(y_g)$ for all $y_g \in \mathcal{G}$. The split location is determined as the global minimum of the estimated density as Eq.~\ref{eq:argmin}:
	\begin{equation}
		\label{eq:argmin}
		s = \arg\min_{y_g \in \mathcal{G}} \hat{p}_{S_I}(y_g),
	\end{equation}
	which corresponds to the most pronounced density valley within the interval. This mechanism ensures that partitioning is guided by intrinsic structural sparsity in the target distribution rather than arbitrary geometric rules. Splitting at density valleys is motivated by the fact that local minima of the estimated density function act as natural separators between adjacent high-density regions (modes), whereas density peaks correspond to coherent data concentrations that should remain intact. By aligning partitions with density modes, the proposed scheme confines subsequent oversampling to locally consistent regions, which mitigates cross-mode sample synthesis and leads to more realistic and structure-preserving synthetic data generation in the following phases.
	
	To prevent over-fragmentation of the target space, we define a data-dependent minimum bin size as Eq.~\ref{eq:minbinsize}. This threshold controls recursive refinement by ensuring that no interval with fewer than $M$ samples is further split. The sublinear growth of $M$ with respect to $N$ allows finer partitioning in large datasets while maintaining stability in sparse regions. Without this constraint, the recursive splitting process may lead to excessive fragmentation, resulting in overly small bins that are unsuitable for reliable density estimation and downstream learning.
	\begin{equation}
		\label{eq:minbinsize}
		M = \left\lceil \frac{N}{(\log N)^2} \right\rceil,
	\end{equation}
		
	An interval $I$ is split only if its associated subset satisfies $|S_I| \ge M$ and the detected split point does not lie on the interval boundary. Otherwise, the interval is accepted as a final bin. This ensures that only sufficiently supported and structurally meaningful regions are further partitioned.
	
	The procedure follows a recursive partitioning strategy. Splitting continues until no interval satisfies the conditions for further refinement, yielding a final adaptive bin set $\mathcal{B} = \{b_1, \dots, b_B\}$. By construction, the resulting bins are both data-adaptive and density-consistent, capturing the intrinsic structure of the target distribution while avoiding over-segmentation. Algorithm~\ref{alg:DAAP} demonstrates step-by-step flow of DAAP.

	
	\begin{algorithm}[!htbp]
		\caption{Phase 1: Density-Aware Adaptive Partitioning (DAAP)}
		\label{alg:DAAP}
		\scriptsize
		\begin{algorithmic}[1]
			
			\Require Target values $\{y_i\}_{i=1}^{N}$
			\Ensure Adaptive bin set $\mathcal{B}$
			
			\Statex \textbf{Step 1: Initialization}
			
			\State Compute minimum bin size:
			\[
			M \gets \left\lceil \frac{N}{(\log N)^2} \right\rceil
			\]
			
			\State Initialize stack:
			\[
			\mathcal{S} \gets \{[\min(y), \max(y)]\}
			\]
			
			\State Initialize final bin set:
			\[
			\mathcal{B} \gets \emptyset
			\]
			
			\vspace{0.3em}
			\Statex \textbf{Step 2: Recursive density-aware partitioning}
			
			\While{$\mathcal{S}$ is not empty}
			
			\State Pop interval $I=[a,b]$ from $\mathcal{S}$
			
			\State Define:
			\[
			S_I = \{y_i \mid a \le y_i \le b\}
			\]
			
			\Statex \textbf{Step 2-1: stopping condition}
			
			\If{$|S_I| < M$}
			
			\State Add $S_I$ to $\mathcal{B}$
			
			\Else

			\State Generate uniform evaluation grid:
			\Comment{\textcolor{gray}{Approximates the continuous KDE function for numerical valley detection}}
			\[
			\mathcal{G} = \{y_g\}_{g=1}^{G} \subset [a,b]
			\]
			
			\State Evaluate KDE on grid:
			\[
			\hat{p}_{S_I}(y_g)
			=
			\frac{1}{|S_I|h}
			\sum_{y_j \in S_I}
			K\!\left(\frac{y_g - y_j}{h}\right),
			\quad \forall y_g \in \mathcal{G}
			\]
			
			\State Determine split point:
			\[
			s = \arg\min_{y_g \in \mathcal{G}} \hat{p}_{S_I}(y_g)
			\]
			
			\If{$s = a$ or $s = b$}
			\Comment{\textcolor{gray}{Edge-case handling}}
			\State Add $S_I$ to $\mathcal{B}$
			\Else
			\State Define child intervals:
			\[
			I_L = [a,s], \qquad I_R = [s,b]
			\]
			
			\State Push $I_L$ into $\mathcal{S}$
			\State Push $I_R$ into $\mathcal{S}$
			\EndIf

			\EndIf
			
			\EndWhile
			
			\vspace{0.3em}
			\Statex \textbf{Step 3: Output}
			
			\State \Return $\mathcal{B}$
			
		\end{algorithmic}
	\end{algorithm}

	\subsection{Phase II: Density-Regularized Conditional Variational Autoencoder (DR-CVAE)}
	\label{sec:phase2}
	
	Effective learning in imbalanced regression tasks depends not only on the distribution of training samples but also on the quality of the underlying feature representations. Raw input features, particularly in sparse or underrepresented regions, often fail to capture the latent structures necessary for accurate generalization. This issue is more pronounced in imbalanced settings, where minority regions provide insufficient information for reliable model estimation. To address this, Phase II performs representation learning by projecting the data into a compact, structured latent space that preserves both feature dependencies and target-related variations.
	
	To this end, we employ a \emph{Density-Regularized Conditional Variational Autoencoder (DR-CVAE)}, which extends the standard Conditional Variational Autoencoder (CVAE) by incorporating density-aware weighting into the learning objective. Similar to the conventional CVAE, the model is conditioned on the continuous target variable \(y \in \mathbb{R}\), enabling the learned latent representations to align with the target distribution. Specifically, the encoder receives the concatenated input \((x, y)\), while the decoder reconstructs \(x\) conditioned on both the latent variable \(z\) and the target \(y\). This conditioning ensures that the latent space captures target-dependent structures, which is particularly beneficial for modeling sparse regions in imbalanced regression.
	
	Formally, given a dataset \(D = \{(x_i, y_i)\}_{i=1}^N\), where \(x_i \in \mathbb{R}^d\) and \(y_i \in \mathbb{R}\), the DR-CVAE consists of: (i) an encoder \(q_\phi(z \mid x, y)\) that maps each input-target pair to a latent distribution over \(z \in \mathbb{R}^{d_z}\); (ii) a decoder \(p_\theta(x \mid z, y)\) that reconstructs the input from the latent representation and target; and (iii) a prior distribution \(p(z) = \mathcal{N}(0, I)\), which regularizes the latent space toward a standard Gaussian structure.
	
	To account for imbalance, we incorporate sample-wise weights derived from the estimated marginal target density \(\hat{p}(y)\) obtained in Phase I. Each sample is assigned a weight as Eq.~\ref{eq:w_i}:
	\begin{equation}
		\label{eq:w_i}
		w_i = \frac{1}{\hat{p}(y_i)}, \quad \text{normalized such that } \frac{1}{N}\sum_{i=1}^N w_i = 1,
	\end{equation}
	which increases the contribution of samples located in low-density regions of the target space while maintaining a stable overall scale.
	
	The model is trained by minimizing a density-regularized negative-ELBO-style objective, augmented with an auxiliary target-prediction term, defined as Eq.~\ref{eq:dr_cvae}:
	\begin{equation}
		\label{eq:dr_cvae}
		\mathcal{L}_{\text{DR-CVAE}} =
		\frac{1}{N} \sum_{i=1}^N
		\left[
		w_i \cdot \mathbb{E}_{q_\phi(z \mid x_i, y_i)}
		\Bigl[ \|x_i - \hat{x}_i(z)\|^2 \Bigr]
		+ \beta \, \mathrm{KL}
		\Bigl( q_\phi(z \mid x_i, y_i) \,\|\, p(z) \Bigr)
		+ \lambda_{\text{aux}} \cdot
		\mathrm{MSE}(y_i, \hat{y}(z_i))
		\right].
	\end{equation}
	where \(\hat{x}_i(z) = p_\theta(x \mid z, y_i)\), and \(\hat{y}(z_i)\) denotes the output of an auxiliary regression head that predicts the target from the latent representation. The reconstruction term encourages accurate feature reconstruction, the KL divergence enforces a smooth and continuous latent space, and the auxiliary regression term promotes target-aligned latent embeddings.

	The density-weighted reconstruction term plays a central role by allocating greater learning emphasis to samples in sparse regions, thereby enhancing representation quality where it is most needed. At the same time, the KL regularization ensures that the latent space remains well-structured and suitable for generative modeling.
	
	To promote meaningful abstraction and avoid trivial identity mappings, the latent dimensionality is constrained such that \(d_z \ll d\). This is particularly important in low-dimensional tabular settings, where overly expressive models may fail to learn informative representations.
	
	The output of this phase is a set of latent representations \(z_i \sim \mathcal{N}(\mu_i, \sigma_i^2)\) for each sample, where \(\mu_i\) and \(\sigma_i\) are produced by the encoder. These representations define a structured latent space that captures both feature geometry and target-dependent variations. This space serves as the foundation for the clustering and density-guided oversampling procedures in Phase III.


	\subsection{Phase III: Data-Level Balancing via Latent-Space Modeling}
	\label{sec:phase3}
	In supervised learning, model generalization is fundamentally constrained by the availability and distribution of training data. This limitation is particularly critical in imbalanced regression, where the target variable exhibits highly non-uniform density across its domain, leading to biased predictions toward dense regions and poor performance in sparse ones. To mitigate this issue, we propose a structured \emph{two-step data-level balancing} strategy that operates in the latent space learned in Phase II. Unlike conventional methods that manipulate samples directly in the original feature space, our approach performs augmentation in a representation space that captures the joint structure of features and targets. This enables the generation of synthetic samples that are both feature-consistent and target-aware. The proposed pipeline consists of two steps: (i) \emph{feature-level clustering in latent space} to identify structurally sparse regions, and (ii) \emph{density-guided oversampling} to augment these regions in a principled manner.
	
	\subsubsection{Step 1: Feature-Level Clustering in Latent Space}
	\label{sec:phase2_clustering}
	
	Although Phase I partitions the target space into coherent bins, the distribution of samples within each bin may remain uneven in the feature space. Therefore, identifying truly sparse regions requires analyzing the geometry of latent representations. Let \(\{z_i\}_{i=1}^{N_k}\) denote the latent representations corresponding to samples in bin \(b_k\), where \(z_i \in \mathbb{R}^{d_z}\). We define:
	\begin{equation}
		\mathcal{Z}_k = \{ z_i \mid y_i \in b_k \}.
	\end{equation}
	
	To uncover the intrinsic structure of \(\mathcal{Z}_k\), we apply \emph{Hierarchical Density-Based Spatial Clustering of Applications with Noise (HDBSCAN)} \cite{hdbscan2017}, which identifies clusters of varying densities without requiring the number of clusters as input. This property is well-suited to imbalanced settings, where cluster structure is unknown and heterogeneous across bins. Let \(\mathcal{C}_k \) denote the set of clusters discovered in bin \(b_k\). Samples labeled as noise correspond to low-density regions with insufficient structural support; these samples are retained in the dataset but, due to their low local density, exert only limited influence on the KDE-based sampling process in the subsequent step, thereby reducing the likelihood of generating unreliable synthetic instances. This step provides a fine-grained decomposition of the latent feature space, forming the basis for targeted and structure-aware oversampling.
	
	\subsubsection{Step 2: Density-Guided Oversampling}
	\label{subsubsec:step2}
	Following the clustering procedure in Step 1, each bin \(b_k\) is partitioned into a set of clusters \(\mathcal{C}_k\) based on localized density in the latent feature space. In this step, we identify structurally sparse clusters within each bin and perform targeted oversampling to enhance coverage in underrepresented regions. To achieve this, we adopt a two-part strategy that combines \emph{progressive cluster balancing} with \emph{density-guided sample generation}. Specifically, we first determine how clusters should be expanded in a controlled manner, and then generate synthetic samples within each selected cluster based on its local density structure. These two interrelated substeps are described below.
	
	\textbf{A. Progressive cluster balancing:}  
	as a preliminary requirement to sample generation, clusters within each bin are systematically prioritized based on their size. For each bin \(b_k\), the set of identified clusters \(\mathcal{C}_k\) is first ordered in ascending order based on their cardinality. Let \(c_k^{(1)}\) denote the smallest cluster, \(c_k^{(2)}\) the second smallest, and so on, such that $ |c_k^{(1)}| \leq |c_k^{(2)}| \leq \cdots \leq |c_k^{(n)}|$. Based on this ordering, oversampling is performed iteratively as follows:
	\begin{itemize}
		\item The smallest cluster \(c_k^{(1)}\) is oversampled (as described in substep B) until its size matches that of \(c_k^{(2)}\).
		\item Subsequently, clusters \(\{c_k^{(1)}, c_k^{(2)}\}\) are jointly oversampled until they match the size of \(c_k^{(3)}\).
		\item This process continues progressively, incorporating larger clusters at each stage, until the total bin size \(|b_k|\) reaches a target size defined by the largest bin in the dataset.
	\end{itemize}
	
	This strategy ensures that augmentation is concentrated on minority clusters while maintaining a balanced and structure-aware growth across the bin, avoiding over-amplification of already dense regions.
	
	\textbf{B. Density-guided sample generation:}  
	given the clusters selected for augmentation, synthetic samples are generated in the latent space using a density-guided mechanism. For each cluster \(c \subset \mathcal{Z}_k\), we model the local distribution of latent representations using kernel density estimation (KDE) as Eq.~\ref{eq:kde_cluster}:
	\begin{equation}
		\hat{p}_{c}(z) = \frac{1}{|c|} \sum_{z_j \in c} K_h(z - z_j),
		\label{eq:kde_cluster}
	\end{equation}
	where \(K_h(\cdot)\) is a Gaussian kernel with bandwidth \(h\) determined via Scott’s rule. This estimator captures the local geometry of the cluster in latent space. Synthetic latent samples are then drawn as Eq.~\ref{eq:z_sampling}, ensuring that generated points follow the intrinsic structure of the cluster.
	\begin{equation}
		z_{\text{new}} \sim \hat{p}_{c}(z),
		\label{eq:z_sampling}
	\end{equation}

	Then to assign target values, two nearest neighbors of \(z_{\text{new}}\) are identified within the same cluster in latent space. Let \(z_i\) and \(z_j\) denote two such neighbors with corresponding targets \(y_i\) and \(y_j\). The synthetic target is computed via interpolation as Eq.~\ref{eq:y_interpolation}:
	\begin{equation}
		y_{\text{new}} = y_i + \lambda (y_j - y_i), \quad \lambda \sim \mathcal{U}(0,1),
		\label{eq:y_interpolation}
	\end{equation}
	where \(\lambda\) is a random interpolation coefficient from the uniform distribution $\mathcal{U}(0,1)$. This formulation preserves local continuity of the target variable. In this manner, synthetic samples are generated directly in the latent space as pairs \((z_{\text{new}}, y_{\text{new}})\), which are subsequently used for training the downstream regression model. Together, the density-guided sampling mechanism and the progressive balancing strategy enable controlled and structure-preserving augmentation, focusing on informative sparse regions while maintaining global consistency. What is more, in cases where HDBSCAN identifies fewer than two clusters within a bin, the entire bin is treated as a single cluster and oversampling is performed globally using the same density-guided strategy. The output of this step is an augmented dataset \(\tilde{D}\), consisting of both original and synthetic samples, which provides improved coverage of underrepresented regions and supports more balanced model learning. Algorithm~\ref{alg:phase3} presents the step-by-step workflow of Phase III.

	
	\begin{algorithm}[!htbp]  
		\caption{Phase III: Data-Level Balancing via Latent-Space Modeling}
		\label{alg:phase3}
		\scriptsize
		\begin{algorithmic}[1]
			\Require Latent representations \(\{(z_i, y_i)\}_{i=1}^N\), bins \(\mathcal{B} = \{b_k\}_{k=1}^B\)
			\Ensure Augmented dataset \(\tilde{D}\)
			
			\State Initialize \(\tilde{D} \gets D\)
			\State \(target\_size \gets \max_k |b_k|\)
			
			\For{each bin \(b_k \in \mathcal{B}\)}
			
			\State Extract DR-CVAE latent representations: $\mathcal{Z}_k = \{ z_i \mid y_i \in b_k \} $

			\Statex \textbf{Step 1: Feature-Level Clustering in Latent Space}
			\State Apply HDBSCAN on \(\mathcal{Z}_k\) to obtain clusters \(\mathcal{C}_k\)
			
			\Statex \textbf{Step 2: Density-Guided Oversampling (Progressive Balancing)}
			
			\If{$|\mathcal{C}_k| \leq 1$}
			\Comment{\textcolor{gray}{Fallback: when DBSCAN yields just 1 cluster or no clusters, treat the entire bin as a single cluster}}
			
			\State Fit KDE \(\hat{p}_{b_k}(z)\) on \(\mathcal{Z}_k\)
			
			\While{$|b_k| < target\_size$}
			
			\State Sample \(z_{\text{new}} \sim \hat{p}_{b_k}(z)\)
			\State Find nearest neighbors \(z_i, z_j \in \mathcal{Z}_k\)
			\State Sample \(\lambda \sim \mathcal{U}(0,1)\)
			\State \(y_{\text{new}} \gets y_i + \lambda (y_j - y_i)\)
			\State Add \((z_{\text{new}}, y_{\text{new}})\) to \(\tilde{D}\)
			\State Update \(|b_k|\)
			
			\EndWhile
			
			\Else
			\Comment{\textcolor{gray}{When HDBSCAN yields more than 1 clusters in current bin}}
			
			\State Sort clusters in ascending order of size:
			\[
			|c_k^{(1)}| \le |c_k^{(2)}| \le \dots \le |c_k^{(M_k)}|
			\]
			\Comment{\textcolor{gray}{Smallest clusters will be expanded first}}
			
			\For{each cluster \(c_k^{(j)}\)}
			\State Fit KDE \(\hat{p}_{c_k^{(j)}}(z)\)
			\EndFor
			
			\For{$j = 2$ to $M_k$}
			\Comment{\textcolor{gray}{Progressively match cluster sizes}}
			
			\For{$l = 1$ to $j$}
			
			\While{$|c_k^{(l)}| < |c_k^{(j)}|$ \textbf{and} $|b_k| < target\_size$}
			
			\State Sample \(z_{\text{new}} \sim \hat{p}_{c_k^{(l)}}(z)\)
			\State Find nearest neighbors \(z_i, z_j \in c_k^{(l)}\)
			\State Sample \(\lambda \sim \mathcal{U}(0,1)\)
			\State \(y_{\text{new}} \gets y_i + \lambda (y_j - y_i)\)
			\State Add \((z_{\text{new}}, y_{\text{new}})\) to \(\tilde{D}\)
			\State Update \(|c_k^{(l)}|\) and \(|b_k|\)

			\EndWhile
			
			\EndFor
			\EndFor
			
			\EndIf
			
			\EndFor
			
			\State \Return \(\tilde{D}\)
		\end{algorithmic}
	\end{algorithm}

	\subsection{Computational Complexity Analysis}
	\label{sec:computational_complexity}
	To characterize the computational requirements of the proposed multi-phase framework, we analyze the time complexity of each phase, highlighting their scalability and practical feasibility. Let $N$ denote the number of training samples.
		
	The computational cost of Phase I is mainly determined by the recursive density estimation and partitioning of the target space. The target values are first ordered, which requires \(O(N\log N)\) time and allows each candidate interval to be processed efficiently without repeatedly scanning the entire dataset. For each interval, DAAP estimates the local one-dimensional density and identifies a density valley as the split point. Since the density is evaluated over a fixed numerical grid and the intervals at each recursion level are disjoint, the total number of samples processed at one level is linear in \(N\). The minimum bin-size constraint, \(M=\left\lceil N/(\log N)^2 \right\rceil\), prevents excessive fragmentation of the target space and limits the number of recursive refinements. Consequently, assuming that recursive density-valley partitioning produces reasonably balanced interval refinements and that the minimum bin-size constraint prevents pathological fragmentation, the cumulative cost of the recursive density evaluations grows approximately linearly with \(N\) in practice and does not dominate the initial sorting cost. Therefore, the practical time complexity of Phase I is dominated by the ordering step and can be expressed as \(O(N\log N)\). This indicates that DAAP scales efficiently with the number of training samples while providing an adaptive, density-aware partitioning of the continuous target space.
		
	The computational complexity of Phase II (DR-CVAE) is primarily governed by neural network training. Let \(E\) the number of training epochs, and \(C\) the computational cost of a single forward--backward pass through the encoder, decoder, and auxiliary regression head. This cost is determined by the input dimensionality, latent dimensionality, hidden-layer sizes, and the total number of trainable parameters. Since all samples are processed once in each epoch, the overall training complexity is \(O(E \cdot N \cdot C)\). The density-based sample weights \(w_i = 1/\hat{p}(y_i)\) are obtained from the density estimates computed in Phase I and require only \(O(N)\) time for assignment and normalization, introducing negligible overhead compared with model training. Therefore, the overall complexity of this phase remains \(O(E \cdot N \cdot C)\), which scales linearly with the dataset size for a fixed network architecture and a fixed number of epochs.
	
	The computational complexity of Phase III is mainly determined by latent-space clustering, cluster-wise density modeling, and synthetic sample generation. Let \(N_k=|b_k|\) denote the number of samples in bin \(b_k\), and let \(N_{\mathrm{syn}}\) denote the number of generated synthetic samples. In Step 1, HDBSCAN is applied independently within each bin. Its practical complexity is commonly near \(O(N_k\log N_k)\) in low-dimensional spaces when efficient neighborhood search is available, while its worst-case complexity may reach \(O(N_k^2)\) due to distance-based operations \cite{hdbscanCampelo2015}. Aggregated over all bins, the clustering cost therefore ranges from near \(O(N\log N)\) in practice to \(O(N^2)\) in the worst case. In Step 2, KDE-based density modeling is performed within the discovered clusters or within the whole bin in the fallback case. Since these clusters form local subsets of the latent space, the cumulative density-modeling cost remains proportional to the number of processed samples. For each generated synthetic instance, the method samples a latent point from the corresponding local density model and assigns a target value using nearest-neighbor interpolation within the same cluster. With efficient nearest-neighbor search in the low-dimensional latent space, this step requires approximately logarithmic time per generated sample, whereas a naive search may require linear time. Since the balancing process generates at most a dataset-scale number of synthetic samples, namely, \(N_{\mathrm{syn}}=O(N)\), the practical complexity of Phase III is close to \(O(N\log N)\), while the worst-case complexity is \(O(N^2)\). Thus, Phase III remains computationally feasible in practice, particularly because clustering and oversampling are performed within bins and clusters rather than over the entire dataset at once. 
	
	The phase-wise computational costs are summarized in Table~\ref{tab:time_complexity_new}. Phase~I and the practical execution of Phase~III exhibit \(O(N\log N)\)-type behavior, while Phase~II is governed by the DR-CVAE training cost \(O(E\cdot N\cdot C)\). In the worst case, Phase~III may dominate the framework due to density-based clustering and nearest-neighbor operations, leading to \(O(N^2)\) complexity. In practical settings, however, the runtime is mainly governed by DR-CVAE training, assuming a fixed network architecture and a fixed number of epochs. This indicates that the proposed framework remains computationally feasible for moderately large tabular regression datasets.
	
	\begin{tableorg}[htbp]
		\centering
		\caption{Summary of dominant time complexity for the proposed framework. \(N\): number of samples, \(E\): number of training epochs, and \(C\): cost of one forward--backward pass in DR-CVAE.}
		\label{tab:time_complexity_new}
		\resizebox{\textwidth}{!}{
			\begin{tabular}{lcc}
				\toprule
				\textbf{Phase} & \textbf{Dominant Time Complexity} & \textbf{Dominant Operation(s)} \\
				\midrule
				Phase I -- DAAP & \(O(N\log N)\) & Target ordering and recursive density-aware partitioning \\[3pt]
				Phase II -- DR-CVAE & \(O(E\cdot N\cdot C)\) & Encoder--decoder training with auxiliary regression head \\[3pt]
				Phase III -- Latent-Space Balancing & \(O(N^2)\) worst-case; practical \(O(N\log N)\) & HDBSCAN and nearest-neighbor interpolation \\[3pt]
				\midrule
				\textbf{Overall Framework} & \(O(N^2)\) worst-case; practical runtime dominated by \(O(E\cdot N\cdot C)\) & Phase III in the worst case; DR-CVAE training in practice \\
				\bottomrule
			\end{tabular}
		}
	\end{tableorg}


	\section{Results and Discussion}
	\label{sec:resultAndDiscussion}

	\subsection{Evaluation Metrics}
	\label{subsec:evaluation_metrics}
	
	The performance of the proposed imbalanced regression framework is assessed using three standard metrics, detailed below.  
	
	\begin{itemize}

		\item \textit{Mean Absolute Error (MAE)} (Eq.~\ref{eq:mae}) computes the average absolute difference between predictions and ground truth, providing robustness against outliers:
		\begin{equation}
			\text{MAE} = \frac{1}{N} \sum_{i=1}^{N} |y_i - \hat{y}_i|
			\label{eq:mae}
		\end{equation}
		
		\item \textit{Root Mean Squared Error (RMSE)} (Eq.~\ref{eq:rmse}) measures the square root of the average squared deviation between predicted and actual values, expressing the prediction error in the same units as the target variable while remaining sensitive to large errors:
		\begin{equation}
			\text{RMSE} =
			\sqrt{\frac{1}{N}\sum_{i=1}^{N}(y_i-\hat{y}_i)^2}
			\label{eq:rmse}
		\end{equation}
		
		\item \textit{Coefficient of determination ($R^2$)} (Eq.~\ref{eq:r2}) measures the fraction of the variance in the target variable that is explained by the model, thus reflecting overall fit:
		\begin{equation}
			R^2 = 1 - \frac{\sum_{i=1}^{N} (y_i - \hat{y}_i)^2}{\sum_{i=1}^{N} (y_i - \bar{y})^2}
			\label{eq:r2}
		\end{equation}
	\end{itemize}
	
	In the above equations, $y_i$ represents the observed target value, $\hat{y}_i$ the predicted value, $N$ the total number of samples, and $\bar{y} = \frac{1}{N} \sum_{i=1}^{N} y_i$ the mean of the observed target values.

	\subsection{Implementation and Hyperparameter Settings}
	\label{sec:hyperparameters}
	Table~\ref{tab:param_setting_table} lists the phase-specific hyperparameters applied in each phase of DADIR. all the settings are kept fixed across all benchmark datasets and regressors to ensure fair and reproducible comparisons. Before executing the proposed phases and training the regression models, the target variable is standardized to have zero mean and unit variance. Furthermore, all experimental phases were conducted under a 10-fold cross-validation scheme. During each fold, the adaptive binning mechanism of Phase 1 was derived exclusively from the training split and subsequently transferred to the associated validation split, thereby preventing information leakage and ensuring a reliable and stratified evaluation process.
	
	To isolate the impact of the proposed imbalance-aware pipeline, all standalone regressors are employed with their default hyperparameter settings (Table~\ref{tab:standalone_regressors_hyperparams}). Performance is reported using 10-fold cross-validation, with folds stratified based on the Phase 1 bins to preserve minority target regions. This strategy ensures fair evaluation, reduces split-dependent bias, and maintains consistent representation of sparse regions across folds.

	\begin{table}[!htbp]
		\centering
		\scriptsize
		\caption{Hyperparameter setting for DADIR}
		\label{tab:param_setting_table}
		\begin{tabular}{l l l}
			\toprule
			\textbf{Phase} & \textbf{Parameter} & \textbf{Value} \\
			\midrule
			
			\multirow{10}{*}{\textbf{Phase I}}
			
			& Density estimator & KDE \\
			& KDE kernel & Gaussian \\
			& KDE bandwidth rule & Scott \\
			& Bandwidth sensitivity factor & 0.5 \\
			& Density grid size & 512 \\
			& Minimum support & $\left\lceil N / (\log N)^2 \right\rceil$ \\

			\midrule
			
			\multirow{21}{*}{\textbf{Phase II}}
						
			& Encoder hidden dimensions & $128 \rightarrow 64$ \\
			& Decoder hidden dimensions & $64 \rightarrow 128$ \\
			& Latent dimension $d_z$ & 16 \\
			& Encoder/decoder activation & ReLU \\
			& Auxiliary regression head & $16 \rightarrow 32 \rightarrow 1$ \\
			& Density weighting function & $w_i = 1/(\hat{p}(y_i)+10^{-8})$ \\
			& Weight normalization & Mean weight equal to 1 \\
			& KDE kernel for target density & Gaussian \\
			& KDE bandwidth rule & Silverman \\
			& Minimum KDE bandwidth & $10^{-3}$ \\
			& Batch size & 64 \\
			& Optimizer & Adam \\
			& Learning rate & $10^{-3}$ \\
			& Max epochs & 200 \\
			& Early stopping patience & 15 \\
			& KL weight $\beta$ & 0.1 \\
			& Auxiliary loss weight $\lambda_{\text{aux}}$ & 1.0 \\
			& Loss function & Proposed $\mathcal{L}_{\text{DR-CVAE}}$ \\
			& Latent output & Deterministic encoder mean $\mu_i \in \mathbb{R}^{16}$ \\
			
			\midrule
			
			\multirow{19}{*}{\textbf{Phase III}}
			
			& Clustering algorithm & HDBSCAN \\
			& Balancing target & Largest original bin size \\
			& Fallback clustering rule & Whole bin as single cluster if $n_{\text{clusters}} \leq 1$ \\
			& Synthetic generation method & KDE sampling when feasible \\
			& KDE bandwidth rule & Scott \\
			& KDE feasibility condition & $n \geq \max(d_z,2)$ \\
			& Nearest-neighbor model & 2-NN with Euclidean distance \\
			& Target interpolation factor & $\lambda \sim U(0,1)$ \\
			& Interpolation fallback & Linear interpolation for $n \geq 2$ \\
			& Single-sample fallback & Gaussian noise perturbation \\
			& Latent noise scale & $0.01 \times \mathrm{std}(z)$; otherwise 0.01 \\
			& Target noise scale & 0.01 \\
			& Random seed & 42 \\

			\bottomrule
		\end{tabular}
	\end{table}

	
	\begin{tableorg}[!htbp]
		\centering
		\caption{Hyperparameter settings for applied standalone regressors. 
			All models are evaluated using 10-fold cross-validation with shuffled folds 
			and \emph{random state=42} where applicable.}
		\label{tab:standalone_regressors_hyperparams}
		{\scriptsize
			\resizebox{\textwidth}{!}{
				\begin{tabular}{@{}ll||@{\hspace{1.2em}}ll||@{\hspace{1.2em}}ll@{}}
					\toprule
					\textbf{Regressor} & \textbf{Param = Value} &
					\textbf{Regressor} & \textbf{Param = Value} &
					\textbf{Regressor} & \textbf{Param = Value} \\
					\midrule
					
					\emph{MLP} & Architecture = $d \rightarrow 128 \rightarrow 64 \rightarrow 32 \rightarrow 1$ &
					\emph{XGBoost} & Number of trees = 300 &
					\emph{Linear Reg.} & Fit intercept = True \\
					
					& Activation = ReLU &
					& Learning rate = 0.05 &
					& Copy X = True \\
					
					& Loss = MSE &
					& Max depth = 6 &
					& Positive = False \\
					
					& Optimizer = Adam &
					& Subsample = 0.8 &
					& Jobs = None \\
					
					& Learning rate = $10^{-3}$ &
					& Column subsample = 0.8 &
					& Tolerance = -- \\
					
					& Batch size = 128 &
					& Random state = 42 &
					& Random state = -- \\
					
					& Max epochs = 200 &
					& Verbosity = 0 &
					& Solver = -- \\
					
					& Prediction batch size = 128 & & & & \\
					
					\midrule
					
					\emph{KNN} & Neighbors = 5 &
					\emph{SVR} & Kernel = RBF &
					\emph{Ridge} & Regularization $\alpha$ = 1.0 \\
					
					& Weights = distance &
					& $C$ = 100 &
					& Fit intercept = True \\
					
					& Metric = Minkowski &
					& Epsilon = 0.1 &
					& Copy X = True \\
					
					& Power $p$ = 2 &
					& Gamma = scale &
					& Max iterations = None \\
					
					& Algorithm = auto &
					& Degree = 3 &
					& Tolerance = $10^{-4}$ \\
					
					& Leaf size = 30 &
					& Coef0 = 0.0 &
					& Solver = auto \\
					
					& Jobs = None &
					& Shrinking = True &
					& Random state = 42 \\
					
					& & 
					& Cache size = 200 &
					& Positive = False \\
					
					& & 
					& Tolerance = $10^{-3}$ &
					& \\

					\bottomrule
				\end{tabular}
			}
		}
	\end{tableorg}

	\subsection{Results and Discussions}

	\subsubsection{Baseline Methods}
	\label{sec:baselines}
	To comprehensively evaluate the effectiveness of the proposed data-level density-aware imbalanced regression framework, two complementary categories of baseline methods are considered. The first category consists of widely adopted general-purpose regression algorithms that do not explicitly address target imbalance, while the second category includes state-of-the-art methods specifically designed for imbalanced regression learning. This evaluation strategy enables assessment of the proposed framework from both conventional predictive modeling and imbalance-aware learning perspectives.
	
	The inclusion of standard regression algorithms allows investigation of whether the proposed framework can consistently improve predictive performance independently of the underlying regressor architecture. Since the proposed approach operates at the data level, its effectiveness should generalize across different learning paradigms, including linear, distance-based, kernel-based, tree-based, and neural-network-based regressors. Therefore, the following commonly used regression models are employed: \emph{Multi-Layer Perceptron (MLP)}, \emph{XGBoost}, \emph{Linear Regression}, \emph{K-Nearest Neighbors (KNN)}, \emph{Support Vector Regression (SVR)}, and \emph{Ridge} Regression.
	
	In addition, to rigorously evaluate the proposed framework under imbalanced regression settings, comparisons are conducted against representative state-of-the-art imbalance-aware regression methods. These methods encompass different methodological strategies, including synthetic oversampling, distribution smoothing, algorithm-level, and data augmentation-based reweighting. By comparing against methods with diverse imbalance-handling mechanisms, the experiments provide a comprehensive assessment of the strengths and limitations of the proposed framework across varying target distribution characteristics. The following imbalance-aware regression methods are considered in this study:
	
	\begin{itemize}
		
		\item \textbf{SMOTER} \cite{SmoteR2013}: An oversampling approach for imbalanced regression that extends the SMOTE algorithm to continuous target spaces by generating synthetic samples through interpolation between rare target instances.
		
		\item \textbf{WSMOTER} \cite{Camacho2024}: A weighted synthetic oversampling method for imbalanced regression that extends SMOTER by assigning adaptive importance to rare target instances during sample generation to improve prediction performance in underrepresented regions.
		
		\item \textbf{SMOGN} \cite{SMOGN2017}: A preprocessing method for imbalanced regression that combines SMOTE-based synthetic oversampling with Gaussian noise injection to enhance the diversity and representation of rare target regions.
		
		\item \textbf{LDS (Label Distribution Smoothing)} \cite{yang2021delving}: A distribution-aware regression method that smooths the target label distribution by incorporating neighboring continuous target values to better estimate effective sample density in imbalanced regression tasks.
		
		\item \textbf{FDS (Feature Distribution Smoothing)} \cite{yang2021delving}: A feature-level regularization approach that smooths learned feature representations across neighboring target regions to improve generalization under continuous label imbalance.
		
		\item \textbf{FOCAL-R} \cite{focalR2020}: An algorithm-level regression method based on focal loss that dynamically down-weights easy samples and emphasizes hard or underrepresented target instances during training to alleviate imbalance-induced learning bias.
		
		\item \textbf{IRDA (Implicit Reweighting with Data Augmentation)} \cite{IRDA2024}: A deep imbalanced regression framework that combines implicit data augmentation with adaptive sample reweighting through a surrogate regression loss, aiming to mitigate distributional bias and improve learning in sparse target regions.
		
	\end{itemize}

	\subsubsection{Phase I: Density-Aware Adaptive Partitioning (DAAP) Analysis}
	To overcome the lack of explicit class boundaries in imbalanced regression, we introduce the proposed \emph{Density-Aware Adaptive Partitioning (DAAP)} strategy, which partitions the continuous target space according to its intrinsic density structure. Unlike conventional fixed-width or quantile-based partitioning methods that impose predefined and often distribution-agnostic boundaries, DAAP employs KDE and density-valley-based recursive splitting to identify natural separation regions within the target distribution. This results in a fully data-driven segmentation process in which each bin captures a locally coherent density regime.
	
	Fig.~\ref{fig:adaptive_bins_results} illustrates the resulting adaptive partitions produced by DAAP across the evaluated datasets. As observed, the proposed method transforms highly skewed and unevenly populated target distributions into structurally meaningful regions aligned with the underlying density modes and sparse transition areas. By splitting at density valleys, the resulting partitions preserve coherent high-density regions while isolating sparse and minority-sensitive areas, thereby providing a more reliable foundation for subsequent localized oversampling.
	
	Complementing these visual analyses, Table~\ref{tab:bin_statistics} summarizes the partitioning statistics generated by DAAP for each dataset. As can be observed, the proposed density-aware partitioning strategy produces non-uniform bin widths that adapt to the local structure of the target distribution. In some datasets, extremely narrow bins emerge around highly concentrated target regions or repeated target values, reflecting localized density plateaus within the empirical distribution.

	
	\begin{figure}[htbp]
		\centering
		\subfloat[\centering california]{
			\includegraphics[width=0.45\textwidth]{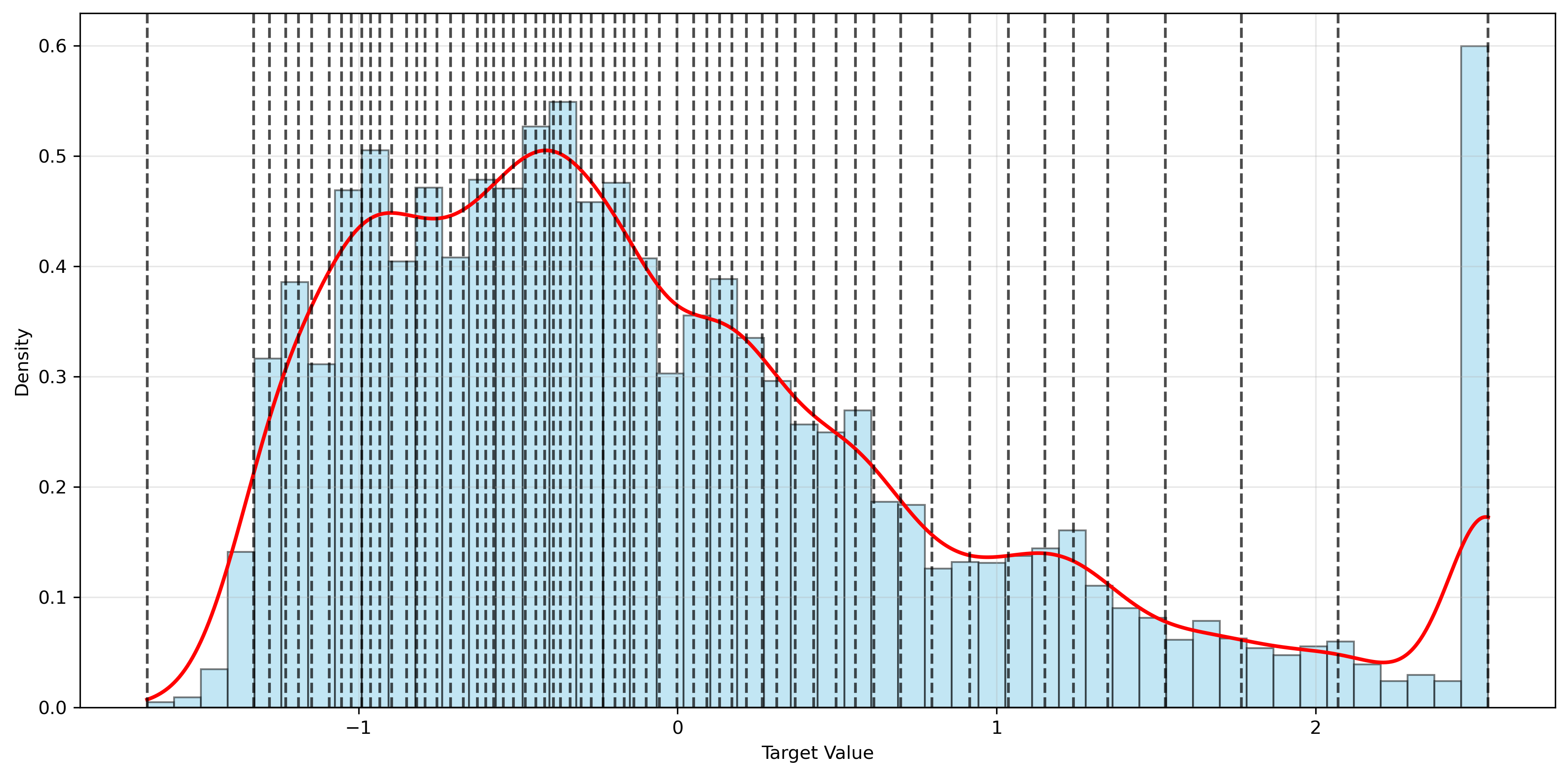}
			\label{fig:raw-California}
		}
		\hfill
		\subfloat[\centering compactive]{
			\includegraphics[width=0.45\textwidth]{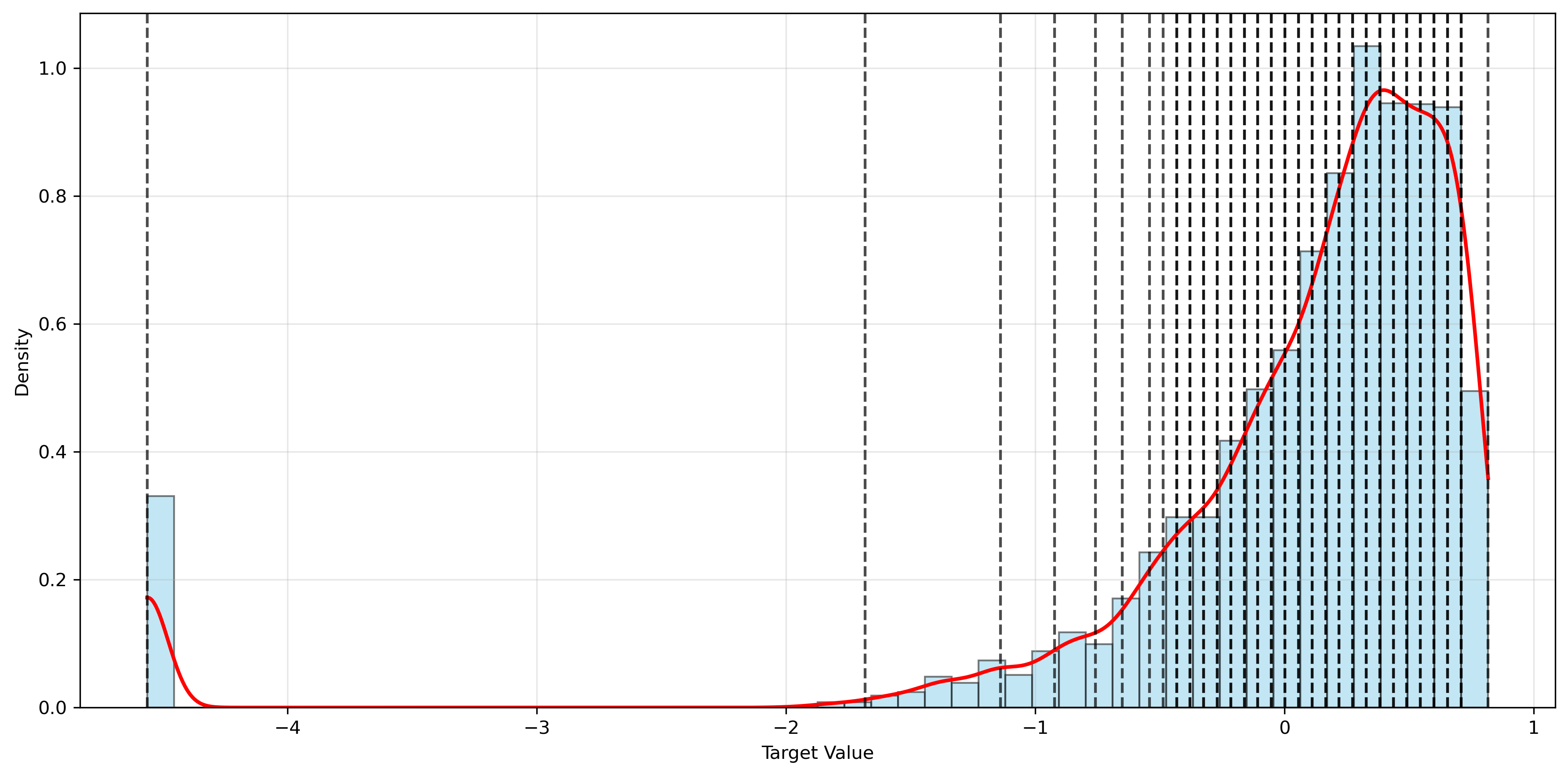}
			\label{fig:binned_compactive}
		}
		
		\subfloat[\centering cpu\_small]{
			\includegraphics[width=0.45\textwidth]{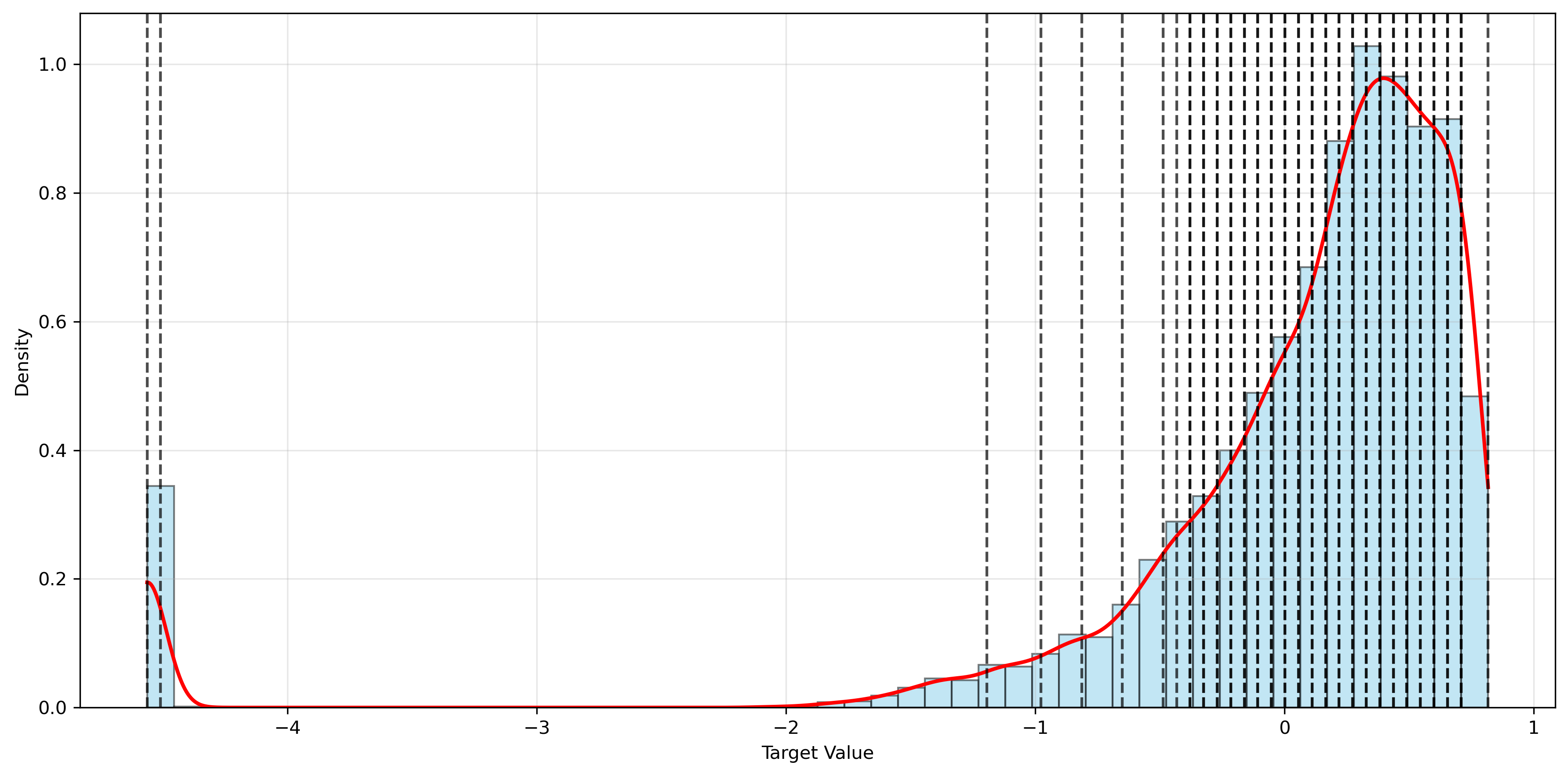}
			\label{fig:raw-cpuSmall}
		}
		\hfill
		\subfloat[\centering heat]{
			\includegraphics[width=0.45\textwidth]{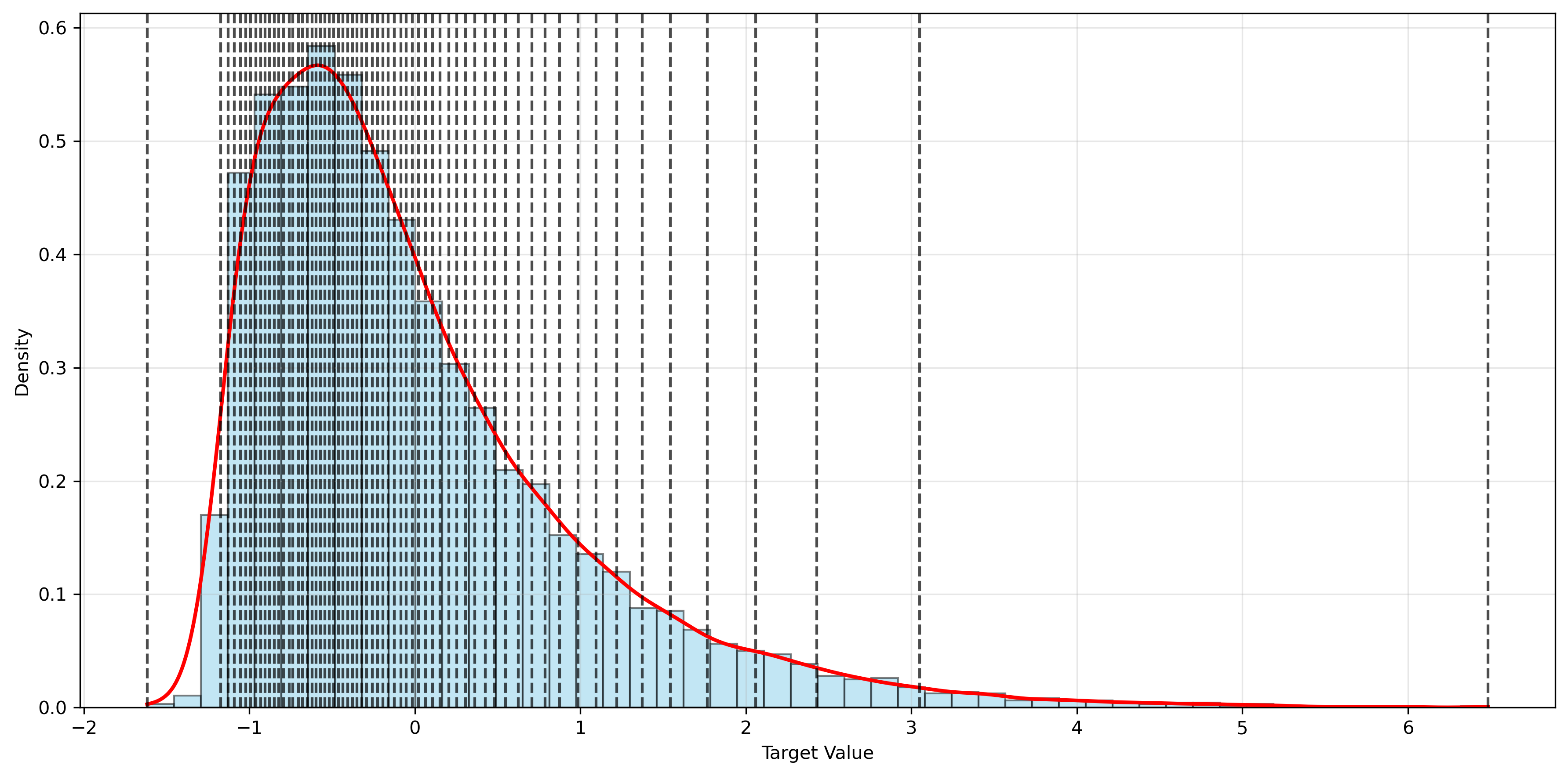}
			\label{fig:binned_heat}
		}
		
		\subfloat[\centering wine\_quality]{
			\includegraphics[width=0.45\textwidth]{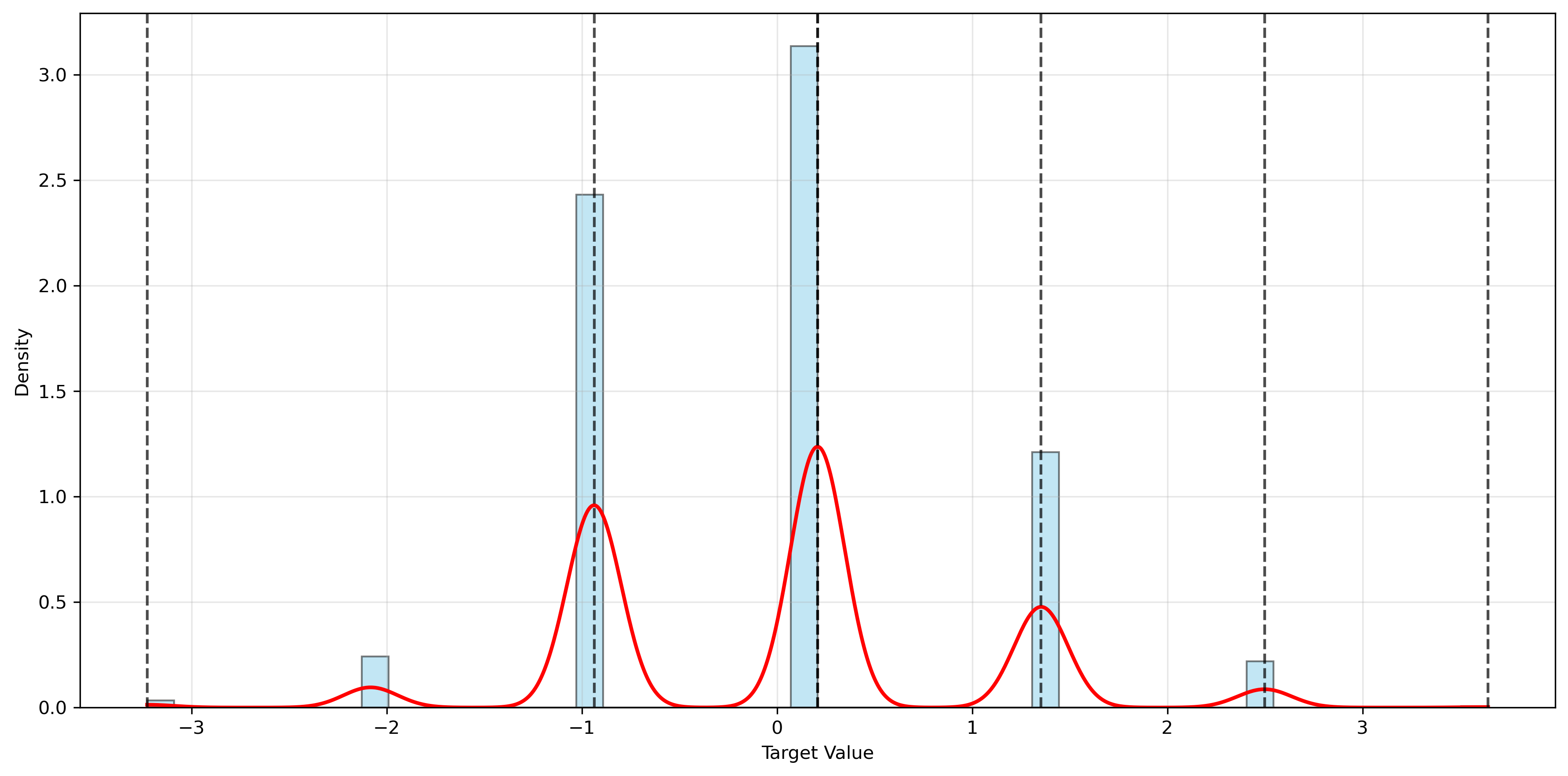}
			\label{fig:winequality}
		}
		\hfill
		\subfloat[\centering abalone]{
			\includegraphics[width=0.45\textwidth]{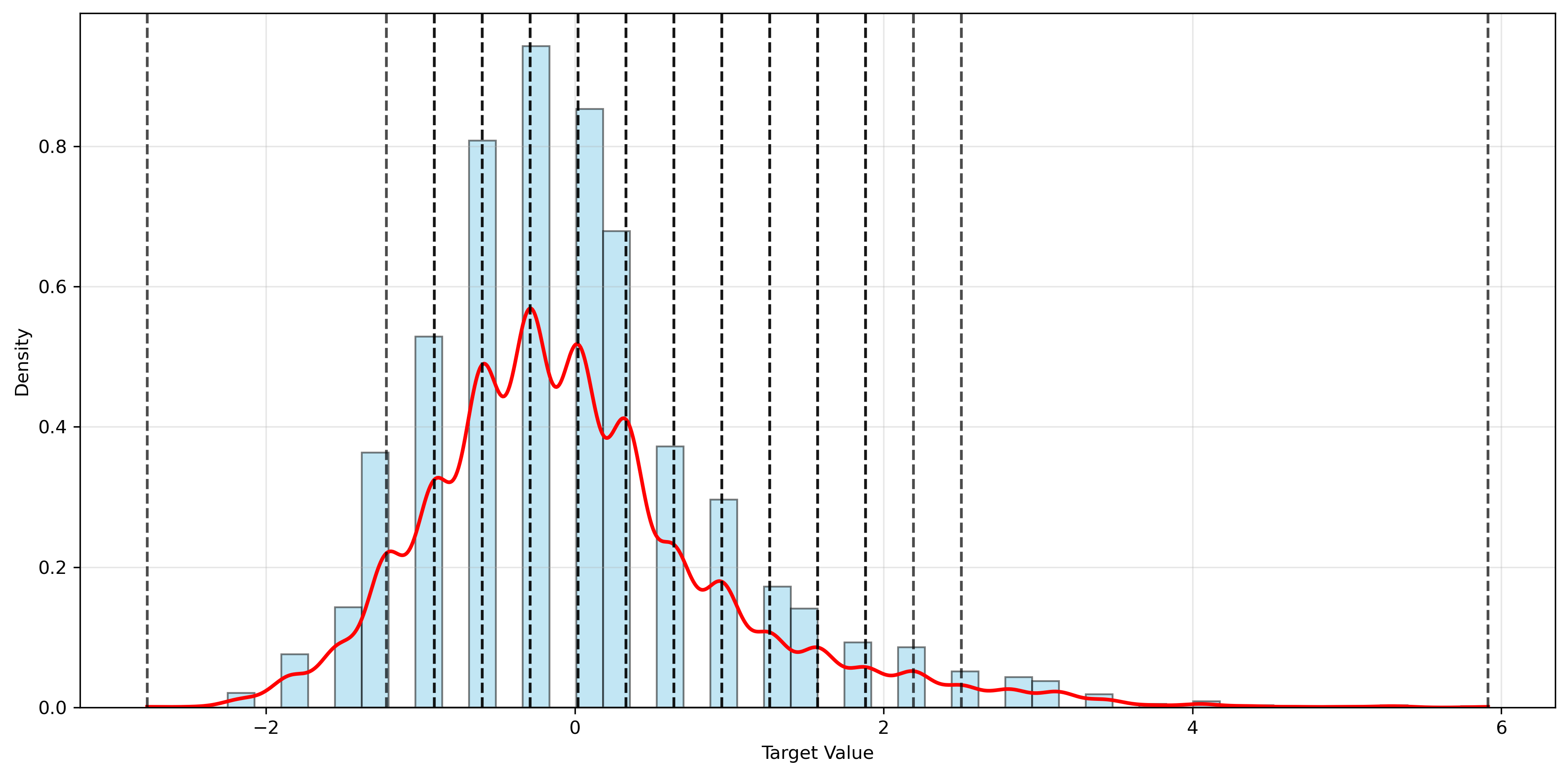}
			\label{fig:binned_abalone}
		}
		
		\subfloat[\centering space\_ga]{
			\includegraphics[width=0.45\textwidth]{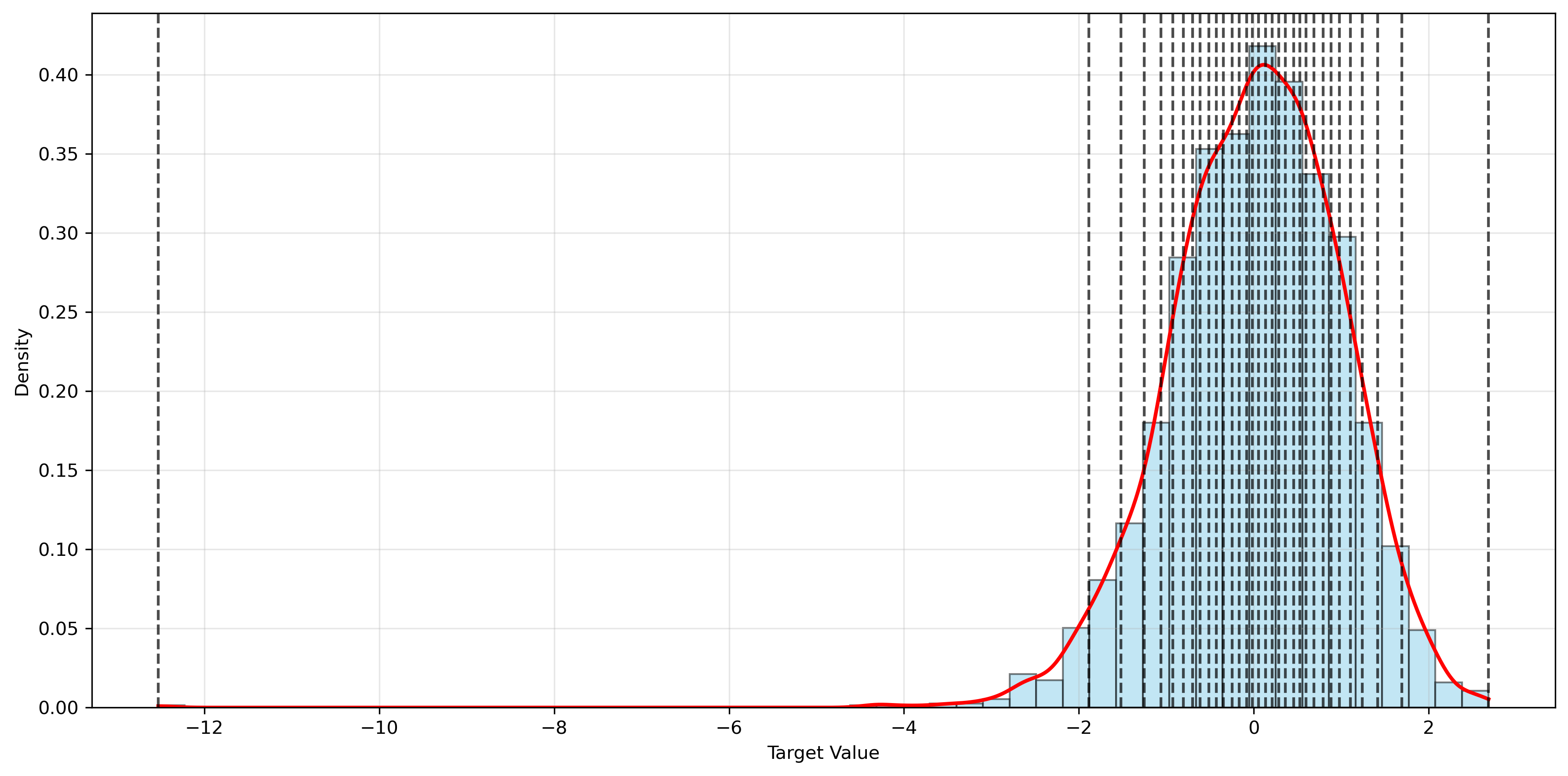}
			\label{fig:binned_spaceGa}
		}
		\hfill
		\subfloat[\centering debutanizer]{
			\includegraphics[width=0.45\textwidth]{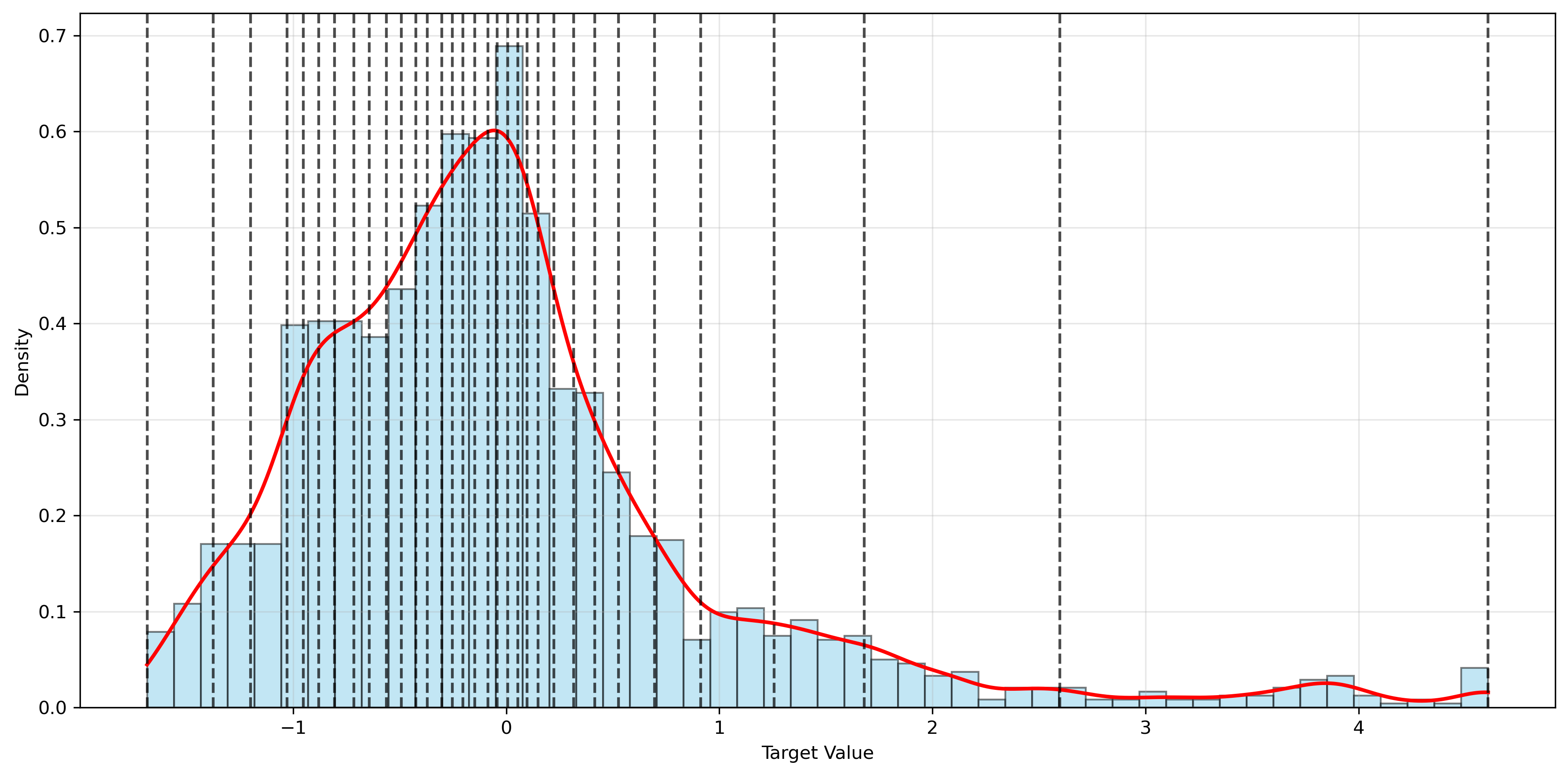}
			\label{fig:binned_debutanizer}
		}
		
		\subfloat[\centering available\_power]{
			\includegraphics[width=0.45\textwidth]{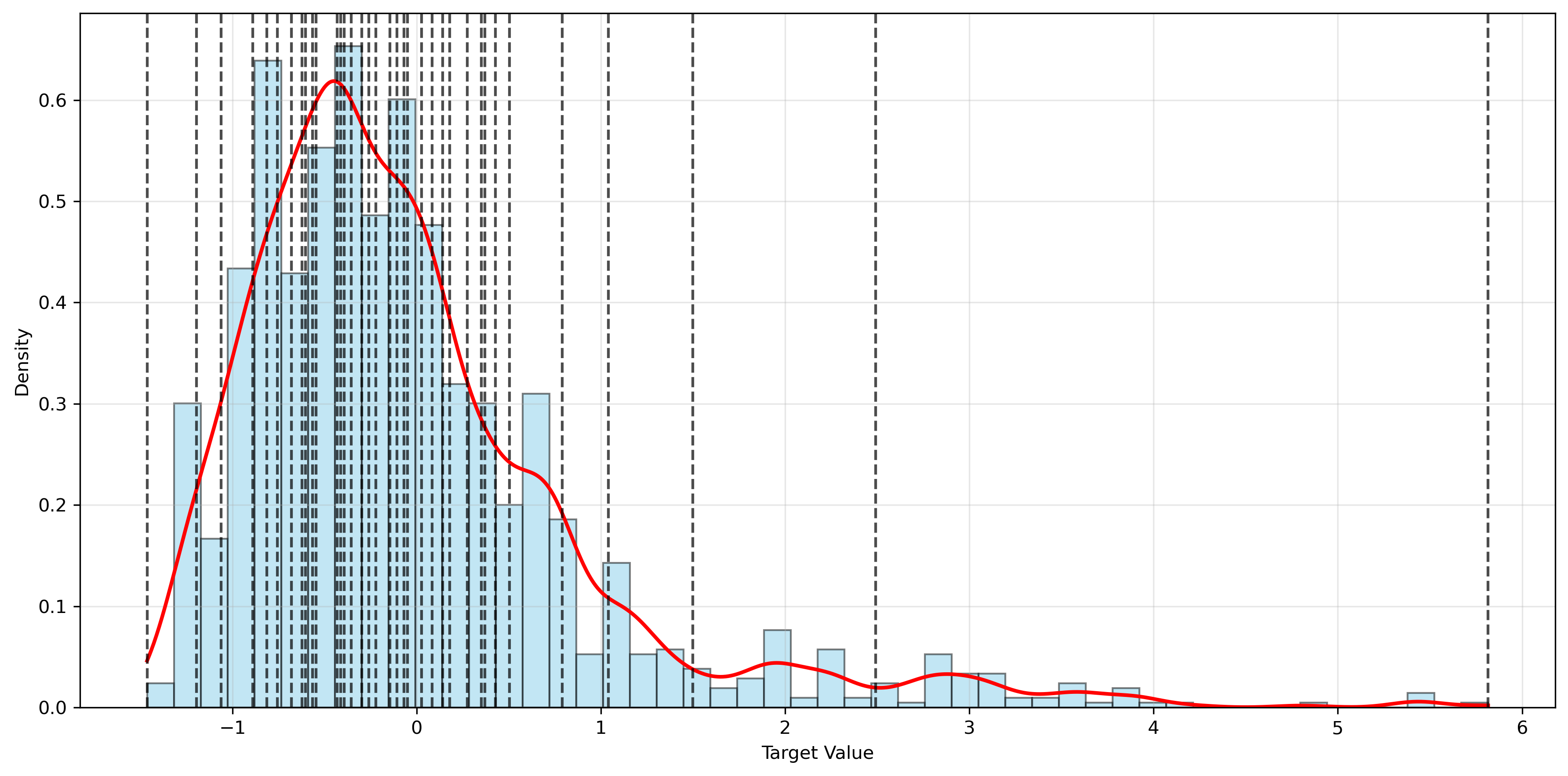}
			\label{fig:binned_available_power}
		}
		\hfill
		\subfloat[\centering maximal\_torque]{
			\includegraphics[width=0.45\textwidth]{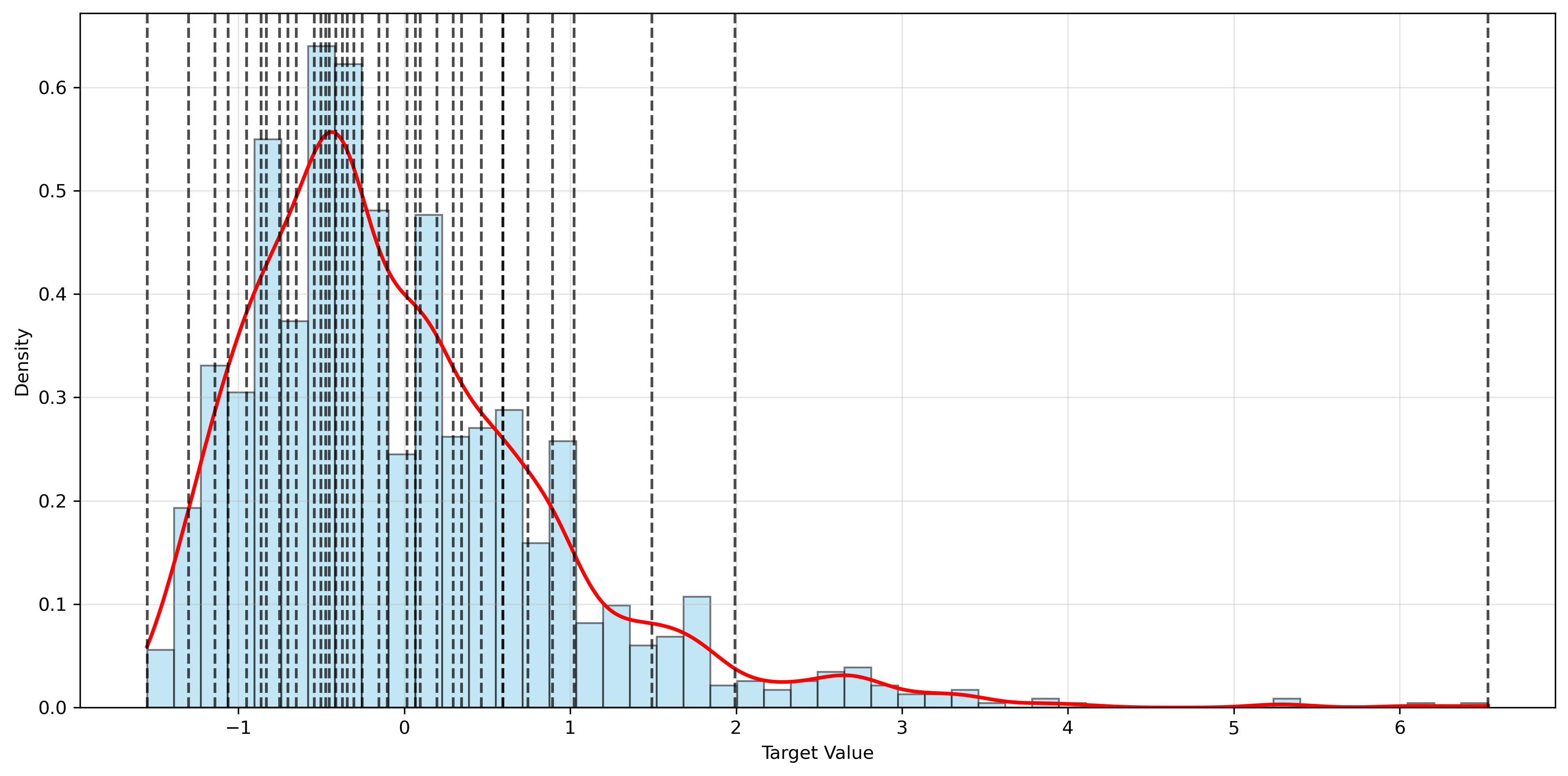}
			\label{fig:binned_maximal_torque}
		}
		
		\caption{Visualization of the proposed Density-Aware Adaptive Partitioning (DAAP) across the evaluated datasets. Histograms represent the standardized target distributions (z-score normalized), while the overlaid KDE curves illustrate the underlying density structure. Vertical dashed lines denote the adaptive bin boundaries identified through recursive density-valley-based partitioning.}
		\label{fig:adaptive_bins_results}
	\end{figure}
	
	\begin{figure}[!htbp]
		\ContinuedFloat
		\centering
		
		\subfloat[\centering fuel\_consumption\_country]{
			\includegraphics[width=0.45\textwidth]{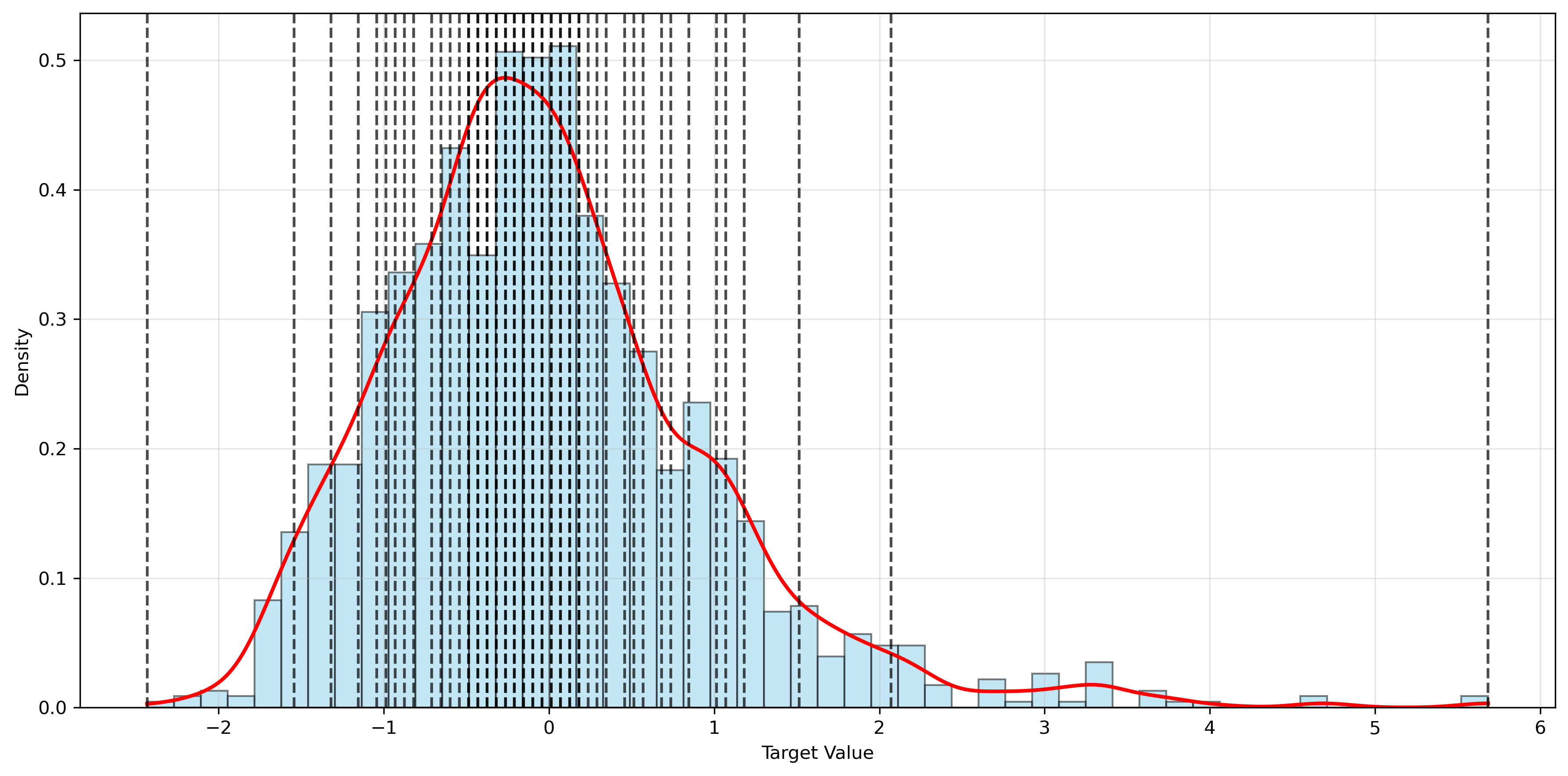}
			\label{fig:bins_fuel_consumption_country}
		}
		\hfill
		\subfloat[\centering acceleration]{
			\includegraphics[width=0.45\textwidth]{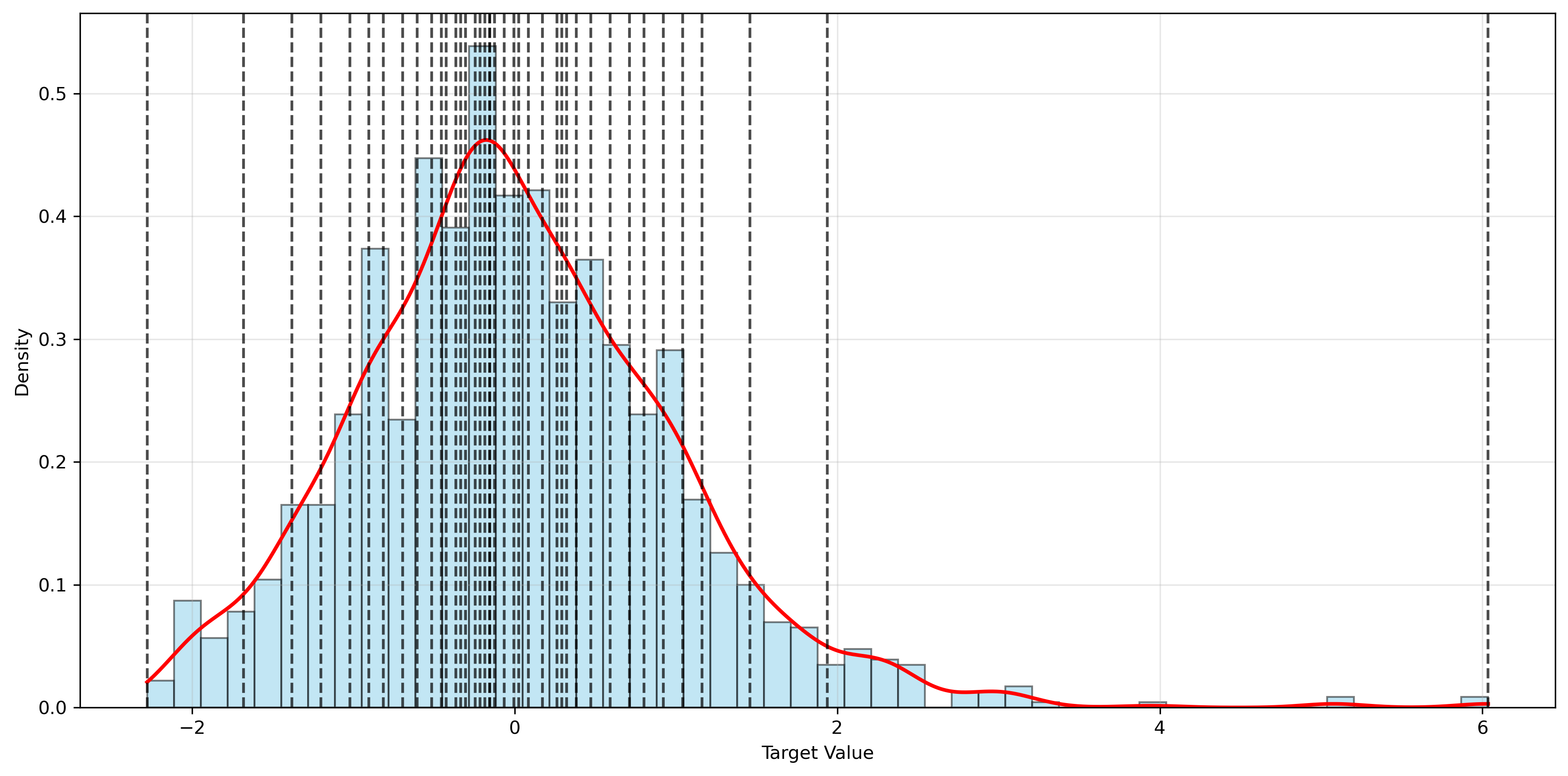}
			\label{fig:binned_acceleration}
		}
		
		\subfloat[\centering airfoild]{
			\includegraphics[width=0.45\textwidth]{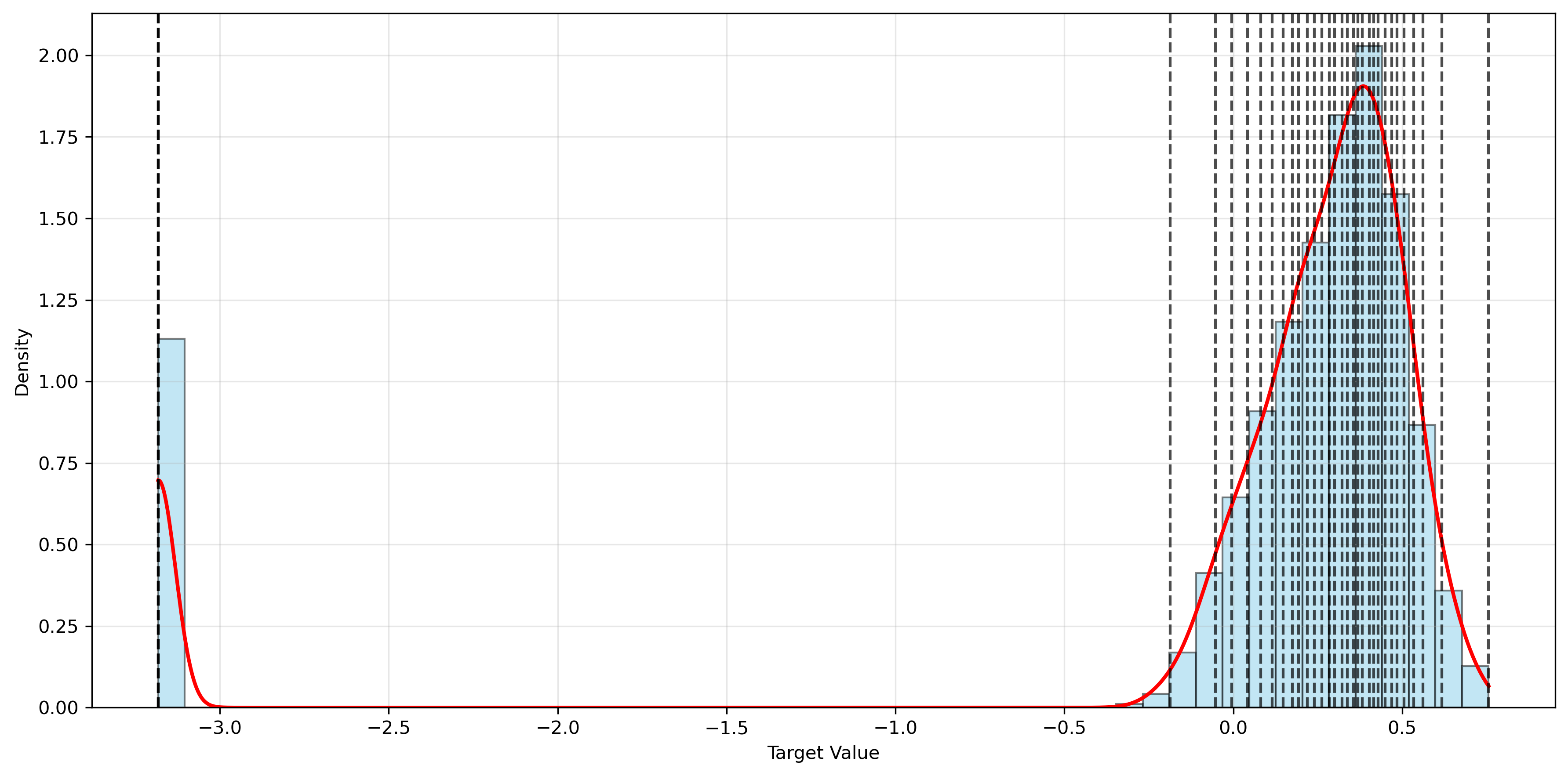}
			\label{fig:binned_airfoild}
		}
		\hfill
		\subfloat[\centering mortgage]{
			\includegraphics[width=0.45\textwidth]{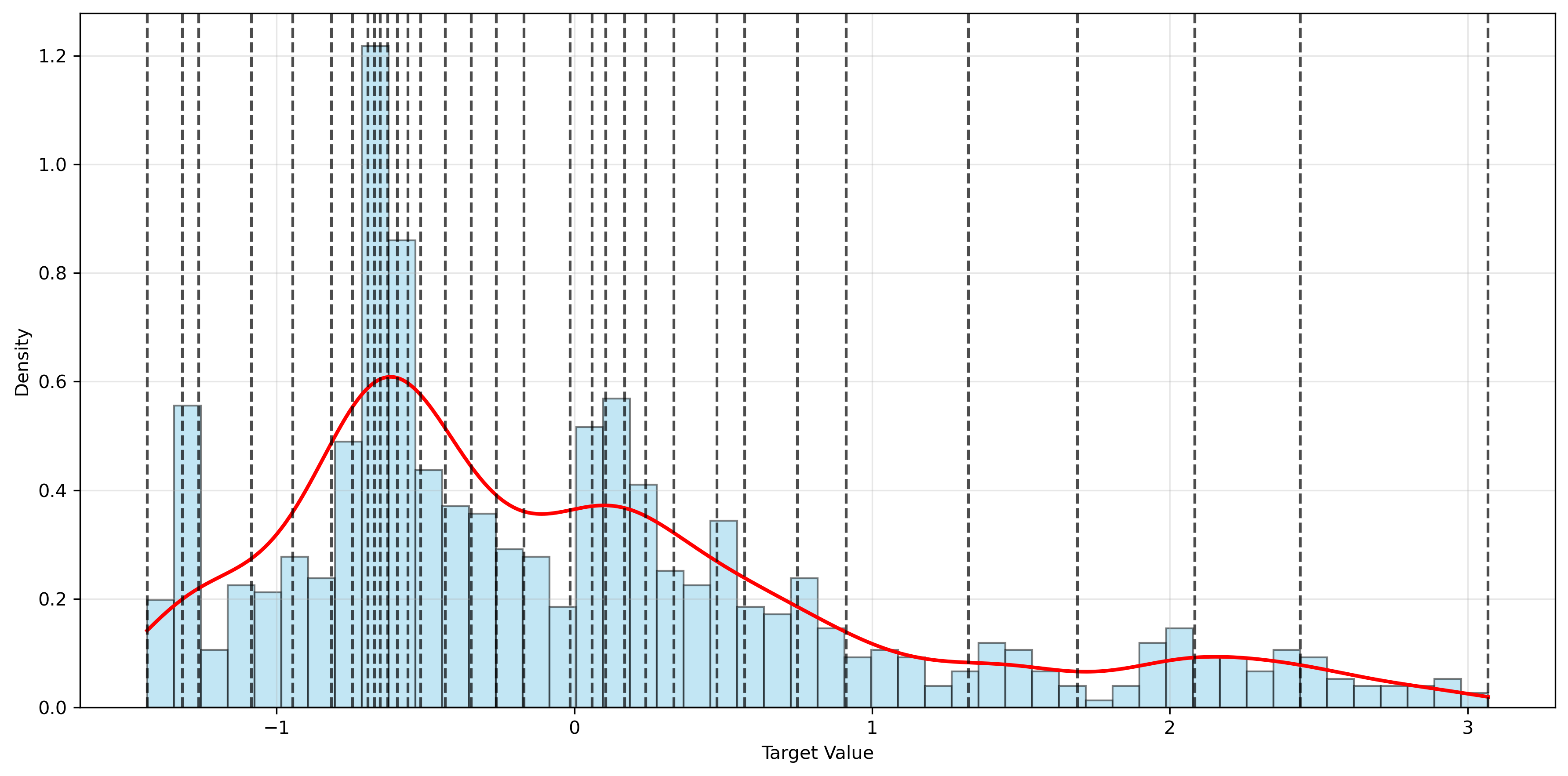}
			\label{fig:binned_mortgage}
		}
		
		\subfloat[\centering treasury]{
			\includegraphics[width=0.45\textwidth]{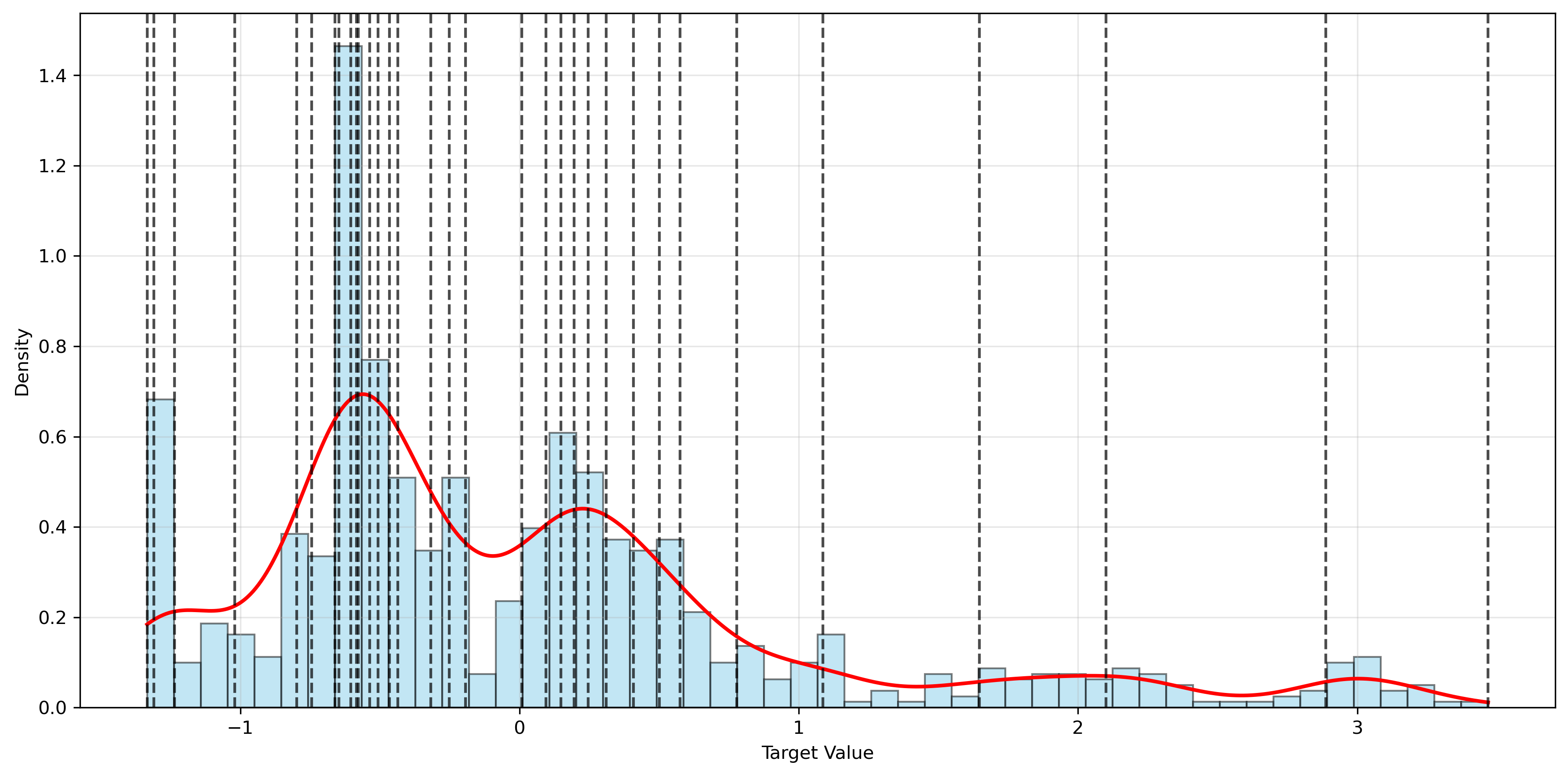}
			\label{fig:binned_treasury}
		}
		\hfill
		\subfloat[\centering concreteStrength]{
			\includegraphics[width=0.45\textwidth]{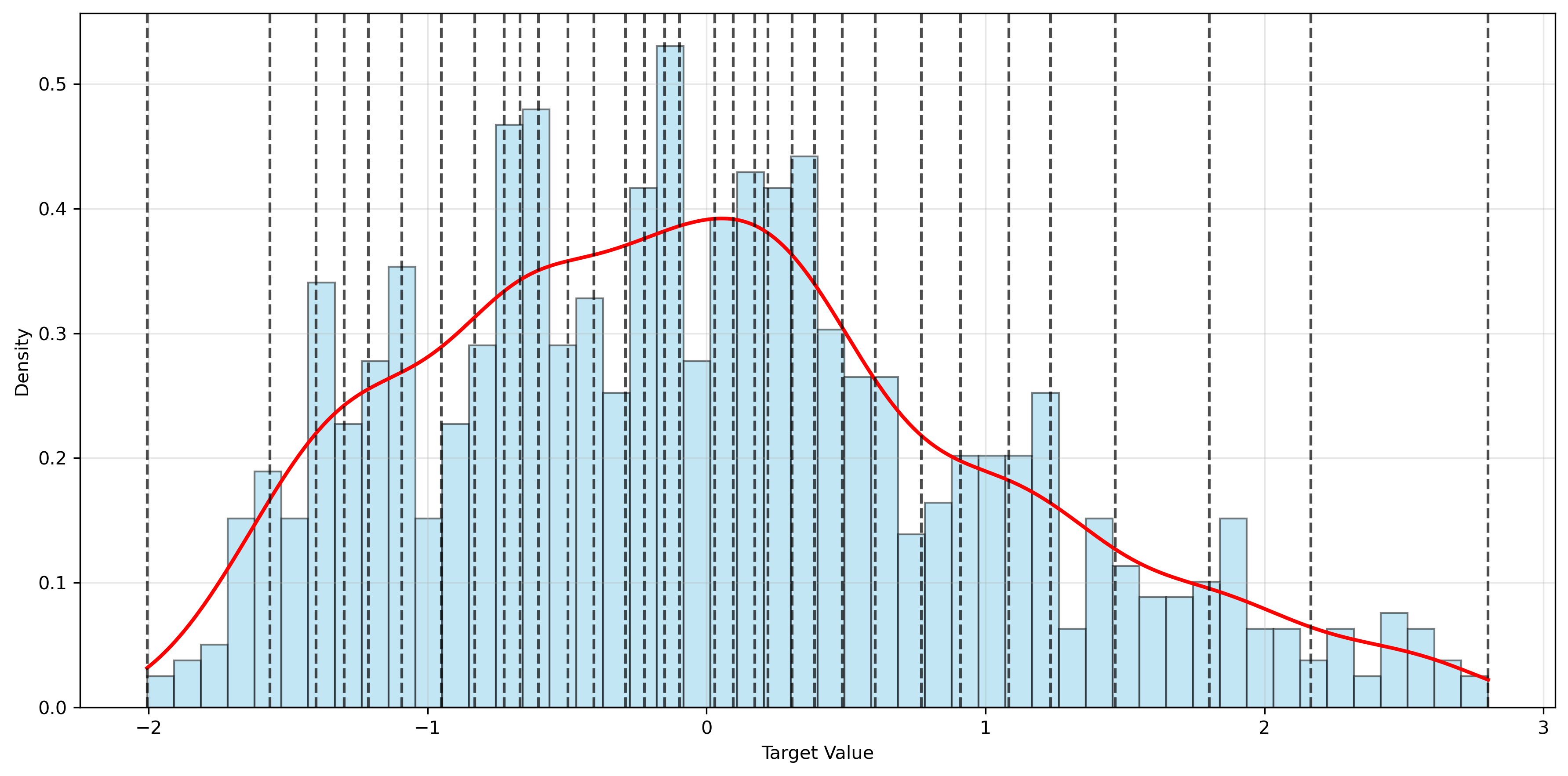}
			\label{fig:binned_concreteStrength}
		}
		
		\caption[]{(continued)}
	\end{figure}

	\begin{table*}[!htbp]
		\centering
		\caption{Statistical summary of the proposed Density-Aware Adaptive Partitioning (DAAP) across all evaluated datasets. Bin sizes denote the number of samples contained within each adaptive partition, while bin widths represent the corresponding target-space coverage of each bin.}
		\label{tab:bin_statistics}
		\resizebox{\textwidth}{!}{
			\begin{tabular}{lcccccccc}
				\hline
				\textbf{Dataset} & \textbf{No. of samples} & \textbf{No. of bins} & \textbf{Min bin size} & \textbf{Max bin size} & \textbf{Avg. bin size} & \textbf{Min bin width} & \textbf{Max bin width} & \textbf{Avg. bin width} \\
				\hline
				
				california        & 20,640 & 61 & 261  & 1,053 & 282.39 & 0.02 & 0.47 & 0.07 \\
				compactive        & 8,192  & 52 & 90   & 675   & 233.85 & 0    & 2.88 & 0.08 \\
				cpu\_small        & 8,192  & 50 & 101  & 664   & 241.28 & 0    & 3.32 & 0.07 \\
				heat              & 7,400  & 64 & 93   & 99    & 93.81  & 0.02 & 3.43 & 0.13 \\
				wine\_quality     & 6,497  & 5  & 161  & 2,239 & 1,239.33 & 0    & 1.15 & 0.38 \\
				abalone           & 4,177  & 24 & 54   & 547   & 269.50 & 0    & 3.41 & 0.23 \\
				space\_ga         & 3,107  & 32 & 78   & 78    & 78     & 0.07 & 10.64 & 0.47 \\
				debutanizer       & 2,394  & 32 & 60   & 60    & 60     & 0.04 & 2.01 & 0.19 \\
				available\_power  & 1,802  & 35 & 29   & 112   & 59.86  & 0.02 & 3.32 & 0.20 \\
				maximal\_torque   & 1,802  & 35 & 28   & 107   & 54.6   & 0    & 4.54 & 0.23 \\
				fuel\_consumption & 1,764  & 53 & 27   & 75    & 45.79  & 0    & 3.61 & 0.14 \\
				acceleration      & 1,732  & 39 & 28   & 68    & 49.74  & 0    & 4.10 & 0.21 \\
				airfoil           & 1,503  & 32 & 38   & 39    & 38.06  & 0    & 3.00 & 0.12 \\
				mortgage          & 1,049  & 32 & 27   & 33    & 27.88  & 0.02 & 0.63 & 0.14 \\
				treasury          & 1,049  & 32 & 27   & 36    & 28.38  & 0.01 & 0.79 & 0.15 \\
				concreteStrength  & 1,030  & 32 & 26   & 29    & 26.44  & 0.04 & 0.64 & 0.15 \\
				
				\hline
			\end{tabular}
		}
	\end{table*}

	\subsubsection{Experimental Results and Analysis}
	This section presents the experimental results and analytical findings of the proposed framework, with the objective of addressing RQ1, RQ2, and RQ3.
	
	\textbf{RQ1:} \textit{How does the proposed data-level imbalanced learning method perform in imbalanced regression tasks compared to (i) standard regression baselines without imbalance handling and (ii) state-of-the-art imbalanced regression methods, both in terms of overall predictive performance and performance within minority target regions?}
	
	To answer the first part of \textbf{RQ1}, we evaluate the proposed DADIR framework by integrating it with a diverse set of standard regression models, including Multi-Layer Perceptron (MLP), XGBoost (XGB), Linear Regression (LR), K-Nearest Neighbors (KNN), Support Vector Regression (SVR), and Ridge Regression. Each model is assessed in two configurations: (i) as a standalone baseline and (ii) when coupled with the proposed data-level imbalanced learning strategy, namely DADIR.
	
	Tables~\ref{tab:overall_mae_dadirVSBaseRegressors}--\ref{tab:overall_r2_dadirVSBaseRegressors} report the \emph{overall performance} (across the entire target space) in terms of MAE, RMSE, and $R^2$ across all datasets for both settings. In addition, the final column summarizes the average percentage improvement achieved by DADIR relative to each corresponding standalone regressor. The results consistently indicate that incorporating DADIR leads to performance gains across all datasets and evaluation metrics. Specifically, every regressor exhibits improved predictive accuracy when combined with the proposed framework, compared to its standalone counterpart. These findings indicate that the proposed framework is regressor-agnostic and consistently enhances predictive performance by mitigating the effects of target imbalance.
	
	The observed improvements can be attributed to the complementary roles of the three phases of the proposed pipeline. The density-aware adaptive partitioning stage exposes the underlying imbalance structure in the target space, while the representation learning stage improves the separability of underrepresented regions in the latent space. Finally, the data-level balancing phase alleviates sample scarcity by generating a more representative and balanced training distribution. Collectively, these phases enable the model to better capture informative variability across the full target range rather than being biased toward dense regions.
	
	This effect is particularly evident for linear models such as Linear Regression and Ridge Regression, which benefit substantially from the enriched representation due to their limited capacity to model complex nonlinear patterns in sparse regions. Nevertheless, nonlinear models such as XGBoost and MLP also exhibit consistent performance gains, indicating that the proposed framework complements rather than replaces their inherent modeling capabilities. Overall, DADIR serves as a regressor-independent enhancement mechanism that improves robustness under imbalanced regression settings.

	\begin{table*}[!htbp]
		\centering
		\scriptsize
		\caption{\textit{Overall MAE} comparison between standalone base regressors (gray columns) and their corresponding DADIR-enhanced variants using identical backbone regressors across all datasets. Lower values indicate better predictive performance ($\downarrow$).}
		\label{tab:overall_mae_dadirVSBaseRegressors}
		\resizebox{\textwidth}{!}{%
			\begin{tabular}{
					l
					>{\columncolor{gray!15}}c >{\columncolor{white}}c
					>{\columncolor{gray!15}}c >{\columncolor{white}}c
					>{\columncolor{gray!15}}c >{\columncolor{white}}c
					>{\columncolor{gray!15}}c >{\columncolor{white}}c
					>{\columncolor{gray!15}}c >{\columncolor{white}}c
					>{\columncolor{gray!15}}c >{\columncolor{white}}c
					c
				}
				\toprule
				\textbf{Dataset} &
				\multicolumn{2}{c}{\textbf{MLP}} &
				\multicolumn{2}{c}{\textbf{XGB}} &
				\multicolumn{2}{c}{\textbf{LR}} &
				\multicolumn{2}{c}{\textbf{KNN}} &
				\multicolumn{2}{c}{\textbf{SVR}} &
				\multicolumn{2}{c}{\textbf{Ridge}} &
				\textbf{Improvement} \\
				\cmidrule(lr){2-3}\cmidrule(lr){4-5}\cmidrule(lr){6-7}
				\cmidrule(lr){8-9}\cmidrule(lr){10-11}\cmidrule(lr){12-13}
				& Standalone & DADIR-MLP 
				& Standalone & DADIR-XGB
				& Standalone & DADIR-LR
				& Standalone & DADIR-KNN
				& Standalone & DADIR-SVR
				& Standalone & DADIR-Ridge
				& \textbf{avg. (\%)} \\
				\midrule
				california        & 0.1418 & 0.0092 & 0.1299 & 0.0248 & 0.2115 & 0.0856 & 0.1735 & 0.0580 & 0.1554 & 0.0340 & 0.2119 & 0.0874 & \textbf{72.90\%} \\
				compactiv          & 0.0364 & 0.0206 & 0.0305 & 0.0300 & 0.1255 & 0.1249 & 0.0395 & 0.0371 & 0.0521 & 0.0508 & 0.1262 & 0.1202 & \textbf{9.81\%} \\
				cpu\_small         & 0.0431 & 0.0158 & 0.0379 & 0.0271 & 0.1247 & 0.1209 & 0.0438 & 0.0366 & 0.0557 & 0.0425 & 0.1254 & 0.1192 & \textbf{23.33\%} \\
				heat              & 0.0068 & 0.0066 & 0.0055 & 0.0052 & 0.0734 & 0.0692 & 0.0082 & 0.0080 & 0.0466 & 0.0364 & 0.0734 & 0.0715 & \textbf{6.84\%} \\
				wine\_quality      & 0.1805 & 0.0130 & 0.1576 & 0.0326 & 0.1886 & 0.0564 & 0.1389 & 0.0075 & 0.1859 & 0.0447 & 0.1886 & 0.0561 & \textbf{80.50\%} \\
				abalone           & 0.1384 & 0.0170 & 0.1197 & 0.0307 & 0.1500 & 0.1233 & 0.1404 & 0.0794 & 0.1393 & 0.0602 & 0.1500 & 0.1233 & \textbf{49.65\%} \\
				space\_ga          & 0.0541 & 0.0224 & 0.0522 & 0.0481 & 0.0675 & 0.0449 & 0.0568 & 0.0521 & 0.0536 & 0.0500 & 0.0678 & 0.0457 & \textbf{24.59\%} \\
				debutanizer       & 0.1156 & 0.0153 & 0.0880 & 0.0311 & 0.1905 & 0.1253 & 0.0819 & 0.0471 & 0.1216 & 0.0440 & 0.1909 & 0.1310 & \textbf{53.89\%} \\
				available\_power   & 0.0372 & 0.0166 & 0.0105 & 0.0102 & 0.0501 & 0.0488 & 0.0069 & 0.0069 & 0.0453 & 0.0442 & 0.0501 & 0.0489 & \textbf{10.94\%} \\
				maximal\_torque    & 0.0203 & 0.0200 & 0.0067 & 0.0066 & 0.0430 & 0.0412 & 0.0230 & 0.0217 & 0.0425 & 0.0419 & 0.0428 & 0.0414 & \textbf{2.92\%} \\
				fuel\_consumption  & 0.0581 & 0.0402 & 0.0382 & 0.0371 & 0.0674 & 0.0652 & 0.0575 & 0.0526 & 0.0615 & 0.0601 & 0.0711 & 0.0702 & \textbf{8.17\%} \\
				acceleration      & 0.0597 & 0.0250 & 0.0336 & 0.0315 & 0.1393 & 0.0637 & 0.0552 & 0.0520 & 0.0639 & 0.0523 & 0.0771 & 0.0650 & \textbf{26.38\%} \\
				airfoild          & 0.0396 & 0.0137 & 0.0216 & 0.0209 & 0.2796 & 0.0529 & 0.0715 & 0.0465 & 0.1562 & 0.0481 & 0.2793 & 0.0595 & \textbf{55.43\%} \\
				mortgage          & 0.0110 & 0.0103 & 0.0102 & 0.0100 & 0.0127 & 0.0118 & 0.0147 & 0.0129 & 0.0435 & 0.0386 & 0.0186 & 0.0886 & \textbf{56.24\%} \\
				treasury          & 0.0212 & 0.0205 & 0.0149 & 0.0123 & 0.0190 & 0.0181 & 0.0145 & 0.0132 & 0.0517 & 0.0531 & 0.0201 & 0.0185 & \textbf{6.62\%} \\
				concreteStrength  & 0.0931 & 0.0184 & 0.0732 & 0.0572 & 0.1930 & 0.0606 & 0.1444 & 0.1000 & 0.1058 & 0.0519 & 0.1937 & 0.0622 & \textbf{53.38\%} \\
				\bottomrule
			\end{tabular}%
		}%
		\parbox{\textwidth}{\raggedleft\itshape Average enhancement: \textbf{33.47\%}}
	\end{table*}

	\begin{table*}[!htbp]
		\centering
		\scriptsize
		\caption{\textit{Overall RMSE} comparison between standalone base regressors (gray columns) and their corresponding DADIR-enhanced variants using identical backbone regressors across all datasets. Lower values indicate better predictive performance ($\downarrow$).}
		\label{tab:overall_rmse_dadirVSBaseRegressors}
		\resizebox{\textwidth}{!}{%
			\begin{tabular}{
					l
					>{\columncolor{gray!15}}c >{\columncolor{white}}c
					>{\columncolor{gray!15}}c >{\columncolor{white}}c
					>{\columncolor{gray!15}}c >{\columncolor{white}}c
					>{\columncolor{gray!15}}c >{\columncolor{white}}c
					>{\columncolor{gray!15}}c >{\columncolor{white}}c
					>{\columncolor{gray!15}}c >{\columncolor{white}}c
					c
				}
				\toprule
				\textbf{Dataset} &
				\multicolumn{2}{c}{\textbf{MLP}} &
				\multicolumn{2}{c}{\textbf{XGB}} &
				\multicolumn{2}{c}{\textbf{LR}} &
				\multicolumn{2}{c}{\textbf{KNN}} &
				\multicolumn{2}{c}{\textbf{SVR}} &
				\multicolumn{2}{c}{\textbf{Ridge}} &
				\textbf{Improvement} \\
				\cmidrule(lr){2-3}\cmidrule(lr){4-5}\cmidrule(lr){6-7}
				\cmidrule(lr){8-9}\cmidrule(lr){10-11}\cmidrule(lr){12-13}
				& Standalone & DADIR-MLP 
				& Standalone & DADIR-XGB
				& Standalone & DADIR-LR
				& Standalone & DADIR-KNN
				& Standalone & DADIR-SVR
				& Standalone & DADIR-Ridge
				& \textbf{avg. (\%)} \\
				\midrule
				california        & 0.2150 & 0.0134 & 0.1950 & 0.0378 & 0.2879 & 0.1185 & 0.2609 & 0.0908 & 0.2308 & 0.0504 & 0.2883 & 0.1207 & \textbf{72.45\%} \\
				compactiv         & 0.0520 & 0.0502 & 0.0428 & 0.0412 & 0.1904 & 0.1895 & 0.0711 & 0.0652 & 0.0663 & 0.0612 & 0.1908 & 0.1799 & \textbf{4.90\%} \\
				cpu\_small         & 0.0599 & 0.0246 & 0.0543 & 0.0515 & 0.1874 & 0.1863 & 0.0633 & 0.0630 & 0.0704 & 0.0631 & 0.1876 & 0.1712 & \textbf{14.04\%} \\
				heat              & 0.0081 & 0.0072 & 0.0078 & 0.0072 & 0.1056 & 0.1001 & 0.0133 & 0.0125 & 0.0541 & 0.0455 & 0.1054 & 0.0990 & \textbf{8.66\%} \\
				wine\_quality      & 0.2355 & 0.0239 & 0.2075 & 0.0806 & 0.2465 & 0.0796 & 0.6487 & 0.0777 & 0.2464 & 0.1193 & 0.2463 & 0.0792 & \textbf{71.03\%} \\
				abalone           & 0.1919 & 0.0317 & 0.1708 & 0.0655 & 0.2080 & 0.1629 & 0.1980 & 0.1897 & 0.1946 & 0.1388 & 0.2080 & 0.1636 & \textbf{36.84\%} \\
				space\_ga          & 0.0707 & 0.0306 & 0.0759 & 0.0738 & 0.0918 & 0.0598 & 0.0828 & 0.0799 & 0.0749 & 0.0697 & 0.0915 & 0.0617 & \textbf{22.89\%} \\
				debutanizer       & 0.1646 & 0.0209 & 0.1395 & 0.0588 & 0.2735 & 0.1730 & 0.1360 & 0.0777 & 0.1845 & 0.0572 & 0.2733 & 0.1762 & \textbf{54.88\%} \\
				available\_power   & 0.0486 & 0.0237 & 0.0371 & 0.0270 & 0.0765 & 0.0752 & 0.0307 & 0.0292 & 0.0605 & 0.0523 & 0.0766 & 0.0685 & \textbf{18.20\%} \\
				maximal\_torque    & 0.0367 & 0.0352 & 0.0256 & 0.0249 & 0.0637 & 0.0629 & 0.0765 & 0.0712 & 0.0549 & 0.0501 & 0.0637 & 0.0574 & \textbf{5.61\%} \\
				fuel\_consumption  & 0.0767 & 0.0715 & 0.0550 & 0.0489 & 0.0923 & 0.0905 & 0.0896 & 0.0792 & 0.0779 & 0.0697 & 0.0950 & 0.0873 & \textbf{8.34\%} \\
				acceleration      & 0.0803 & 0.0538 & 0.0476 & 0.0444 & 0.8269 & 0.0917 & 0.0862 & 0.0810 & 0.0821 & 0.0752 & 0.0998 & 0.0944 & \textbf{24.75\%} \\
				airfoild          & 0.0638 & 0.0187 & 0.0300 & 0.0288 & 0.4773 & 0.0726 & 0.1917 & 0.0640 & 0.3466 & 0.0573 & 0.4772 & 0.0801 & \textbf{65.46\%} \\
				mortgage          & 0.0142 & 0.0119 & 0.0175 & 0.0170 & 0.0188 & 0.0171 & 0.0259 & 0.0219 & 0.0513 & 0.0473 & 0.0264 & 0.0216 & \textbf{11.59\%} \\
				treasury          & 0.0354 & 0.0334 & 0.0328 & 0.0315 & 0.0349 & 0.0328 & 0.0394 & 0.0348 & 0.0610 & 0.0762 & 0.0355 & 0.0292 & \textbf{3.37\%} \\
				concreteStrength  & 0.1401 & 0.0251 & 0.1133 & 0.0922 & 0.2441 & 0.0766 & 0.1998 & 0.1566 & 0.1463 & 0.0666 & 0.2440 & 0.0785 & \textbf{52.21\%} \\
				\bottomrule
			\end{tabular}%
		}%
		\parbox{\textwidth}{\raggedleft\itshape Average enhancement: \textbf{29.69\%}}
	\end{table*}

	\begin{table*}[!htbp]
		\centering
		\scriptsize
		\caption{\textit{Overall $R^2$} comparison between standalone base regressors (gray columns) and their corresponding DADIR-enhanced variants using identical backbone regressors across all datasets. Higher values indicate superior predictive performance ($\uparrow$).}
		\label{tab:overall_r2_dadirVSBaseRegressors}
		\resizebox{\textwidth}{!}{%
			\begin{tabular}{
					l
					>{\columncolor{gray!15}}c >{\columncolor{white}}c
					>{\columncolor{gray!15}}c >{\columncolor{white}}c
					>{\columncolor{gray!15}}c >{\columncolor{white}}c
					>{\columncolor{gray!15}}c >{\columncolor{white}}c
					>{\columncolor{gray!15}}c >{\columncolor{white}}c
					>{\columncolor{gray!15}}c >{\columncolor{white}}c
					c
				}
				\toprule
				\textbf{Dataset} &
				\multicolumn{2}{c}{\textbf{MLP}} &
				\multicolumn{2}{c}{\textbf{XGB}} &
				\multicolumn{2}{c}{\textbf{LR}} &
				\multicolumn{2}{c}{\textbf{KNN}} &
				\multicolumn{2}{c}{\textbf{SVR}} &
				\multicolumn{2}{c}{\textbf{Ridge}} &
				\textbf{Improvement} \\
				\cmidrule(lr){2-3}\cmidrule(lr){4-5}\cmidrule(lr){6-7}
				\cmidrule(lr){8-9}\cmidrule(lr){10-11}\cmidrule(lr){12-13}
				& Standalone & DADIR-MLP 
				& Standalone & DADIR-XGB
				& Standalone & DADIR-LR
				& Standalone & DADIR-KNN
				& Standalone & DADIR-SVR
				& Standalone & DADIR-Ridge
				& \textbf{avg. (\%)} \\
				\midrule
				california        & 0.7981 & 0.9998 & 0.8340 & 0.9986 & 0.6380 & 0.9861 & 0.7028 & 0.9918 & 0.7672 & 0.9975 & 0.6370 & 0.9856 & \textbf{37.57\%} \\
				compactive        & 0.9810 & 0.9976 & 0.9871 & 0.9934 & 0.7452 & 0.8629 & 0.9645 & 0.9689 & 0.9691 & 0.9712 & 0.7443 & 0.8620 & \textbf{5.77\%} \\
				cpu\_small         & 0.9707 & 0.9993 & 0.9758 & 0.9948 & 0.7128 & 0.8110 & 0.9672 & 0.9910 & 0.9595 & 0.9906 & 0.7122 & 0.8101 & \textbf{6.35\%} \\
				heat              & 0.9990 & 0.9999 & 0.9990 & 0.9991 & 0.8276 & 0.9574 & 0.9973 & 0.9991 & 0.9548 & 0.9980 & 0.8276 & 0.9549 & \textbf{5.98\%} \\
				wine\_quality      & 0.3242 & 0.9994 & 0.4751 & 0.9933 & 0.2598 & 0.9935 & 0.4301 & 0.9938 & 0.2603 & 0.9853 & 0.2605 & 0.9935 & \textbf{215.12\%} \\
				abalone           & 0.3330 & 0.9990 & 0.4721 & 0.9959 & 0.2169 & 0.9745 & 0.2902 & 0.9522 & 0.3144 & 0.9815 & 0.2169 & 0.9743 & \textbf{241.62\%} \\
				space\_ga          & 0.7251 & 0.9991 & 0.6839 & 0.9924 & 0.5367 & 0.9966 & 0.6235 & 0.9657 & 0.6919 & 0.9283 & 0.5398 & 0.9964 & \textbf{57.04\%} \\
				debutanizer       & 0.6920 & 0.9995 & 0.7789 & 0.9960 & 0.1497 & 0.9656 & 0.7899 & 0.9931 & 0.6132 & 0.9962 & 0.1509 & 0.9644 & \textbf{207.44\%} \\
				available\_power   & 0.9735 & 0.9995 & 0.9846 & 0.9994 & 0.9343 & 0.9637 & 0.9894 & 0.9950 & 0.9589 & 0.9890 & 0.9342 & 0.9630 & \textbf{2.35\%} \\
				maximal\_torque    & 0.9825 & 0.9955 & 0.9915 & 0.9919 & 0.9474 & 0.9852 & 0.9241 & 0.9761 & 0.9610 & 0.9745 & 0.9474 & 0.9840 & \textbf{2.71\%} \\
				fuel\_consumption  & 0.9091 & 0.9952 & 0.9532 & 0.9925 & 0.8684 & 0.9853 & 0.8759 & 0.9514 & 0.9061 & 0.9681 & 0.8606 & 0.9853 & \textbf{9.50\%} \\
				acceleration      & 0.8883 & 0.9971 & 0.9607 & 0.9965 & 0.0714 & 0.9916 & 0.8713 & 0.9902 & 0.8834 & 0.9880 & 0.8274 & 0.9911 & \textbf{225.01\%} \\
				airfoild          & 0.9829 & 0.9996 & 0.9962 & 0.9978 & 0.0456 & 0.9943 & 0.8461 & 0.9956 & 0.4969 & 0.9964 & 0.0459 & 0.9931 & \textbf{710.69\%} \\
				mortgage          & 0.9990 & 0.9995 & 0.9984 & 0.9984 & 0.9982 & 0.9973 & 0.9966 & 0.9978 & 0.9866 & 0.9977 & 0.9965 & 0.9976 & \textbf{0.22\%} \\
				treasury          & 0.9927 & 0.9989 & 0.9901 & 0.9912 & 0.9929 & 0.9971 & 0.9910 & 0.9940 & 0.9784 & 0.9941 & 0.9827 & 0.9892 & \textbf{0.62\%} \\
				concreteStrength  & 0.8773 & 0.9993 & 0.9197 & 0.9908 & 0.6276 & 0.9937 & 0.7504 & 0.9735 & 0.8661 & 0.9952 & 0.6277 & 0.9933 & \textbf{30.48\%} \\
				\bottomrule
			\end{tabular}%
		}%
		\parbox{\textwidth}{\raggedleft\itshape Average enhancement: \textbf{110.41\%}}
	\end{table*}

	To address the second part of RQ1, we further evaluate the proposed DADIR framework against a set of representative state-of-the-art imbalanced regression methods, including SMOTER, WSMOTER, SMOGN, Label Distribution Smoothing (LDS), Feature Distribution Smoothing (FDS), FOCAL-R, and IRDA (as described in the \hyperref[sec:baselines]{Baseline Methods}). All methods are tested under an identical experimental protocol to ensure a fair and controlled comparison. Following common practice in imbalanced regression studies, MLP is used as the base learner, since it is widely adopted in prior work and provides a consistent reference for assessing the effect of imbalance-handling strategies. Tables~\ref{tab:dadirVSsota_mae_comparison}--\ref{tab:dadirVSsota_r2_comparison} report the results in terms of MAE, RMSE, and $R^2$ across all datasets. Overall, DADIR achieves consistently competitive or superior performance compared to both the standalone MLP and the considered state-of-the-art methods, indicating its effectiveness in standard predictive accuracy under imbalanced target distributions.
	
	Beyond \emph{overall performance}, we further analyze model behavior in \emph{minority target regions}, which are the primary challenge in imbalanced regression. For this purpose, the target space is divided into $B=10$ equal-width bins, and bins with fewer than $0.5 \times (N/B)$ samples are defined as \emph{minority regions}, where $N$ denotes the total number of samples. Tables~\ref{tab:minority_mae_comparison}--\ref{tab:minority_r2_comparison} present MAE, RMSE, and $R^2$ restricted to these minority regions. The results show that DADIR consistently improves prediction quality in rare target ranges, achieving strong and competitive performance against both the baseline MLP and existing state-of-the-art approaches. This confirms that the proposed framework is effective not only in terms of overall accuracy but also in addressing the more challenging underrepresented regions of the target space.
	
	\begin{tableorg}[!htbp]
		\centering
		\caption{\textit{Overall MAE} ($\downarrow$) comparison of the proposed DADIR framework (with MLP) against state-of-the-art imbalanced regression methods across all datasets.}
		\label{tab:dadirVSsota_mae_comparison}
		\resizebox{0.65\textwidth}{!}{
			\begin{tabular}{lcccccccc}
				\toprule
				\textbf{Dataset} & \textbf{DADIR} & \textbf{SMOTER} & \textbf{WSMOTER} & \textbf{SMOGN} & \textbf{LDS} & \textbf{FDS} & \textbf{FOCAL-R} & \textbf{IRDA} \\
				\midrule
				california                & \textbf{0.0092} & 0.3464 & 0.3735 & 0.3667 & 0.3686 & 0.3765 & 0.3471 & 0.6465 \\
				compactiv                 & \textbf{0.0206} & 0.1052 & 0.1052 & 0.1030 & 0.0965 & 0.1358 & 0.1018 & 0.1571 \\
				cpu\_small                & \textbf{0.0158} & 0.1177 & 0.1152 & 0.1147 & 0.1283 & 0.1490 & 0.1264 & 0.2034 \\
				heat                      & \textbf{0.0066} & 0.0354 & 0.0462 & 0.0344 & 0.0809 & 0.1075 & 0.0663 & 0.2235 \\
				wine\_quality             & \textbf{0.0130} & 0.6292 & 0.6347 & 0.6267 & 0.7974 & 0.8594 & 0.6098 & 0.2783 \\
				abalone                   & \textbf{0.0170} & 0.6724 & 0.7193 & 0.6828 & 0.9615 & 0.7192 & 0.6270 & 0.2190 \\
				space\_ga                 & \textbf{0.0224} & 0.3972 & 0.3948 & 0.3998 & 0.5267 & 0.4657 & 0.4207 & 0.1177 \\
				debutanizer               & \textbf{0.0153} & 0.3730 & 0.3960 & 0.4442 & 0.5273 & 0.4908 & 0.4601 & 0.3227 \\
				available\_power          & \textbf{0.0166} & 0.1078 & 0.1128 & 0.1098 & 0.1253 & 0.2169 & 0.1305 & 0.1265 \\
				maximal\_torque           & \textbf{0.0200} & 0.0698 & 0.0671 & 0.0697 & 0.0859 & 0.2689 & 0.0974 & 0.2873 \\
				fuel\_consumption         & \textbf{0.0402} & 0.3800 & 0.2303 & 0.2227 & 0.2630 & 0.4723 & 0.2451 & 0.2874 \\
				acceleration              & \textbf{0.0250} & 0.2521 & 0.2534 & 0.2583 & 0.2627 & 0.3080 & 0.2578 & 0.1661 \\
				airfoild                  & \textbf{0.0137} & 0.3130 & 0.3698 & 0.4013 & 0.5128 & 0.5703 & 0.5672 & 0.3764 \\
				mortgage                  & \textbf{0.0103} & 0.0497 & 0.0532 & 0.0562 & 0.0281 & 0.0450 & 0.0512 & 0.2401 \\
				treasury                  & \textbf{0.0205} & 0.0503 & 0.0513 & 0.0568 & 0.0426 & 0.1088 & 0.0649 & 0.2322 \\
				concreteStrength          & \textbf{0.0184} & 0.2316 & 0.2310 & 0.2703 & 0.2371 & 0.3366 & 0.2859 & 0.4478 \\
				\bottomrule
			\end{tabular}
		}
	\end{tableorg}

	\begin{tableorg}[!htbp]
		\centering
		\caption{\textit{Overall RMSE} ($\downarrow$) comparison of the proposed DADIR framework (with MLP) against state-of-the-art imbalanced regression methods across all datasets.}
		\label{tab:dadirVSsota_rmse_comparison}
		\resizebox{0.65\textwidth}{!}{
			\begin{tabular}{lcccccccc}
				\toprule
				\textbf{Dataset} & \textbf{DADIR} & \textbf{SMOTER} & \textbf{WSMOTER} & \textbf{SMOGN} & \textbf{LDS} & \textbf{FDS} & \textbf{FOCAL-R} & \textbf{IRDA} \\
				\midrule
				california                 & \textbf{0.0134} & 0.4900 & 0.5235 & 0.5198 & 0.5070 & 0.5250 & 0.4843 & 0.7276 \\
				compactiv                  & \textbf{0.0502} & 0.1485 & 0.1508 & 0.1439 & 0.1317 & 0.2154 & 0.1331 & 0.1748 \\
				cpu\_small                 & \textbf{0.0246} & 0.1654 & 0.1618 & 0.1615 & 0.1764 & 0.2281 & 0.1667 & 0.2280 \\
				heat                       & \textbf{0.0072} & 0.0498 & 0.0603 & 0.0469 & 0.1089 & 0.2294 & 0.0933 & 0.2371 \\
				wine\_quality              & \textbf{0.0239} & 0.8109 & 0.8242 & 0.8012 & 1.0133 & 1.1425 & 0.7670 & 0.3495 \\
				abalone                    & \textbf{0.0317} & 0.8733 & 0.9336 & 0.8897 & 1.2228 & 0.9983 & 0.8176 & 0.2888 \\
				space\_ga                  & \textbf{0.0306} & 0.5351 & 0.5269 & 0.5405 & 0.6856 & 0.6348 & 0.5627 & 0.1466 \\
				debutanizer                & \textbf{0.0209} & 0.5131 & 0.5403 & 0.5978 & 0.6920 & 0.7165 & 0.6169 & 0.3858 \\
				available\_power           & \textbf{0.0237} & 0.8432 & 0.1687 & 0.1772 & 0.1818 & 0.3965 & 0.1962 & 0.1595 \\
				maximal\_torque            & \textbf{0.0352} & 0.1102 & 0.1044 & 0.1099 & 0.1252 & 0.5276 & 0.1396 & 0.3337 \\
				fuel\_consumption          & \textbf{0.0715} & 0.3095 & 0.3314 & 0.3318 & 0.3877 & 0.6830 & 0.3262 & 0.3448 \\
				acceleration               & \textbf{0.0538} & 0.3364 & 0.3386 & 0.3446 & 0.3505 & 0.4296 & 0.3380 & 0.1944 \\
				airfoild                   & \textbf{0.0187} & 0.5569 & 0.6692 & 0.7925 & 0.7286 & 0.9285 & 0.7738 & 0.5097 \\
				mortgage                   & \textbf{0.0119} & 0.0733 & 0.0789 & 0.0831 & 0.0404 & 0.0847 & 0.7229 & 0.2665 \\
				treasury                   & \textbf{0.0334} & 0.0825 & 0.0840 & 0.0905 & 0.0661 & 0.1967 & 0.0922 & 0.2558 \\
				concreteStrength           & \textbf{0.0251} & 0.3357 & 0.3260 & 0.3556 & 0.3376 & 0.4424 & 0.3583 & 0.5046 \\
				\bottomrule
			\end{tabular}
		}
	\end{tableorg}

	\begin{tableorg}[!htbp]
		\centering
		\caption{\textit{Overall} $R^2$ ($\uparrow$) comparison of the proposed DADIR framework (with MLP) against state-of-the-art imbalanced regression methods across all datasets.}
		\label{tab:dadirVSsota_r2_comparison}
		\resizebox{0.65\textwidth}{!}{
			\begin{tabular}{lcccccccc}
				\toprule
				\textbf{Dataset} & \textbf{DADIR} & \textbf{SMOTER} & \textbf{WSMOTER} & \textbf{SMOGN} & \textbf{LDS} & \textbf{FDS} & \textbf{FOCAL-R} & \textbf{IRDA} \\
				\midrule
				california                 & \textbf{0.9998} & 0.7632 & 0.7297 & 0.7335 & 0.7465 & 0.7282 & 0.7687 & 0.3129 \\
				compactiv                  & \textbf{0.9976} & 0.9786 & 0.9781 & 0.9800 & 0.9833 & 0.9553 & 0.9829 & 0.7853 \\
				cpu\_small                 & \textbf{0.9993} & 0.9682 & 0.9698 & 0.9697 & 0.9638 & 0.9396 & 0.9677 & 0.5749 \\
				heat                       & \textbf{0.9999} & 0.9867 & 0.9890 & 0.9985 & 0.9802 & 0.9512 & 0.9854 & 0.1310 \\
				wine\_quality              & \textbf{0.9994} & 0.3159 & 0.2932 & 0.3322 & -0.0683 & -0.3579 & 0.3880 & 0.3882 \\
				abalone                    & \textbf{0.9990} & 0.2755 & 0.1719 & 0.2480 & -0.4206 & 0.0532 & 0.3650 & 0.4104 \\
				space\_ga                  & \textbf{0.9991} & 0.7068 & 0.7116 & 0.7038 & 0.5601 & 0.6228 & 0.7036 & 0.1801 \\
				debutanizer                & \textbf{0.9995} & 0.6880 & 0.6540 & 0.5765 & 0.4325 & 0.3916 & 0.5450 & 0.2917 \\
				available\_power           & \textbf{0.9995} & 0.9758 & 0.9770 & 0.9747 & 0.9733 & 0.8732 & 0.9690 & 0.7144 \\
				maximal\_torque            & \textbf{0.9955} & 0.9910 & 0.9919 & 0.9911 & 0.9884 & 0.7939 & 0.9856 & 0.4440 \\
				fuel\_consumption          & \textbf{0.9952} & 0.9117 & 0.8987 & 0.8985 & 0.8614 & 0.5700 & 0.9019 & 0.4369 \\
				acceleration               & \textbf{0.9971} & 0.8866 & 0.8851 & 0.8810 & 0.8768 & 0.8150 & 0.8855 & 0.3457 \\
				airfoild                   & \textbf{0.9996} & 0.6588 & 0.5073 & 0.3091 & 0.4160 & 0.0515 & 0.3412 & 0.2352 \\
				mortgage                   & \textbf{0.9995} & 0.9946 & 0.9937 & 0.9930 & 0.9924 & 0.9928 & 0.9947 & 0.6381 \\
				treasury                   & \textbf{0.9989} & 0.9931 & 0.9929 & 0.9918 & 0.9945 & 0.9610 & 0.9914 & 0.6199 \\
				concreteStrength           & \textbf{0.9993} & 0.8758 & 0.8818 & 0.8606 & 0.8744 & 0.8033 & 0.8586 & 0.5917 \\
				\bottomrule
			\end{tabular}
		}
	\end{tableorg}

	\begin{table*}[!htbp]
		\centering
		\scriptsize
		\caption{\textit{Minority-region MAE} ($\downarrow$) performance comparison of the proposed DADIR framework (with MLP), standalone MLP, and state-of-the-art imbalanced regression approaches across all datasets.}
		\label{tab:minority_mae_comparison}
		\resizebox{0.85\textwidth}{!}{%
			\begin{tabular}{lccccccccc}
				\toprule
				\textbf{Dataset} &
				\textbf{DADIR-MLP} &
				\textbf{Standalone MLP} &
				SMOTER &
				WSMOTER &
				SMOGN &
				LDS &
				FDS &
				FOCAL-R &
				IRDA \\
				\midrule
				
				california        & \textbf{0.0101} & 0.2099 & 0.3997 & 0.3687 & 0.3714 & 0.3888 & 0.5508 & 0.4551 & 0.7498 \\
				compactiv         & \textbf{0.0598} & 0.0611 & 0.1757 & 0.1719 & 0.1582 & 0.1269 & 0.6541 & 0.1514 & 0.1390 \\
				cpu\_small        & \textbf{0.0468} & 0.0691 & 0.1654 & 0.1525 & 0.1532 & 0.1395 & 0.7932 & 0.1667 & 0.1774 \\
				heat              & \textbf{0.0133} & 0.0626 & 0.0551 & 0.0437 & 0.0387 & 0.0664 & 0.9811 & 0.1700 & 0.2472 \\
				wine\_quality     & \textbf{0.0518} & 0.4763 & 1.3506 & 1.3470 & 1.3546 & 1.2074 & 2.1806 & 0.5301 & 0.3205 \\
				abalone           & \textbf{0.0637} & 0.4660 & 1.8837 & 1.8740 & 1.8860 & 1.1898 & 2.9319 & 0.2743 & 0.2283 \\
				space\_ga         & \textbf{0.0422} & 0.1521 & 1.0477 & 0.9446 & 0.9733 & 0.7749 & 1.4504 & 0.2520 & 0.1025 \\
				debutanizer       & \textbf{0.0238} & 0.0632 & 0.8500 & 0.6957 & 0.6718 & 0.6195 & 1.6180 & 0.2605 & 0.3851 \\
				available\_power  & \textbf{0.0187} & 0.0668 & 0.2320 & 0.2079 & 0.2291 & 0.2060 & 1.2735 & 0.3610 & 0.1196 \\
				maximal\_torque   & \textbf{0.0789} & 0.6770 & 0.0886 & 0.0850 & 0.0822 & 0.1513 & 1.5978 & 0.2136 & 0.1068 \\
				fuel\_consumption & \textbf{0.0887} & 0.0914 & 0.3626 & 0.3585 & 0.3706 & 0.3635 & 0.8457 & 0.3857 & 0.1187 \\
				acceleration      & \textbf{0.0189} & 0.1307 & 0.5061 & 0.3822 & 0.3757 & 0.2951 & 1.3199 & 0.4246 & 0.2070 \\
				airfoild          & \textbf{0.0137} & 0.1759 & 0.1292 & 0.1277 & 0.1169 & 0.2758 & 0.1317 & 0.5783 & 0.0911 \\
				mortgage          & \textbf{0.0169} & 0.2454 & 0.1087 & 0.1161 & 0.1261 & 0.0518 & 0.1226 & 0.1237 & 0.2048 \\
				treasury          & \textbf{0.0428} & 0.2612 & 0.1082 & 0.1128 & 0.1219 & 0.1080 & 0.2143 & 0.1669 & 0.1757 \\
				concreteStrength  & \textbf{0.0191} & 0.2099 & 0.1412 & 0.1566 & 0.1394 & 0.1262 & 0.3585 & 0.2836 & 0.3295 \\
				
				\bottomrule
			\end{tabular}%
		}
	\end{table*}

	\begin{table*}[!htbp]
		\centering
		\scriptsize
		\caption{\textit{Minority-region RMSE} ($\downarrow$) performance comparison of the proposed DADIR framework (with MLP), standalone MLP, and state-of-the-art imbalanced regression approaches across all datasets.}
		\label{tab:minority_rmse_comparison}
		\resizebox{0.85\textwidth}{!}{%
			\begin{tabular}{lccccccccc}
				\toprule
				\textbf{Dataset} &
				\textbf{DADIR-MLP} &
				\textbf{Standalone MLP} &
				SMOTER &
				WSMOTER &
				SMOGN &
				LDS &
				FDS &
				FOCAL-R &
				IRDA \\
				\midrule
				
				california        & \textbf{0.0155} & 0.2936 & 0.5815 & 0.5450 & 0.5545 & 0.5637 & 0.7160 & 0.6342 & 0.8183 \\
				compactiv         & \textbf{0.0903} & 0.0911 & 0.2537 & 0.2428 & 0.2247 & 0.1699 & 0.7259 & 0.2110 & 0.1577 \\
				cpu\_small        & \textbf{0.0697} & 0.0950 & 0.2463 & 0.2316 & 0.2321 & 0.2190 & 0.9056 & 0.2405 & 0.2049 \\
				heat              & \textbf{0.0170} & 0.0815 & 0.0941 & 0.0615 & 0.0697 & 0.0899 & 1.1898 & 0.2360 & 0.2797 \\
				wine\_quality     & \textbf{0.0747} & 0.5049 & 1.5182 & 1.5611 & 1.5235 & 1.5344 & 2.2126 & 0.6398 & 0.4079 \\
				abalone           & \textbf{0.0893} & 0.5027 & 2.0210 & 2.0607 & 2.0295 & 1.5250 & 2.9892 & 0.3906 & 0.3007 \\
				space\_ga         & \textbf{0.0635} & 0.1070 & 1.2263 & 1.1215 & 1.1487 & 0.9417 & 1.6233 & 0.3935 & 0.1235 \\
				debutanizer       & \textbf{0.0319} & 0.4231 & 1.0217 & 0.8862 & 0.8413 & 0.7831 & 1.8983 & 0.4908 & 0.4722 \\
				available\_power  & \textbf{0.0347} & 0.1120 & 0.3840 & 0.3385 & 0.3913 & 0.3487 & 1.4629 & 0.4862 & 0.1807 \\
				maximal\_torque   & \textbf{0.0939} & 0.0987 & 0.1194 & 0.1339 & 0.1198 & 0.2145 & 1.8187 & 0.2628 & 0.1257 \\
				fuel\_consumption & \textbf{0.1187} & 0.1290 & 0.5224 & 0.5085 & 0.5883 & 0.4270 & 0.9947 & 0.4721 & 0.1438 \\
				acceleration      & \textbf{0.0268} & 0.1555 & 0.3829 & 0.5042 & 0.5018 & 0.5324 & 1.4270 & 0.5319 & 0.2355 \\
				airfoild          & \textbf{0.0187} & 0.1877 & 0.1656 & 0.1702 & 0.1697 & 0.3512 & 0.1791 & 0.6424 & 0.1108 \\
				mortgage          & \textbf{0.0248} & 0.3052 & 0.1407 & 0.1516 & 0.1605 & 0.0686 & 0.1629 & 0.1625 & 0.2512 \\
				treasury          & \textbf{0.0608} & 0.1068 & 0.1615 & 0.1649 & 0.1744 & 0.1352 & 0.3347 & 0.2289 & 0.2059 \\
				concreteStrength  & \textbf{0.0282} & 0.2936 & 0.1767 & 0.1856 & 0.1738 & 0.1551 & 0.4607 & 0.3532 & 0.3905 \\
				
				\bottomrule
			\end{tabular}%
		}
	\end{table*}

	\begin{table*}[!htbp]
		\centering
		\scriptsize
		\caption{\textit{Minority-region} $R^2$ ($\uparrow$) performance comparison of the proposed DADIR framework (with MLP), standalone MLP, and state-of-the-art imbalanced regression approaches across all datasets.}
		\label{tab:minority_r2_comparison}
		\resizebox{0.85\textwidth}{!}{%
			\begin{tabular}{lccccccccc}
				\toprule
				\textbf{Dataset} &
				\textbf{DADIR-MLP} &
				\textbf{Standalone MLP} &
				SMOTER &
				WSMOTER &
				SMOGN &
				LDS &
				FDS &
				FOCAL-R &
				IRDA \\
				\midrule
				
				california        & \textbf{0.9999} & 0.8212 & 0.8416 & 0.8609 & 0.8560 & 0.8512 & 0.7599 & 0.8116 & 0.3889 \\
				compactiv         & \textbf{0.9915} & 0.9789 & 0.9776 & 0.9795 & 0.9824 & 0.9699 & 0.8164 & 0.9845 & 0.9368 \\
				cpu\_small        & \textbf{0.9983} & 0.9774 & 0.9784 & 0.9809 & 0.9808 & 0.9829 & 0.7080 & 0.9794 & 0.8947 \\
				heat              & \textbf{0.9996} & 0.8620 & 0.9890 & 0.9953 & 0.9939 & 0.9495 & 0.1242 & 0.9307 & 0.0243 \\
				wine\_quality     & \textbf{0.9990} & 0.4656 & 0.5875 & 0.5638 & 0.5846 & 0.0579 & 0.1238 & 0.5188 & 0.6513 \\
				abalone           & \textbf{0.9949} & 0.0569 & 0.0735 & 0.0218 & 0.0376 & 0.0099 & 0.0042 & 0.4330 & 0.2662 \\
				space\_ga         & \textbf{0.9996} & 0.0944 & 0.0641 & 0.0371 & 0.0419 & 0.0319 & 0.0767 & 0.1201 & 0.0231 \\
				debutanizer       & \textbf{0.9987} & 0.2439 & 0.0652 & 0.0261 & 0.0752 & 0.1987 & 0.0082 & 0.1241 & 0.1793 \\
				available\_power  & \textbf{0.9985} & 0.7956 & 0.8273 & 0.8658 & 0.8208 & 0.8577 & 0.1057 & 0.7232 & 0.4684 \\
				maximal\_torque   & \textbf{0.9892} & 0.8171 & 0.9847 & 0.9808 & 0.9846 & 0.9507 & 0.0037 & 0.9260 & 0.7035 \\
				fuel\_consumption & \textbf{0.9868} & 0.8698 & 0.8726 & 0.8793 & 0.8384 & 0.8633 & 0.5380 & 0.8959 & 0.8380 \\
				acceleration      & \textbf{0.9975} & 0.0277 & 0.1245 & 0.1309 & 0.1391 & 0.3768 & 0.0782 & 0.0329 & 0.2743 \\
				airfoild          & \textbf{0.9996} & 0.0567 & 0.0187 & 0.0141 & 0.0014 & 0.0034 & 0.0036 & 0.0089 & 0.8535 \\
				mortgage          & \textbf{0.9982} & 0.9561 & 0.9432 & 0.9341 & 0.9260 & 0.9665 & 0.9238 & 0.9242 & 0.0832 \\
				treasury          & \textbf{0.9942} & 0.8803 & 0.9525 & 0.9505 & 0.9447 & 0.9458 & 0.7963 & 0.9048 & 0.5550 \\
				concreteStrength  & \textbf{0.9998} & 0.8212 & 0.9913 & 0.9904 & 0.9916 & 0.9933 & 0.9410 & 0.9653 & 0.7598 \\
				
				\bottomrule
			\end{tabular}%
		}
	\end{table*}

	\textbf{RQ2:} \textit{How do the degree and characteristics of target variable skewness across datasets affect the effectiveness and robustness of the proposed data-level imbalanced learning method?}
	
	In regression tasks, a positively or negatively skewed target distribution implies that most observations are concentrated within a narrow region of the target space, whereas only a limited number of samples appear in the tail regions. Consequently, larger absolute skewness values indicate a more pronounced imbalance between densely populated regions and sparsely represented extreme values. As the distribution tails become longer, the scarcity of rare target values increases, making the prediction of extreme samples considerably more challenging. Table \ref{tab:skewness_report} reports the skewness values of the target variables for the benchmark datasets considered in this study, where skewness was estimated using \emph{Pearson’s second coefficient} of skewness.
		
\begin{figure*}[htbp]
		\centering
		\begin{tikzpicture}
			
			\begin{axis}[
				name=plot1,
				width=\textwidth,
				height=5.5cm,
				title={\small\textbf{(a) MAE improvement vs. skewness}},
				xlabel={Absolute Skewness $|\gamma|$},
				ylabel={Avg. \% of Improvement in MAE},
				xmin=0, xmax=3.6,
				ymin=0, ymax=90,
				xtick={0,1,2,3},
				ytick={0,15,30,45,60,75,90},
				grid=major,
				legend style={font=\scriptsize, draw=none, at={(0.01,0.99)}, anchor=north west},
				label style={font=\scriptsize},
				tick label style={font=\scriptsize},
				title style={font=\small}
				]
				\addplot[only marks, mark=*, mark size=2.5pt, color=blue!70!black] coordinates {
					(0.9777,72.90) (3.4161,9.81) (3.4161,23.33) (1.6340,6.84)
					(0.2000,80.50) (1.1137,49.65) (1.0000,24.59) (1.7088,53.89)
					(1.9036,10.94) (1.6338,2.92) (1.1378,8.17) (0.7694,26.38)
					(2.7545,55.43) (1.0282,56.24) (0.3000,6.62) (1.0000,53.38)
				};
				\addlegendentry{MAE}
				
				\addplot[thick, dashed, color=blue!70!black] coordinates {(0,45.27) (3.5,18.61)};
				\addlegendentry{Trend ($\rho=-0.29$)}
				
			\end{axis}
			
			\begin{axis}[
				anchor=north west,
				at={(plot1.south west)},
				yshift=-2.5cm,
				name=plot2,
				width=\textwidth,
				height=5.5cm,
				title={\small\textbf{(b) RMSE improvement vs. skewness}},
				xlabel={Absolute Skewness $|\gamma|$},
				ylabel={Avg. \% of Improvement in RMSE},
				xmin=0, xmax=3.6,
				ymin=0, ymax=80,
				xtick={0,1,2,3},
				ytick={0,10,20,30,40,50,60,70,80},
				grid=major,
				legend style={font=\scriptsize, draw=none, at={(0.01,0.99)}, anchor=north west},
				label style={font=\scriptsize},
				tick label style={font=\scriptsize},
				title style={font=\small}
				]
				\addplot[only marks, mark=*, mark size=2.5pt, color=green!70!black] coordinates {
					(0.9777,72.45) (3.4161,4.90) (3.4161,14.04) (1.6340,8.66)
					(0.2000,71.03) (1.1137,36.84) (1.0000,22.89) (1.7088,54.88)
					(1.9036,18.20) (1.6338,5.61) (1.1378,8.34) (0.7694,24.75)
					(2.7545,65.46) (1.0282,11.59) (0.3000,3.37) (1.0000,52.21)
				};
				\addlegendentry{RMSE}
				
				\addplot[thick, dashed, color=green!70!black] coordinates {(0,37.49) (3.5,19.31)};
				\addlegendentry{Trend ($\rho=-0.24$)}
				
			\end{axis}
			
			\begin{axis}[
				anchor=north west,
				at={(plot2.south west)},
				yshift=-2.5cm,
				name=plot3,
				width=\textwidth,
				height=5.5cm,
				title={\small\textbf{(c) $R^2$ improvement vs. skewness}},
				xlabel={Absolute Skewness $|\gamma|$},
				ylabel={Avg. \% of Improvement in $R^2$},
				xmin=0, xmax=3.6,
				ymin=0, ymax=750,
				xtick={0,1,2,3},
				ytick={0,150,300,450,600,750},
				grid=major,
				legend style={font=\scriptsize, draw=none, at={(0.01,0.99)}, anchor=north west},
				label style={font=\scriptsize},
				tick label style={font=\scriptsize},
				title style={font=\small}
				]
				\addplot[only marks, mark=*, mark size=2.5pt, color=orange!80!black] coordinates {
					(0.9777,37.57) (3.4161,5.77) (3.4161,6.35) (1.6340,5.98)
					(0.2000,215.12) (1.1137,241.62) (1.0000,57.04) (1.7088,207.44)
					(1.9036,2.35) (1.6338,2.71) (1.1378,9.50) (0.7694,225.01)
					(2.7545,710.69) (1.0282,0.22) (0.3000,0.62) (1.0000,30.48)
				};
				\addlegendentry{$R^2$}
				
				\addplot[thick, dashed, color=orange!80!black] coordinates {(0,72.05) (3.5,160.39)};
				\addlegendentry{Trend ($\rho=-0.14$)}
				
			\end{axis}
			
		\end{tikzpicture}
		\caption{Impact of target variable skewness on DADIR performance across the 16 benchmark datasets. The points represent the average percentage improvement obtained for each dataset, while the dashed lines provide illustrative linear guides to the overall tendency. Spearman’s rank correlations between absolute skewness $|\gamma|$ and percentage improvement are: MAE ($\rho=-0.287$, $p=0.281$), RMSE ($\rho=-0.244$, $p=0.361$), and $R^2$ ($\rho=-0.144$, $p=0.594$).}
		\label{fig:skewness_3panel}
	\end{figure*}
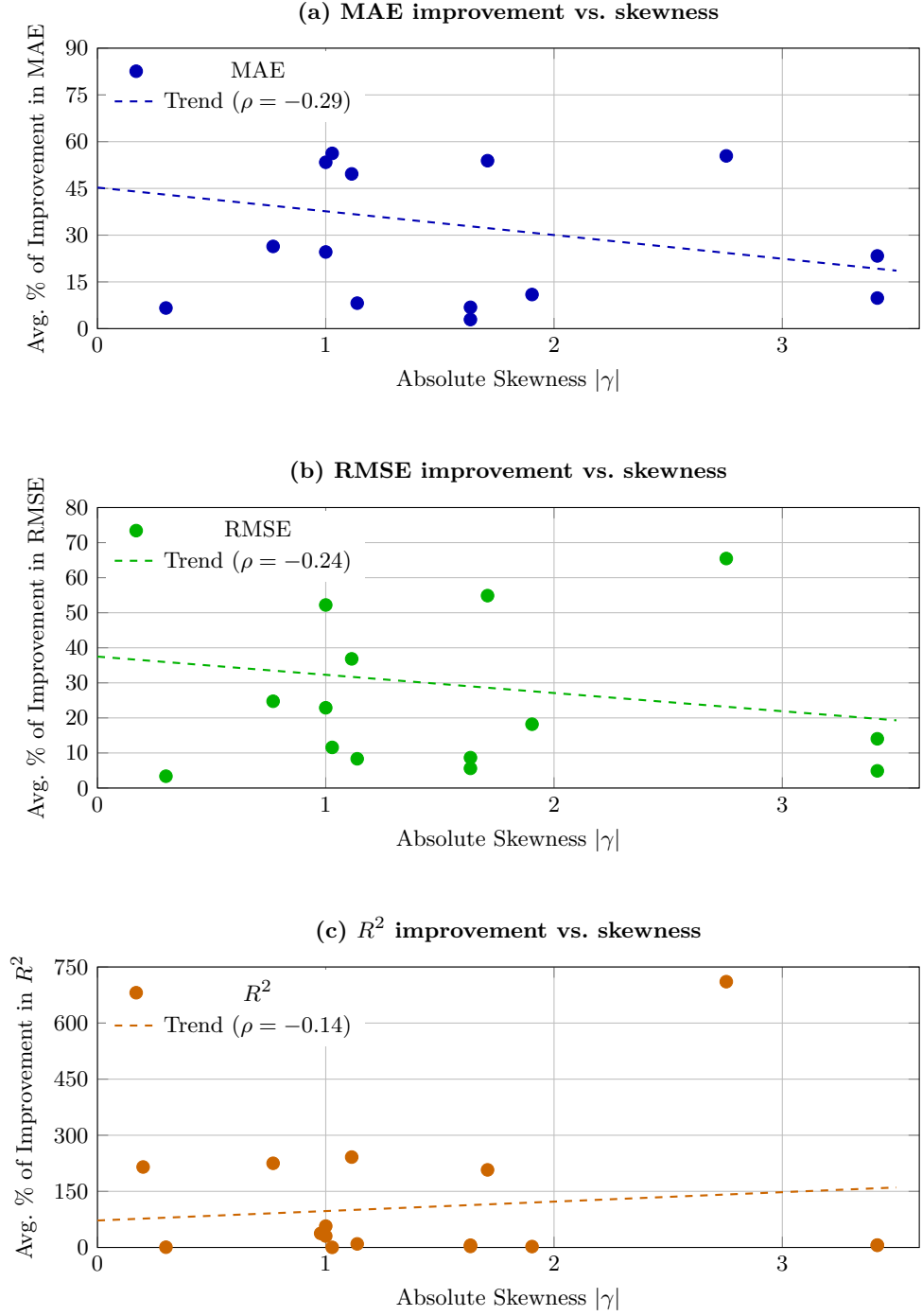

	To address RQ2, we computed the \textit{Spearman rank correlation coefficient} ($\rho$) between the absolute skewness values $|\gamma|$ and the average percentage improvement across the three evaluation metrics. Fig.~\ref{fig:skewness_3panel} illustrates these relationships. The analysis does not indicate a statistically significant monotonic relationship between target skewness and performance improvement. The obtained correlations are $\rho=-0.287$ ($p=0.281$) for MAE, $\rho=-0.244$ ($p=0.361$) for RMSE, and $\rho=-0.144$ ($p=0.594$) for $R^2$. Although weak negative trends are observed, the relatively high $p$-values indicate that skewness alone cannot reliably explain the variation in performance gains across datasets.
	
	The results suggest that the effectiveness of the proposed DADIR framework is not determined solely by the degree of target asymmetry. Several datasets with relatively low skewness still achieve substantial improvements. For example, \textit{wine\_quality} ($|\gamma| \approx 0.2$) achieves 80.50\% and 71.03\% improvement in MAE and RMSE, respectively, together with a 215.12\% gain in $R^2$. Likewise, datasets such as \textit{abalone} and \textit{acceleration} exhibit considerable performance improvements despite moderate skewness levels. In contrast, some highly skewed datasets, including \textit{compactiv} ($|\gamma| \approx 3.42$), show comparatively smaller gains. These observations indicate that additional factors, such as local sample density, rarity structure, and target-space complexity, also play an important role in determining learning difficulty in imbalanced regression tasks. Nevertheless, highly skewed datasets can still benefit significantly from the proposed framework. For instance, \textit{airfoild} ($|\gamma| \approx 2.75$) achieves a remarkable 710.69\% improvement in $R^2$, demonstrating the capability of DADIR to effectively handle severe imbalance conditions. Overall, the results suggest that skewness should be viewed as a descriptive indicator of target imbalance rather than a standalone predictor of performance improvement. The proposed framework maintains robust and consistent effectiveness across datasets with diverse distributional characteristics.
	
	Overall, the observed positive monotonic relationship verifies that target skewness can serve as a reliable indicator of imbalance severity in regression problems. As the asymmetry of the target distribution increases, the proposed framework demonstrates progressively greater advantages over conventional standalone regressors. This behavior can be attributed to the adaptive binning strategy and cost-sensitive learning mechanism employed by the framework, which place greater emphasis on underrepresented target regions. Therefore, the proposed method is particularly effective for highly skewed regression datasets, where its robustness and performance advantages become increasingly evident across diverse real-world scenarios.

	\begin{tableorg}[htbp]
		\centering
		\scriptsize
		\caption{Skewness ($\gamma$) of target variable and performance improvement of the proposed framework (DADIR) over 16 benchmark datasets.}
		\label{tab:skewness_report}
		\resizebox{0.6\textwidth}{!}{
			\begin{tabular}{l c c c c}
				\toprule
				\textbf{Dataset} & \textbf{Skewness $(\gamma)$} & \multicolumn{3}{c}{\textbf{Improvement avg. \%}} \\
				\cmidrule(lr){3-5}
				&  & \textbf{MAE $\downarrow$} & \textbf{RMSE $\downarrow$} & \textbf{$R^2$ $\uparrow$} \\
				\midrule
				california & 0.9777 & 72.90 & 72.45 & 37.57 \\
				compactiv & -3.4161 & 9.81 & 4.90 & 5.77 \\
				cpu\_small & -3.4161 & 23.33 & 14.04 & 6.35 \\
				heat & 1.6340 & 6.84 & 8.66 & 5.98 \\
				wine\_quality & 0.2 & 80.50 & 71.03 & 215.12 \\
				abalone & 1.1137 & 49.65 & 36.84 & 241.62 \\
				space\_ga & $\approx$1.0 & 24.59 & 22.89 & 57.04 \\
				debutanizer & 1.7088 & 53.89 & 54.88 & 207.44 \\
				available\_power & 1.9036 & 10.94 & 18.20 & 2.35 \\
				maximal\_torque & 1.6338 & 2.92 & 5.61 & 2.71 \\
				fuel\_consumption & 1.1378 & 8.17 & 8.34 & 9.50 \\
				acceleration & 0.7694 & 26.38 & 24.75 & 225.01 \\
				airfoild & -2.7545 & 55.43 & 65.46 & 710.69 \\
				mortgage & 1.0282 & 56.24 & 11.59 & 0.22 \\
				treasury & $\approx$0.3 & 6.62 & 3.37 & 0.62 \\
				concreteStrength & $\approx$1.0 & 53.38 & 52.21 & 30.48 \\
				\bottomrule
			\end{tabular}
		}
	\end{tableorg}
	
	\textbf{RQ3:} \textit{What are the limitations of the proposed data-level imbalanced learning method?}
		
	Although the proposed framework demonstrates strong performance across diverse imbalanced regression benchmarks, several limitations should be acknowledged. First, the effectiveness of the DAAP phase depends on the quality of kernel density estimation. In datasets with extremely sparse target regions or highly irregular target distributions, density valleys may become less distinct, potentially affecting the quality of the adaptive partitioning process. Second, the DR-CVAE phase assumes that a meaningful latent representation can be learned from the available data. As with most deep generative models, representation quality may deteriorate when training data are severely limited or when feature spaces become substantially more complex than those considered in this study (5--37 features).
	
	In addition, the latent-space balancing strategy in Phase III relies on clustering and density-guided sample generation. Consequently, the reliability of synthetic samples may decrease in bins containing very few observations, where local structures are difficult to estimate accurately. Finally, the proposed framework introduces additional computational overhead due to density estimation, representation learning, clustering, and synthetic sample generation. While the cost remains practical for the benchmark datasets evaluated, scalability to very large-scale or high-dimensional regression problems warrants further investigation.
	
	Overall, these limitations do not diminish the effectiveness of the proposed method but highlight promising directions for future research, particularly toward improving robustness in extremely sparse regions and extending the framework to larger and more complex data domains.


	\subsection{Ablation Study}
	\label{sec:ablation}
	\textbf{RQ4:} \emph{How does each phase of the proposed framework (Density-Aware Adaptive Partitioning (DAAP), Density-Regularized Conditional Variational Autoencoder (DR-CVAE), Data-Level Balancing via Latent-Space Modeling) contribute to the overall performance in imbalanced regression tasks?}
	
	To address this research question, we conduct a systematic phase-wise analysis of the proposed framework. The objective is to assess the contribution of each phase by selectively replacing or removing it while preserving the remaining components. Specifically, each modified variant differs from the complete framework in only one phase, enabling the impact of that phase to be examined in isolation.
	
	The following ablation configurations are considered. First, the proposed Density-Aware Adaptive Partitioning (DAAP) strategy is replaced with a conventional quantile-based partitioning scheme ($q=10$). Second, the latent representation learning phase based on DR-CVAE is omitted, and all subsequent operations are performed directly in the original feature space. Third, the proposed minority-aware oversampling mechanism is substituted with SMOGN, a widely adopted data-level method for imbalanced regression. All resulting variants are evaluated using the same benchmark datasets, regressors, and experimental settings as the full framework.
	
	To quantify the effect of each modification, we compute the average relative performance variation with respect to the complete framework across all datasets, regressors, and evaluation metrics (MAE, RMSE, and $R^2$). The relative variation is calculated as Eq. \ref{eq:relvar}:

	\begin{equation}
		\text{Relative variation (\%)} = \frac{\text{Metric}_{\text{ablation}} - \text{Metric}_{\text{full}}}{\left| \text{Metric}_{\text{full}} \right|} \times 100,
		\label{eq:relvar}
	\end{equation}

	A larger performance degradation indicates a greater contribution of the corresponding phase to the overall effectiveness of the proposed framework. Table \ref{tab:ablation-main} presents the average relative performance variations associated with each ablation setting. For error-based metrics (MAE and RMSE), larger positive $\Delta$ values indicate greater performance loss, while for $R^2$, more negative $\Delta$ values indicate a stronger degradation. Because only one phase is modified at a time, these variations directly reflect the importance of the corresponding component within the proposed framework. Results are averaged across all benchmark datasets and regressors, with detailed per-dataset analyses provided in Appendix~\ref{appen:ablation-study}.
	
	The ablation results confirm that all three phases contribute meaningfully to the effectiveness of the proposed framework, while exhibiting different levels of influence on the final predictive performance. In particular, the removal of the Density-Regularized Conditional Variational Autoencoder phase (DR-CVAE) leads to the most substantial degradation across all evaluation metrics, highlighting the critical role of latent representation learning in modeling the complex structure of imbalanced regression data. The exclusion of the Density-Aware Adaptive Partitioning (DAAP) phase also results in considerable performance deterioration, demonstrating the importance of adaptive target-space partitioning for accurately identifying informative minority regions. Furthermore, replacing the proposed data-level balancing mechanism with SMOGN consistently reduces predictive performance relative to the complete framework, indicating that the proposed oversampling strategy provides more effective minority-region enrichment than existing data-level approaches. Overall, the results presented in Table~\ref{tab:ablation-main} demonstrate that the performance gains achieved by the proposed framework arise from the complementary interaction of all three phases rather than from a single isolated component.

	\begin{tableorg}[htbp]
		\centering
		\caption{Ablation study: average relative change (\%) under different ablation settings with respect to the complete framework. Positive $\Delta$ values indicate deterioration in MAE and RMSE, whereas negative $\Delta$ values indicate deterioration in $R^2$. Detailed results are reported in Appendix~\ref{appen:ablation-study}.}
		\label{tab:ablation-main}
		\resizebox{\textwidth}{!}{
			
			\begin{tabular}{lccc}
				\toprule
				\toprule
				\textbf{Phase Removed} & \textbf{Avg. $\Delta$ MAE (\%)} & \textbf{Avg. $\Delta$ RMSE (\%)} & \textbf{Avg. $\Delta$ R$^2$ (\%)} \\
				\midrule
				\midrule
				Without Density-Aware Adaptive Partitioning (DAAP)- Phase I & 66.57 & 58.85 & -1.15 \\
				Without Density-Regularized Conditional Variational Autoencoder (DR-CVAE)- Phase II   & 655.64 & 563.46 & -23.63 \\
				Without proposed data-level balancing method- Phase III        & 51.99  & 83.02  & -0.86 \\

				\bottomrule
			\end{tabular}
		}
	\end{tableorg}

	\subsection{Execution Time Analysis}
	\label{sec:executionTime}
	To evaluate the balance between predictive effectiveness and computational overhead, wall-clock training times are reported for both the proposed framework and a simpler baseline under the same experimental setup. Runtime analysis is performed using an MLP regressor as a representative iterative learning model, since it constitutes the most computationally intensive regressor among those considered in this study. All experiments are conducted on identical hardware equipped with an Intel Core i5-2500K CPU, 16 GB RAM, and an NVIDIA GeForce GT 710 GPU (2 GB), while execution is performed exclusively on the CPU to ensure consistent and fair runtime comparisons.
	
	As reported in Table~\ref{tab:runtime}, the proposed framework introduces additional training‑time overhead across the evaluated imbalanced datasets of different scales. On average, DADIR+MLP increases training time by 107.01\% compared with a standalone MLP model. This overhead is mainly attributable to the representation learning (phase 2) and data‑level balancing (phase 3), which constitute the core components of the proposed methodology. The overhead varies across datasets, from as low as 22.19\% (space\_ga) to as high as 336.86\% (cpu\_small). However, it does not exhibit disproportionate escalation for larger‑scale problems, and the observed values remain predictable and consistent with the theoretical computational complexity analysis discussed in Section~\ref{sec:computational_complexity}. Even the highest overhead is compensated by the substantial and consistent predictive performance gains that DADIR achieves over simpler baseline regressors. Taken together, these findings suggest that the overhead introduced by DADIR is both manageable and predictable.

	\begin{table*}[htbp]
		\centering
		\caption{Wall-clock execution time (in minutes) for the proposed data-level framework (DADIR) combined with MLP compared to a standalone MLP across multiple datasets.}
		\label{tab:runtime}
		\resizebox{\textwidth}{!}{
			\begin{tabular}{lcccccccccccccccc}
				\toprule
				\textbf{Method} &
				\textbf{california} &
				\textbf{compactiv} &
				\textbf{cpu\_small} &
				\textbf{heat} &
				\textbf{wine} &
				\textbf{abalone} &
				\textbf{space\_ga} &
				\textbf{debutanizer} &
				\textbf{avail.\_power} &
				\textbf{max.\_torque} &
				\textbf{fuel\_cons.} &
				\textbf{acceleration} &
				\textbf{airfoil} &
				\textbf{mortgage} &
				\textbf{treasury} &
				\textbf{concrete} \\
				\midrule
				Standalone MLP &
				20.24 & 7.95 & 7.76 & 7.31 & 6.14 & 4.00 & 3.02 & 2.35 & 1.82 & 1.87 & 1.78 & 1.70 & 1.51 & 1.25 & 1.04 & 1.07 \\
				
				DADIR + MLP &
				53.54 & 32.75 & 33.90 & 9.16 & 13.44 & 12.88 & 3.69 & 3.44 & 2.77 & 2.35 & 2.59 & 3.40 & 2.11 & 1.92 & 1.96 & 1.70 \\
				
				\midrule
				Overhead (\%) &
				164.53 & 311.95 & 336.86 & 25.31 & 118.89 & 222.00 & 22.19 & 46.38 & 52.20 & 25.67 & 45.51 & 100.00 & 39.74 & 53.60 & 88.46 & 58.88 \\
				\multicolumn{17}{r}{\textbf{Avg. overhead:} \textbf{107.01\%}} \\
				\bottomrule
			\end{tabular}
		}
	\end{table*}

	\section{Conclusion}
	\label{sec:conclusion}
	This study addressed the persistent challenge of imbalanced regression, where non-uniform target distributions bias learning models toward densely populated regions and reduce their reliability on rare yet practically important target values. To address this issue, we proposed \emph{DADIR}, a \emph{Density-Aware Data-Level Imbalanced Regression} framework that unifies adaptive target-space partitioning, density-regularized representation learning, and structure-preserving latent-space oversampling within a single preprocessing pipeline. Specifically, \emph{Density-Aware Adaptive Partitioning (DAAP)} recursively segments the target space using kernel density estimation to capture intrinsic density variations; \emph{Density-Regularized Conditional Variational Autoencoder (DR-CVAE)} learns target-conditional latent representations while emphasizing samples from sparse target regions through inverse-density weighting; and progressive cluster-based oversampling generates synthetic instances in the latent space by preserving local structural patterns and maintaining target consistency through nearest-neighbor interpolation.
	
	Extensive experiments across diverse imbalanced regression benchmark datasets demonstrate that DADIR improves predictive performance when applied upstream of multiple standard regressors. The reductions in MAE and RMSE, together with improvements in ($R^2$), confirm that the proposed density-aware data-level strategy enhances both overall predictive accuracy and minority-region performance. Compared with standalone regressors and existing imbalance-handling methods, DADIR provides a more effective balancing mechanism by jointly exploiting target-density information, latent representation learning, and local feature-space structure. Moreover, the ablation studies validate the necessity of each phase: DAAP improves the identification of underrepresented target regions, DR-CVAE preserves sparse-region information in the learned latent space, and latent-space oversampling improves the structural consistency of generated samples. These results demonstrate that the effectiveness of DADIR arises from the coordinated interaction of its density-aware components rather than from oversampling alone.
	
	Overall, DADIR offers a flexible and model-agnostic data-level solution for imbalanced regression. Since it produces a balanced training set that can be directly used by standard regressors, it improves robustness under target imbalance without requiring modifications to model architectures or learning objectives. Future work may extend DADIR toward more scalable density estimation, stronger generative mechanisms for extremely sparse regions, and efficient latent encoders for high-dimensional or streaming regression scenarios. More broadly, this study highlights density-awareness as a central design principle for data-level learning under continuous target imbalance.

	\section*{Statements and Declarations}
	
	\noindent\textbf{Competing Interests:} The authors declare that they have no competing interests.
	
	\noindent\textbf{Funding:} No funding was received for conducting this study.
	
	
	
	
	
	\noindent\textbf{Availability of Data and Materials:} All datasets used in this study are publicly available on the internet.

	\appendix
	
	\section{Detailed Ablation Study Results}
	\label{appen:ablation-study}
	Tables \ref{tab:detailed_Ablation_Ph1} to \ref{tab:detailed_Ablation_Ph3} provide a comprehensive phase-wise ablation analysis of the proposed density-aware data-level imbelanced leraning framework using three evaluation metrics: MAE, RMSE, and $R^2$. Across all datasets and regressors, the removal of any individual phase leads to a noticeable decline in predictive performance. In particular, the ablated variants generally exhibit higher MAE and RMSE values together with lower $R^2$ scores, demonstrating the positive contribution of each component to the overall framework. While the extent of performance degradation differs depending on the dataset characteristics and the employed regressor, the complete framework consistently achieves superior results compared to all corresponding ablated versions, further validating the effectiveness and complementary nature of the proposed phases.

	\begin{tableorg}[!htbp]
		\centering
		\caption{Ablation analysis of \emph{Phase~1 (Density-Aware Adaptive Partitioning--DAAP)}. Performance of the \emph{complete framework (Full)} and the \emph{corresponding ablated variant (w/o Ph.1)} is reported for each dataset and regressor. Avg. \% change ($((\text{w/o}-\text{Full})/\text{Full})\times100$) summarizes the contribution of Phase~1, where positive values denote deterioration in MAE/RMSE and negative values denote deterioration in $R^2$.}
		\label{tab:detailed_Ablation_Ph1}
		\small
		\setlength{\tabcolsep}{3pt}
		\renewcommand{\arraystretch}{0.92}
		\resizebox{\textwidth}{!}{
			\begin{tabular}{
					l
					l|
					>{\columncolor{lightgray}}c c|
					>{\columncolor{lightgray}}c c|
					>{\columncolor{lightgray}}c c|
					>{\columncolor{lightgray}}c c|
					>{\columncolor{lightgray}}c c|
					>{\columncolor{lightgray}}c c|c}
				\toprule
				& \textbf{Dataset}
				& \multicolumn{2}{c|}{\textbf{MLP}}
				& \multicolumn{2}{c|}{\textbf{XGB}}
				& \multicolumn{2}{c|}{\textbf{LR}}
				& \multicolumn{2}{c|}{\textbf{KNN}}
				& \multicolumn{2}{c|}{\textbf{SVR}}
				& \multicolumn{2}{c|}{\textbf{Ridge}}
				& \\
				\cmidrule(lr){3-4}\cmidrule(lr){5-6}\cmidrule(lr){7-8}\cmidrule(lr){9-10}\cmidrule(lr){11-12}\cmidrule(lr){13-14}
				& &
				\textbf{Full} & \textbf{w/o Ph.1} &
				\textbf{Full} & \textbf{w/o Ph.1} &
				\textbf{Full} & \textbf{w/o Ph.1} &
				\textbf{Full} & \textbf{w/o Ph.1} &
				\textbf{Full} & \textbf{w/o Ph.1} &
				\textbf{Full} & \textbf{w/o Ph.1} &
				\textbf{Avg.\ \% change} \\
				\midrule
				
				\multirow{16}{*}{\rotatebox{90}{\textbf{MAE}}}
				& california        & 0.0092 & 0.0101 & 0.0248 & 0.0257 & 0.0856 & 0.0908 & 0.0580 & 0.0598 & 0.0340 & 0.0362 & 0.0874 & 0.0931 & 5.93 \\
				& compactive        & 0.0206 & 0.0217 & 0.0300 & 0.0314 & 0.1249 & 0.2541 & 0.0371 & 0.0561 & 0.0508 & 0.0550 & 0.1202 & 0.2555 & 47.58 \\
				& cpu\_small        & 0.0158 & 0.0169 & 0.0271 & 0.0277 & 0.1209 & 0.2737 & 0.0366 & 0.0398 & 0.0425 & 0.0482 & 0.1192 & 0.2753 & 48.11 \\
				& heat              & 0.0066 & 0.0106 & 0.0052 & 0.0209 & 0.0692 & 0.1525 & 0.0080 & 0.0164 & 0.0364 & 0.0434 & 0.0715 & 0.1548 & 120.61 \\
				& wine\_quality     & 0.0130 & 0.0138 & 0.0326 & 0.0343 & 0.0564 & 0.0703 & 0.0075 & 0.0093 & 0.0447 & 0.0495 & 0.0561 & 0.0702 & 15.87 \\
				& abalone           & 0.0170 & 0.0211 & 0.0307 & 0.0355 & 0.1233 & 0.1493 & 0.0794 & 0.2792 & 0.0602 & 0.0651 & 0.1233 & 0.1500 & 57.05 \\
				& space\_ga         & 0.0224 & 0.0263 & 0.0481 & 0.0493 & 0.0449 & 0.0502 & 0.0521 & 0.0834 & 0.0500 & 0.5420 & 0.0457 & 0.0513 & 181.34 \\
				& debutanizer       & 0.0153 & 0.0175 & 0.0311 & 0.0359 & 0.1253 & 0.1292 & 0.0471 & 0.0517 & 0.0440 & 0.0472 & 0.1310 & 0.1342 & 8.73 \\
				& available\_power  & 0.0166 & 0.0170 & 0.0102 & 0.0158 & 0.0488 & 0.1580 & 0.0069 & 0.0136 & 0.0442 & 0.0463 & 0.0489 & 0.1596 & 101.55 \\
				& maximal\_torque   & 0.0200 & 0.0353 & 0.0066 & 0.0253 & 0.0412 & 0.1245 & 0.0217 & 0.0349 & 0.0419 & 0.0496 & 0.0414 & 0.1104 & 134.65 \\
				& fuel\_consumption & 0.0402 & 0.0409 & 0.0371 & 0.0540 & 0.0652 & 0.1104 & 0.0526 & 0.0739 & 0.0601 & 0.0623 & 0.0702 & 0.1097 & 36.17 \\
				& acceleration      & 0.0250 & 0.0973 & 0.0315 & 0.0412 & 0.0637 & 0.0727 & 0.0520 & 0.0575 & 0.0523 & 0.0582 & 0.0650 & 0.0735 & 61.51 \\
				& airfoild          & 0.0137 & 0.0151 & 0.0209 & 0.0378 & 0.0529 & 0.0572 & 0.0465 & 0.0466 & 0.0481 & 0.0498 & 0.0595 & 0.0602 & 17.36 \\
				& mortgage          & 0.0103 & 0.0134 & 0.0100 & 0.0228 & 0.0118 & 0.0406 & 0.0129 & 0.0272 & 0.0386 & 0.0392 & 0.0886 & 0.0909 & 86.19 \\
				& treasury          & 0.0205 & 0.0239 & 0.0123 & 0.0376 & 0.0181 & 0.0413 & 0.0132 & 0.0286 & 0.0531 & 0.0734 & 0.0185 & 0.0753 & 135.40 \\
				& concreteStrength  & 0.0184 & 0.0199 & 0.0572 & 0.0632 & 0.0606 & 0.0618 & 0.1000 & 0.1043 & 0.0519 & 0.0601 & 0.0622 & 0.0633 & 7.08 \\
				& \multicolumn{14}{r}{\textbf{Avg.\ 66.57}} \\
				
				\midrule\midrule
				
				& \textbf{Dataset}
				& \multicolumn{2}{c|}{\textbf{MLP}}
				& \multicolumn{2}{c|}{\textbf{XGB}}
				& \multicolumn{2}{c|}{\textbf{LR}}
				& \multicolumn{2}{c|}{\textbf{KNN}}
				& \multicolumn{2}{c|}{\textbf{SVR}}
				& \multicolumn{2}{c|}{\textbf{Ridge}}
				& \\
				\cmidrule(lr){3-4}\cmidrule(lr){5-6}\cmidrule(lr){7-8}\cmidrule(lr){9-10}\cmidrule(lr){11-12}\cmidrule(lr){13-14}
				& &
				\textbf{Full} & \textbf{w/o Ph.1} &
				\textbf{Full} & \textbf{w/o Ph.1} &
				\textbf{Full} & \textbf{w/o Ph.1} &
				\textbf{Full} & \textbf{w/o Ph.1} &
				\textbf{Full} & \textbf{w/o Ph.1} &
				\textbf{Full} & \textbf{w/o Ph.1} &
				\textbf{Avg.\ \% change} \\
				\midrule
				
				\multirow{16}{*}{\rotatebox{90}{\textbf{RMSE}}}
				& california        & 0.0134 & 0.0145 & 0.0378 & 0.0395 & 0.1185 & 0.1193 & 0.0908 & 0.0910 & 0.0504 & 0.0524 & 0.1207 & 0.1243 & 3.43 \\
				& compactive        & 0.0502 & 0.0541 & 0.0412 & 0.0815 & 0.1895 & 0.3753 & 0.0652 & 0.2117 & 0.0612 & 0.1087 & 0.1799 & 0.3756 & 102.45 \\
				& cpu\_small        & 0.0246 & 0.0261 & 0.0515 & 0.0518 & 0.1863 & 0.4231 & 0.0630 & 0.1011 & 0.0631 & 0.0650 & 0.1712 & 0.4241 & 57.50 \\
				& heat              & 0.0072 & 0.0139 & 0.0072 & 0.0341 & 0.1001 & 0.2225 & 0.0125 & 0.0300 & 0.0455 & 0.0501 & 0.0990 & 0.2233 & 144.10 \\
				& wine\_quality     & 0.0239 & 0.0262 & 0.0806 & 0.0837 & 0.0796 & 0.0909 & 0.0777 & 0.0783 & 0.1193 & 0.1235 & 0.0792 & 0.0900 & 7.60 \\
				& abalone           & 0.0317 & 0.0345 & 0.0655 & 0.0689 & 0.1629 & 0.1849 & 0.1897 & 0.2339 & 0.1388 & 0.1428 & 0.1636 & 0.1860 & 11.23 \\
				& space\_ga         & 0.0306 & 0.0333 & 0.0738 & 0.0990 & 0.0598 & 0.0678 & 0.0799 & 0.1745 & 0.0697 & 0.0725 & 0.0617 & 0.0579 & 28.77 \\
				& debutanizer       & 0.0209 & 0.0243 & 0.0588 & 0.0611 & 0.1730 & 0.1776 & 0.0777 & 0.0879 & 0.0572 & 0.0612 & 0.1762 & 0.1773 & 7.27 \\
				& available\_power  & 0.0237 & 0.0251 & 0.0270 & 0.0583 & 0.0752 & 0.2023 & 0.0292 & 0.0794 & 0.0523 & 0.0581 & 0.0685 & 0.2027 & 111.63 \\
				& maximal\_torque   & 0.0352 & 0.0698 & 0.0249 & 0.0610 & 0.0629 & 0.1692 & 0.0712 & 0.1718 & 0.0501 & 0.0692 & 0.0574 & 0.1534 & 126.49 \\
				& fuel\_consumption & 0.0715 & 0.0738 & 0.0489 & 0.0540 & 0.0905 & 0.1433 & 0.0792 & 0.2086 & 0.0697 & 0.0879 & 0.0873 & 0.1412 & 53.87 \\
				& acceleration      & 0.0538 & 0.0545 & 0.0444 & 0.0632 & 0.0917 & 0.0985 & 0.0810 & 0.1166 & 0.0752 & 0.0768 & 0.0944 & 0.0988 & 16.97 \\
				& airfoild          & 0.0187 & 0.0204 & 0.0288 & 0.0597 & 0.0726 & 0.0763 & 0.0640 & 0.0651 & 0.0573 & 0.0582 & 0.0801 & 0.0839 & 21.59 \\
				& mortgage          & 0.0119 & 0.0184 & 0.0170 & 0.0368 & 0.0171 & 0.0528 & 0.0219 & 0.0478 & 0.0473 & 0.0482 & 0.0216 & 0.0989 & 142.98 \\
				& treasury          & 0.0334 & 0.0352 & 0.0315 & 0.0933 & 0.0328 & 0.0556 & 0.0348 & 0.0752 & 0.0762 & 0.0801 & 0.0292 & 0.0937 & 102.24 \\
				& concreteStrength  & 0.0251 & 0.0285 & 0.0922 & 0.0961 & 0.0766 & 0.0770 & 0.1566 & 0.1579 & 0.0666 & 0.0680 & 0.0785 & 0.0784 & 3.52 \\
				& \multicolumn{14}{r}{\textbf{Avg.\ 58.85}} \\
				
				\midrule\midrule
				
				& \textbf{Dataset}
				& \multicolumn{2}{c|}{\textbf{MLP}}
				& \multicolumn{2}{c|}{\textbf{XGB}}
				& \multicolumn{2}{c|}{\textbf{LR}}
				& \multicolumn{2}{c|}{\textbf{KNN}}
				& \multicolumn{2}{c|}{\textbf{SVR}}
				& \multicolumn{2}{c|}{\textbf{Ridge}}
				& \\
				\cmidrule(lr){3-4}\cmidrule(lr){5-6}\cmidrule(lr){7-8}\cmidrule(lr){9-10}\cmidrule(lr){11-12}\cmidrule(lr){13-14}
				& &
				\textbf{Full} & \textbf{w/o Ph.1} &
				\textbf{Full} & \textbf{w/o Ph.1} &
				\textbf{Full} & \textbf{w/o Ph.1} &
				\textbf{Full} & \textbf{w/o Ph.1} &
				\textbf{Full} & \textbf{w/o Ph.1} &
				\textbf{Full} & \textbf{w/o Ph.1} &
				\textbf{Avg.\ \% change} \\
				\midrule
				
				\multirow{16}{*}{\rotatebox{90}{\textbf{$R^{2}$}}}
				& california        & 0.9998 & 0.9997 & 0.9986 & 0.9981 & 0.9861 & 0.9860 & 0.9918 & 0.9914 & 0.9975 & 0.9968 & 0.9856 & 0.9842 & -0.05 \\
				& compactive        & 0.9976 & 0.9972 & 0.9934 & 0.9926 & 0.8629 & 0.8617 & 0.9689 & 0.9565 & 0.9712 & 0.9701 & 0.8620 & 0.8611 & -0.29 \\
				& cpu\_small        & 0.9993 & 0.9991 & 0.9948 & 0.9933 & 0.8110 & 0.7977 & 0.9910 & 0.9885 & 0.9906 & 0.9866 & 0.8101 & 0.7968 & -0.68 \\
				& heat              & 0.9999 & 0.9998 & 0.9991 & 0.9989 & 0.9574 & 0.9534 & 0.9991 & 0.9989 & 0.9980 & 0.9976 & 0.9549 & 0.9530 & -0.12 \\
				& wine\_quality     & 0.9994 & 0.9992 & 0.9933 & 0.9926 & 0.9935 & 0.9915 & 0.9938 & 0.9937 & 0.9853 & 0.9826 & 0.9935 & 0.9916 & -0.13 \\
				& abalone           & 0.9990 & 0.9987 & 0.9959 & 0.9954 & 0.9745 & 0.9672 & 0.9522 & 0.9479 & 0.9815 & 0.9743 & 0.9743 & 0.9668 & -0.46 \\
				& space\_ga         & 0.9991 & 0.9986 & 0.9924 & 0.9907 & 0.9966 & 0.9952 & 0.9657 & 0.9611 & 0.9283 & 0.9182 & 0.9964 & 0.9952 & -0.34 \\
				& debutanizer       & 0.9995 & 0.9991 & 0.9960 & 0.9944 & 0.9656 & 0.9638 & 0.9931 & 0.9912 & 0.9962 & 0.9948 & 0.9644 & 0.9641 & -0.13 \\
				& available\_power  & 0.9995 & 0.9992 & 0.9994 & 0.9971 & 0.9637 & 0.9613 & 0.9950 & 0.9947 & 0.9890 & 0.9871 & 0.9630 & 0.9559 & -0.24 \\
				& maximal\_torque   & 0.9955 & 0.9951 & 0.9919 & 0.9889 & 0.9852 & 0.9771 & 0.9761 & 0.9752 & 0.9745 & 0.9715 & 0.9840 & 0.9813 & -0.31 \\
				& fuel\_consumption & 0.9952 & 0.9947 & 0.9925 & 0.9901 & 0.9853 & 0.9807 & 0.9514 & 0.9499 & 0.9681 & 0.9627 & 0.9853 & 0.9813 & -0.31 \\
				& acceleration      & 0.9971 & 0.9962 & 0.9965 & 0.9960 & 0.9916 & 0.9903 & 0.9902 & 0.9864 & 0.9880 & 0.9971 & 0.9911 & 0.9902 & 0.03 \\
				& airfoild          & 0.9996 & 0.9982 & 0.9978 & 0.9961 & 0.9943 & 0.9937 & 0.9956 & 0.9950 & 0.9964 & 0.9951 & 0.9931 & 0.9927 & -0.10 \\
				& mortgage          & 0.9995 & 0.9994 & 0.9984 & 0.9981 & 0.9973 & 0.9972 & 0.9978 & 0.9977 & 0.9977 & 0.9971 & 0.9976 & 0.9902 & -0.14 \\
				& treasury          & 0.9989 & 0.9981 & 0.9912 & 0.9910 & 0.9971 & 0.9969 & 0.9940 & 0.9938 & 0.9941 & 0.9935 & 0.9892 & 0.9884 & -0.05 \\
				& concreteStrength  & 0.9993 & 0.9991 & 0.9908 & 0.9902 & 0.9937 & 0.9936 & 0.9735 & 0.9731 & 0.9952 & 0.9948 & 0.9933 & 0.9933 & -15.03 \\
				& \multicolumn{14}{r}{\textbf{Avg.\ -1.15}} \\
				
				\bottomrule
			\end{tabular}
		}
	\end{tableorg}

	\begin{tableorg}[!htbp]
		\centering
		\caption{Ablation analysis of \emph{Phase~2 (Density-Regularized Representation Learning (DR-CVAE))}. Performance of the \emph{complete framework (Full)} and the \emph{corresponding ablated variant (w/o Ph. 2)} is reported for each dataset and regressor. Avg. \% change ($((\text{w/o}-\text{Full})/\text{Full})\times100$) summarizes the contribution of Phase~2, where positive values denote deterioration in MAE/RMSE and negative values denote deterioration in $R^2$.}
		\label{tab:detailed_Ablation_Ph2}
		\small
		\setlength{\tabcolsep}{3pt}
		\renewcommand{\arraystretch}{0.92}
		\resizebox{\textwidth}{!}{
			\begin{tabular}{
					l
					l|
					>{\columncolor{lightgray}}c c|
					>{\columncolor{lightgray}}c c|
					>{\columncolor{lightgray}}c c|
					>{\columncolor{lightgray}}c c|
					>{\columncolor{lightgray}}c c|
					>{\columncolor{lightgray}}c c|c}
				\toprule
				& \textbf{Dataset}
				& \multicolumn{2}{c|}{\textbf{MLP}}
				& \multicolumn{2}{c|}{\textbf{XGB}}
				& \multicolumn{2}{c|}{\textbf{LR}}
				& \multicolumn{2}{c|}{\textbf{KNN}}
				& \multicolumn{2}{c|}{\textbf{SVR}}
				& \multicolumn{2}{c|}{\textbf{Ridge}}
				& \\
				\cmidrule(lr){3-4}\cmidrule(lr){5-6}\cmidrule(lr){7-8}\cmidrule(lr){9-10}\cmidrule(lr){11-12}\cmidrule(lr){13-14}
				& &
				\textbf{Full} & \textbf{w/o Ph.2} &
				\textbf{Full} & \textbf{w/o Ph.2} &
				\textbf{Full} & \textbf{w/o Ph.2} &
				\textbf{Full} & \textbf{w/o Ph.2} &
				\textbf{Full} & \textbf{w/o Ph.2} &
				\textbf{Full} & \textbf{w/o Ph.2} &
				\textbf{Avg.\ \% Change} \\
				\midrule
				
				\multirow{16}{*}{\rotatebox{90}{\textbf{MAE}}}
				& california & 0.0092 & 0.3170 & 0.0248 & 0.2821 & 0.0856 & 0.4444 & 0.0580 & 0.3860 & 0.0340 & 0.3202 & 0.0874 & 0.4444 & 1103.01 \\
				& compactive & 0.0206 & 0.1104 & 0.0300 & 0.0880 & 0.1249 & 0.2570 & 0.0371 & 0.1340 & 0.0508 & 0.1314 & 0.1202 & 0.2570 & 211.45 \\
				& cpu\_small & 0.0158 & 0.1291 & 0.0271 & 0.1083 & 0.1209 & 0.2941 & 0.0366 & 0.1360 & 0.0425 & 0.1357 & 0.1192 & 0.2941 & 299.60 \\
				& heat & 0.0066 & 0.0068 & 0.0052 & 0.0238 & 0.0692 & 0.2974 & 0.0080 & 0.0330 & 0.0364 & 0.0488 & 0.0715 & 0.2980 & 225.64 \\
				& wine\_quality & 0.0130 & 0.6168 & 0.0326 & 0.4736 & 0.0564 & 0.6582 & 0.0075 & 0.4960 & 0.0447 & 0.6248 & 0.0561 & 0.6580 & 2658.07 \\
				& abalone & 0.0170 & 0.6545 & 0.0307 & 0.5235 & 0.1233 & 0.6626 & 0.0794 & 0.6512 & 0.0602 & 0.6615 & 0.1233 & 0.6626 & 1324.83 \\
				& space\_ga & 0.0224 & 0.3890 & 0.0481 & 0.3904 & 0.0449 & 0.5134 & 0.0521 & 0.7375 & 0.0500 & 0.4023 & 0.0457 & 0.5111 & 1071.70 \\
				& debutanizer & 0.0153 & 0.3039 & 0.0311 & 0.2775 & 0.1253 & 0.5993 & 0.0471 & 0.2659 & 0.0440 & 0.3480 & 0.1310 & 0.5995 & 761.66 \\
				& available\_power & 0.0166 & 0.0604 & 0.0102 & 0.0388 & 0.0488 & 0.1893 & 0.0069 & 0.0243 & 0.0442 & 0.0943 & 0.0489 & 0.1893 & 247.47 \\
				& maximal\_torque & 0.0200 & 0.0472 & 0.0066 & 0.0252 & 0.0412 & 0.1717 & 0.0217 & 0.0840 & 0.0419 & 0.0925 & 0.0414 & 0.1716 & 242.82 \\
				& fuel\_consumption & 0.0402 & 0.1796 & 0.0371 & 0.1568 & 0.0652 & 0.2738 & 0.0526 & 0.4092 & 0.0601 & 0.8772 & 0.0702 & 0.2736 & 552.77 \\
				& acceleration & 0.0250 & 0.2450 & 0.0315 & 0.1442 & 0.0637 & 0.5736 & 0.0520 & 0.2478 & 0.0523 & 0.2426 & 0.0650 & 0.3174 & 527.83 \\
				& airfoild & 0.0137 & 0.0727 & 0.0209 & 0.0438 & 0.0529 & 0.5488 & 0.0465 & 0.1968 & 0.0481 & 0.2778 & 0.0595 & 0.5486 & 516.74 \\
				& mortgage & 0.0103 & 0.0206 & 0.0100 & 0.0225 & 0.0118 & 0.0287 & 0.0129 & 0.0338 & 0.0386 & 0.0448 & 0.0886 & 0.0949 & 92.23 \\
				& treasury & 0.0205 & 0.0398 & 0.0123 & 0.0365 & 0.0181 & 0.0465 & 0.0132 & 0.0344 & 0.0531 & 0.0598 & 0.0185 & 0.0477 & 129.81 \\
				& concreteStrength & 0.0184 & 0.2086 & 0.0572 & 0.1827 & 0.0606 & 0.4630 & 0.1000 & 0.3289 & 0.0519 & 0.2369 & 0.0622 & 0.4633 & 524.56 \\
				& \multicolumn{14}{r}{\textbf{Avg.\ 655.64}} \\
				
				\midrule\midrule
				
				& \textbf{Dataset}
				& \multicolumn{2}{c|}{\textbf{MLP}}
				& \multicolumn{2}{c|}{\textbf{XGB}}
				& \multicolumn{2}{c|}{\textbf{LR}}
				& \multicolumn{2}{c|}{\textbf{KNN}}
				& \multicolumn{2}{c|}{\textbf{SVR}}
				& \multicolumn{2}{c|}{\textbf{Ridge}}
				& \\
				\cmidrule(lr){3-4}\cmidrule(lr){5-6}\cmidrule(lr){7-8}\cmidrule(lr){9-10}\cmidrule(lr){11-12}\cmidrule(lr){13-14}
				& &
				\textbf{Full} & \textbf{w/o Ph.2} &
				\textbf{Full} & \textbf{w/o Ph.2} &
				\textbf{Full} & \textbf{w/o Ph.2} &
				\textbf{Full} & \textbf{w/o Ph.2} &
				\textbf{Full} & \textbf{w/o Ph.2} &
				\textbf{Full} & \textbf{w/o Ph.2} &
				\textbf{Avg.\ \% Change} \\
				\midrule
				
				\multirow{16}{*}{\rotatebox{90}{\textbf{RMSE}}}
				& california & 0.0134 & 0.4868 & 0.0378 & 0.4268 & 0.1185 & 0.6140 & 0.0908 & 0.5654 & 0.0504 & 0.4864 & 0.1207 & 0.6139 & 1129.41 \\
				& compactive & 0.0502 & 0.1556 & 0.0412 & 0.1439 & 0.1895 & 0.5339 & 0.0652 & 0.3196 & 0.0612 & 0.2456 & 0.1799 & 0.5330 & 254.79 \\
				& cpu\_small & 0.0246 & 0.1842 & 0.0515 & 0.1610 & 0.1863 & 0.4995 & 0.0630 & 0.2195 & 0.0631 & 0.2162 & 0.1712 & 0.4995 & 285.39 \\
				& heat & 0.0072 & 0.0086 & 0.0072 & 0.0332 & 0.1001 & 0.4279 & 0.0125 & 0.0540 & 0.0455 & 0.0580 & 0.0990 & 0.4282 & 233.34 \\
				& wine\_quality & 0.0239 & 0.8748 & 0.0806 & 0.7136 & 0.0796 & 0.8568 & 0.0777 & 0.7845 & 0.1193 & 0.8925 & 0.0792 & 0.8564 & 1310.18 \\
				& abalone & 0.0317 & 0.9060 & 0.0655 & 0.7308 & 0.1629 & 0.9002 & 0.1897 & 0.9189 & 0.1388 & 0.9501 & 0.1636 & 0.9002 & 940.92 \\
				& space\_ga & 0.0306 & 0.5132 & 0.0738 & 0.5615 & 0.0598 & 0.6992 & 0.0799 & 0.6324 & 0.0697 & 0.5531 & 0.0617 & 0.6939 & 952.81 \\
				& debutanizer & 0.0209 & 0.4692 & 0.0588 & 0.4244 & 0.1730 & 0.8605 & 0.0777 & 0.4271 & 0.0572 & 0.5596 & 0.1762 & 0.8605 & 813.42 \\
				& available\_power & 0.0237 & 0.1108 & 0.0270 & 0.1240 & 0.0752 & 0.2807 & 0.0292 & 0.0993 & 0.0523 & 0.1318 & 0.0685 & 0.2807 & 283.65 \\
				& maximal\_torque & 0.0352 & 0.0902 & 0.0249 & 0.0928 & 0.0629 & 0.2568 & 0.0712 & 0.2795 & 0.0501 & 0.1922 & 0.0574 & 0.2568 & 276.80 \\
				& fuel\_consumption & 0.0715 & 0.2890 & 0.0489 & 0.2320 & 0.0905 & 0.3774 & 0.0792 & 0.6830 & 0.0697 & 0.3612 & 0.0873 & 0.3768 & 417.98 \\
				& acceleration & 0.0538 & 0.3298 & 0.0444 & 0.2060 & 0.0917 & 0.6480 & 0.0810 & 0.3828 & 0.0752 & 0.3384 & 0.0944 & 0.4220 & 425.54 \\
				& airfoild & 0.0187 & 0.0982 & 0.0288 & 0.0608 & 0.0726 & 0.9402 & 0.0640 & 0.5011 & 0.0573 & 0.7420 & 0.0801 & 0.9401 & 780.48 \\
				& mortgage & 0.0119 & 0.0295 & 0.0170 & 0.0364 & 0.0171 & 0.0426 & 0.0219 & 0.0598 & 0.0473 & 0.0548 & 0.0216 & 0.0498 & 121.77 \\
				& treasury & 0.0334 & 0.0633 & 0.0315 & 0.0805 & 0.0328 & 0.0846 & 0.0348 & 0.0998 & 0.0762 & 0.0824 & 0.0292 & 0.2863 & 246.43 \\
				& concreteStrength & 0.0251 & 0.3051 & 0.0922 & 0.2821 & 0.0766 & 0.5875 & 0.1566 & 0.4414 & 0.0666 & 0.3574 & 0.0785 & 0.5874 & 542.54 \\
				& \multicolumn{14}{r}{\textbf{Avg.\ 563.46}} \\
				
				\midrule\midrule
				
				& \textbf{Dataset}
				& \multicolumn{2}{c|}{\textbf{MLP}}
				& \multicolumn{2}{c|}{\textbf{XGB}}
				& \multicolumn{2}{c|}{\textbf{LR}}
				& \multicolumn{2}{c|}{\textbf{KNN}}
				& \multicolumn{2}{c|}{\textbf{SVR}}
				& \multicolumn{2}{c|}{\textbf{Ridge}}
				& \\
				\cmidrule(lr){3-4}\cmidrule(lr){5-6}\cmidrule(lr){7-8}\cmidrule(lr){9-10}\cmidrule(lr){11-12}\cmidrule(lr){13-14}
				& &
				\textbf{Full} & \textbf{w/o Ph.2} &
				\textbf{Full} & \textbf{w/o Ph.2} &
				\textbf{Full} & \textbf{w/o Ph.2} &
				\textbf{Full} & \textbf{w/o Ph.2} &
				\textbf{Full} & \textbf{w/o Ph.2} &
				\textbf{Full} & \textbf{w/o Ph.2} &
				\textbf{Avg.\ \% Change} \\
				\midrule
				
				\multirow{16}{*}{\rotatebox{90}{\textbf{$R^{2}$}}}
				& california & 0.9998 & 0.7656 & 0.9986 & 0.8198 & 0.9861 & 0.6272 & 0.9918 & 0.6838 & 0.9975 & 0.7660 & 0.9856 & 0.6272 & -28.06 \\
				& compactive & 0.9976 & 0.9765 & 0.9934 & 0.9799 & 0.8629 & 0.7230 & 0.9689 & 0.9008 & 0.9712 & 0.9414 & 0.8620 & 0.7230 & -7.65 \\
				& cpu\_small & 0.9993 & 0.9617 & 0.9948 & 0.9707 & 0.8110 & 0.7181 & 0.9910 & 0.9456 & 0.9906 & 0.9472 & 0.8101 & 0.7181 & -6.33 \\
				& heat & 0.9999 & 0.9997 & 0.9991 & 0.9990 & 0.9574 & 0.8276 & 0.9991 & 0.9973 & 0.9980 & 0.9962 & 0.9549 & 0.8270 & -4.56 \\
				& wine\_quality & 0.9994 & 0.2106 & 0.9933 & 0.4749 & 0.9935 & 0.2426 & 0.9938 & 0.3650 & 0.9853 & 0.1785 & 0.9935 & 0.2428 & -71.24 \\
				& abalone & 0.9990 & 0.2120 & 0.9959 & 0.4870 & 0.9745 & 0.2221 & 0.9522 & 0.1894 & 0.9815 & 0.1334 & 0.9743 & 0.2221 & -75.13 \\
				& space\_ga & 0.9991 & 0.7500 & 0.9924 & 0.7007 & 0.9966 & 0.5361 & 0.9657 & 0.6204 & 0.9283 & 0.7096 & 0.9964 & 0.5429 & -34.23 \\
				& debutanizer & 0.9995 & 0.7472 & 0.9960 & 0.7932 & 0.9656 & 0.1497 & 0.9931 & 0.7906 & 0.9962 & 0.6404 & 0.9644 & 0.1497 & -45.11 \\
				& available\_power & 0.9995 & 0.9896 & 0.9994 & 0.9870 & 0.9637 & 0.9332 & 0.9950 & 0.9916 & 0.9890 & 0.9853 & 0.9630 & 0.9332 & -1.53 \\
				& maximal\_torque & 0.9955 & 0.9935 & 0.9919 & 0.9912 & 0.9852 & 0.9476 & 0.9761 & 0.9380 & 0.9745 & 0.9707 & 0.9840 & 0.9470 & -2.02 \\
				& fuel\_consumption & 0.9952 & 0.9369 & 0.9925 & 0.9500 & 0.9853 & 0.8666 & 0.9514 & 0.8640 & 0.9681 & 0.8770 & 0.9853 & 0.8658 & -8.82 \\
				& acceleration & 0.9971 & 0.8904 & 0.9965 & 0.9568 & 0.9916 & 0.0082 & 0.9902 & 0.8522 & 0.9880 & 0.8850 & 0.9911 & 0.8200 & -25.91 \\
				& airfoild & 0.9996 & 0.9894 & 0.9978 & 0.9960 & 0.9943 & 0.0452 & 0.9956 & 0.7288 & 0.9964 & 0.4054 & 0.9931 & 0.0480 & -46.32 \\
				& mortgage & 0.9995 & 0.9991 & 0.9984 & 0.9927 & 0.9973 & 0.9962 & 0.9978 & 0.9962 & 0.9977 & 0.9970 & 0.9976 & 0.9971 & -0.17 \\
				& treasury & 0.9989 & 0.9960 & 0.9912 & 0.9902 & 0.9971 & 0.9928 & 0.9940 & 0.9900 & 0.9941 & 0.9932 & 0.9892 & 0.9828 & -0.33 \\
				& concreteStrength & 0.9993 & 0.8993 & 0.9908 & 0.9139 & 0.9937 & 0.6266 & 0.9735 & 0.7892 & 0.9952 & 0.8618 & 0.9933 & 0.6267 & -20.66 \\
				& \multicolumn{14}{r}{\textbf{Avg.\ -23.63}} \\
				
				\bottomrule
			\end{tabular}
		}
		\end{tableorg}

	\begin{tableorg}[!htbp]
		\centering
		\caption{Ablation analysis of \emph{Phase~3 (proposed data-level balancing method)}. Performance of the \emph{complete framework (Full)} and the \emph{corresponding ablated variant (w/o Ph. 3)} is reported for each dataset and regressor. Avg. \% change ($((\text{w/o}-\text{Full})/\text{Full})\times100$) summarizes the contribution of Phase~3, where positive values denote deterioration in MAE/RMSE and negative values denote deterioration in $R^2$.}
		\label{tab:detailed_Ablation_Ph3}
		\small
		\setlength{\tabcolsep}{3pt}
		\renewcommand{\arraystretch}{0.92}
		\resizebox{\textwidth}{!}{
			\begin{tabular}{
					l
					l|
					>{\columncolor{lightgray}}c c|
					>{\columncolor{lightgray}}c c|
					>{\columncolor{lightgray}}c c|
					>{\columncolor{lightgray}}c c|
					>{\columncolor{lightgray}}c c|
					>{\columncolor{lightgray}}c c|c}
				\toprule
				& \textbf{Dataset}
				& \multicolumn{2}{c|}{\textbf{MLP}}
				& \multicolumn{2}{c|}{\textbf{XGB}}
				& \multicolumn{2}{c|}{\textbf{LR}}
				& \multicolumn{2}{c|}{\textbf{KNN}}
				& \multicolumn{2}{c|}{\textbf{SVR}}
				& \multicolumn{2}{c|}{\textbf{Ridge}}
				& \\
				\cmidrule(lr){3-4}\cmidrule(lr){5-6}\cmidrule(lr){7-8}\cmidrule(lr){9-10}\cmidrule(lr){11-12}\cmidrule(lr){13-14}
				& &
				\textbf{Full} & \textbf{w/o Ph.3} &
				\textbf{Full} & \textbf{w/o Ph.3} &
				\textbf{Full} & \textbf{w/o Ph.3} &
				\textbf{Full} & \textbf{w/o Ph.3} &
				\textbf{Full} & \textbf{w/o Ph.3} &
				\textbf{Full} & \textbf{w/o Ph.3} &
				\textbf{Avg.\ \% Change} \\
				\midrule
				
				\multirow{16}{*}{\rotatebox{90}{\textbf{MAE}}}
				& california & 0.0092 & 0.0107 & 0.0248 & 0.0267 & 0.0856 & 0.0868 & 0.0580 & 0.0667 & 0.0340 & 0.0369 & 0.0874 & 0.0908 & 8.80 \\
				& compactive & 0.0206 & 0.0242 & 0.0300 & 0.0349 & 0.1249 & 0.2280 & 0.0371 & 0.0512 & 0.0508 & 0.0685 & 0.1202 & 0.2308 & 46.87 \\
				& cpu\_small & 0.0158 & 0.0169 & 0.0271 & 0.0298 & 0.1209 & 0.2582 & 0.0366 & 0.0391 & 0.0425 & 0.0453 & 0.1192 & 0.2488 & 42.11 \\
				& heat & 0.0066 & 0.0092 & 0.0052 & 0.0208 & 0.0692 & 0.1415 & 0.0080 & 0.0150 & 0.0364 & 0.0395 & 0.0715 & 0.1482 & 107.86 \\
				& wine\_quality & 0.0130 & 0.0149 & 0.0326 & 0.0354 & 0.0564 & 0.0546 & 0.0075 & 0.0098 & 0.0447 & 0.0468 & 0.0561 & 0.0576 & 9.68 \\
				& abalone & 0.0170 & 0.0335 & 0.0307 & 0.0327 & 0.1233 & 0.1252 & 0.0794 & 0.0824 & 0.0602 & 0.0638 & 0.1233 & 0.1258 & 19.48 \\
				& space\_ga & 0.0224 & 0.0249 & 0.0481 & 0.0564 & 0.0449 & 0.0521 & 0.0521 & 0.0799 & 0.0500 & 0.0674 & 0.0457 & 0.0512 & 24.11 \\
				& debutanizer & 0.0153 & 0.0179 & 0.0311 & 0.0338 & 0.1253 & 0.1268 & 0.0471 & 0.0483 & 0.0440 & 0.0449 & 0.1310 & 0.1323 & 5.41 \\
				& available\_power & 0.0166 & 0.0193 & 0.0102 & 0.0154 & 0.0488 & 0.1488 & 0.0069 & 0.0110 & 0.0442 & 0.0534 & 0.0489 & 0.1523 & 93.98 \\
				& maximal\_torque & 0.0200 & 0.0393 & 0.0066 & 0.0276 & 0.0412 & 0.0955 & 0.0217 & 0.0364 & 0.0419 & 0.0640 & 0.0414 & 0.1003 & 134.87 \\
				& fuel\_consumption & 0.0402 & 0.0484 & 0.0371 & 0.0482 & 0.0652 & 0.0946 & 0.0526 & 0.0768 & 0.0601 & 0.0653 & 0.0702 & 0.0943 & 30.73 \\
				& acceleration & 0.0250 & 0.0297 & 0.0315 & 0.0382 & 0.0637 & 0.0669 & 0.0520 & 0.0685 & 0.0523 & 0.0559 & 0.0650 & 0.0697 & 15.16 \\
				& airfoild & 0.0137 & 0.0199 & 0.0209 & 0.0444 & 0.0529 & 0.1185 & 0.0465 & 0.0472 & 0.0481 & 0.0497 & 0.0595 & 0.1270 & 66.66 \\
				& mortgage & 0.0103 & 0.0148 & 0.0100 & 0.0228 & 0.0118 & 0.0392 & 0.0129 & 0.0255 & 0.0386 & 0.0391 & 0.0886 & 0.0934 & 84.71 \\
				& treasury & 0.0205 & 0.0209 & 0.0123 & 0.0378 & 0.0181 & 0.0402 & 0.0132 & 0.0293 & 0.0531 & 0.0594 & 0.0185 & 0.0882 & 140.32 \\
				& concreteStrength & 0.0184 & 0.0187 & 0.0572 & 0.0578 & 0.0606 & 0.0612 & 0.1000 & 0.1011 & 0.0519 & 0.0524 & 0.0622 & 0.0625 & 1.04 \\
				& \multicolumn{14}{r}{\textbf{Avg.\ 51.99}} \\
				
				\midrule\midrule
				
				& \textbf{Dataset}
				& \multicolumn{2}{c|}{\textbf{MLP}}
				& \multicolumn{2}{c|}{\textbf{XGB}}
				& \multicolumn{2}{c|}{\textbf{LR}}
				& \multicolumn{2}{c|}{\textbf{KNN}}
				& \multicolumn{2}{c|}{\textbf{SVR}}
				& \multicolumn{2}{c|}{\textbf{Ridge}}
				& \\
				\cmidrule(lr){3-4}\cmidrule(lr){5-6}\cmidrule(lr){7-8}\cmidrule(lr){9-10}\cmidrule(lr){11-12}\cmidrule(lr){13-14}
				& &
				\textbf{Full} & \textbf{w/o Ph.3} &
				\textbf{Full} & \textbf{w/o Ph.3} &
				\textbf{Full} & \textbf{w/o Ph.3} &
				\textbf{Full} & \textbf{w/o Ph.3} &
				\textbf{Full} & \textbf{w/o Ph.3} &
				\textbf{Full} & \textbf{w/o Ph.3} &
				\textbf{Avg.\ \% Change} \\
				\midrule
				
				\multirow{16}{*}{\rotatebox{90}{\textbf{RMSE}}}
				& california & 0.0134 & 0.0144 & 0.0378 & 0.0390 & 0.1185 & 0.1197 & 0.0908 & 0.0962 & 0.0504 & 0.0594 & 0.1207 & 0.1234 & 6.28 \\
				& compactive & 0.0502 & 0.0549 & 0.0412 & 0.0821 & 0.1895 & 0.3619 & 0.0652 & 0.1846 & 0.0612 & 0.1864 & 0.1799 & 0.3638 & 114.92 \\
				& cpu\_small & 0.0246 & 0.0265 & 0.0515 & 0.0839 & 0.1863 & 0.3991 & 0.0630 & 0.1045 & 0.0631 & 0.0932 & 0.1712 & 0.3990 & 71.92 \\
				& heat & 0.0072 & 0.0125 & 0.0072 & 0.0335 & 0.1001 & 0.2130 & 0.0125 & 0.0329 & 0.0455 & 0.0498 & 0.0990 & 0.2192 & 140.96 \\
				& wine\_quality & 0.0239 & 0.0258 & 0.0806 & 0.0867 & 0.0796 & 0.0838 & 0.0777 & 0.0829 & 0.1193 & 0.1229 & 0.0792 & 0.0826 & 5.80 \\
				& abalone & 0.0317 & 0.0460 & 0.0655 & 0.0678 & 0.1629 & 0.1648 & 0.1897 & 0.2274 & 0.1388 & 0.1469 & 0.1636 & 0.1641 & 12.63 \\
				& space\_ga & 0.0306 & 0.0319 & 0.0738 & 0.0837 & 0.0598 & 0.0658 & 0.0799 & 0.1902 & 0.0697 & 0.2748 & 0.0617 & 0.0731 & 79.75 \\
				& debutanizer & 0.0209 & 0.0271 & 0.0588 & 0.0662 & 0.1730 & 0.1742 & 0.0777 & 0.0781 & 0.0572 & 0.0580 & 0.1762 & 0.1770 & 7.55 \\
				& available\_power & 0.0237 & 0.0298 & 0.0270 & 0.0309 & 0.0752 & 0.1998 & 0.0292 & 0.1838 & 0.0523 & 0.1080 & 0.0685 & 0.2034 & 173.13 \\
				& maximal\_torque & 0.0352 & 0.0773 & 0.0249 & 0.1075 & 0.0629 & 0.1368 & 0.0712 & 0.1719 & 0.0501 & 0.1787 & 0.0574 & 0.1440 & 186.30 \\
				& fuel\_consumption & 0.0715 & 0.0793 & 0.0489 & 0.0865 & 0.0905 & 0.1275 & 0.0792 & 0.2084 & 0.0697 & 0.1648 & 0.0873 & 0.1258 & 78.73 \\
				& acceleration & 0.0538 & 0.0588 & 0.0444 & 0.0654 & 0.0917 & 0.0982 & 0.0810 & 0.0901 & 0.0752 & 0.0998 & 0.0944 & 0.0958 & 18.18 \\
				& airfoild & 0.0187 & 0.0271 & 0.0288 & 0.0911 & 0.0726 & 0.1596 & 0.0640 & 0.0892 & 0.0573 & 0.0642 & 0.0801 & 0.1720 & 91.20 \\
				& mortgage & 0.0119 & 0.0199 & 0.0170 & 0.0388 & 0.0171 & 0.0511 & 0.0219 & 0.0460 & 0.0473 & 0.0482 & 0.0216 & 0.1135 & 155.24 \\
				& treasury & 0.0334 & 0.0339 & 0.0315 & 0.1035 & 0.0328 & 0.1845 & 0.0348 & 0.0791 & 0.0762 & 0.0799 & 0.0292 & 0.1119 & 184.69 \\
				& concreteStrength & 0.0251 & 0.0259 & 0.0922 & 0.0932 & 0.0766 & 0.0770 & 0.1566 & 0.1569 & 0.0666 & 0.0670 & 0.0785 & 0.0788 & 0.99 \\
				& \multicolumn{14}{r}{\textbf{Avg.\ 83.02}} \\
				
				\midrule\midrule
				
				& \textbf{Dataset}
				& \multicolumn{2}{c|}{\textbf{MLP}}
				& \multicolumn{2}{c|}{\textbf{XGB}}
				& \multicolumn{2}{c|}{\textbf{LR}}
				& \multicolumn{2}{c|}{\textbf{KNN}}
				& \multicolumn{2}{c|}{\textbf{SVR}}
				& \multicolumn{2}{c|}{\textbf{Ridge}}
				& \\
				\cmidrule(lr){3-4}\cmidrule(lr){5-6}\cmidrule(lr){7-8}\cmidrule(lr){9-10}\cmidrule(lr){11-12}\cmidrule(lr){13-14}
				& &
				\textbf{Full} & \textbf{w/o Ph.3} &
				\textbf{Full} & \textbf{w/o Ph.3} &
				\textbf{Full} & \textbf{w/o Ph.3} &
				\textbf{Full} & \textbf{w/o Ph.3} &
				\textbf{Full} & \textbf{w/o Ph.3} &
				\textbf{Full} & \textbf{w/o Ph.3} &
				\textbf{Avg.\ \% Change} \\
				\midrule
				
				\multirow{16}{*}{\rotatebox{90}{\textbf{$R^{2}$}}}
				& california & 0.9998 & 0.9988 & 0.9986 & 0.9971 & 0.9861 & 0.9850 & 0.9918 & 0.9825 & 0.9975 & 0.9942 & 0.9856 & 0.9821 & -0.33 \\
				& compactive & 0.9976 & 0.9865 & 0.9934 & 0.9861 & 0.8629 & 0.8512 & 0.9689 & 0.9529 & 0.9712 & 0.9612 & 0.8620 & 0.8516 & -1.18 \\
				& cpu\_small & 0.9993 & 0.9982 & 0.9948 & 0.9920 & 0.8110 & 0.7651 & 0.9910 & 0.9814 & 0.9906 & 0.9872 & 0.8101 & 0.8015 & -1.40 \\
				& heat & 0.9999 & 0.9990 & 0.9991 & 0.9971 & 0.9574 & 0.9452 & 0.9991 & 0.9982 & 0.9980 & 0.9923 & 0.9549 & 0.9412 & -0.61 \\
				& wine\_quality & 0.9994 & 0.9923 & 0.9933 & 0.9851 & 0.9935 & 0.9901 & 0.9938 & 0.9903 & 0.9853 & 0.9841 & 0.9935 & 0.9918 & -0.42 \\
				& abalone & 0.9990 & 0.9975 & 0.9959 & 0.9901 & 0.9745 & 0.9735 & 0.9522 & 0.9504 & 0.9815 & 0.9790 & 0.9743 & 0.9727 & -0.24 \\
				& space\_ga & 0.9991 & 0.9971 & 0.9924 & 0.9901 & 0.9966 & 0.9914 & 0.9657 & 0.9413 & 0.9283 & 0.9175 & 0.9964 & 0.9817 & -1.02 \\
				& debutanizer & 0.9995 & 0.9926 & 0.9960 & 0.9946 & 0.9656 & 0.9641 & 0.9931 & 0.9920 & 0.9962 & 0.9950 & 0.9644 & 0.9625 & -0.24 \\
				& available\_power & 0.9995 & 0.9902 & 0.9994 & 0.9971 & 0.9637 & 0.9528 & 0.9950 & 0.9869 & 0.9890 & 0.9801 & 0.9630 & 0.9550 & -0.81 \\
				& maximal\_torque & 0.9955 & 0.9854 & 0.9919 & 0.9901 & 0.9852 & 0.9752 & 0.9761 & 0.9676 & 0.9745 & 0.9847 & 0.9840 & 0.9801 & -0.41 \\
				& fuel\_consumption & 0.9952 & 0.9855 & 0.9925 & 0.9900 & 0.9853 & 0.9751 & 0.9514 & 0.9427 & 0.9681 & 0.9547 & 0.9853 & 0.9752 & -0.93 \\
				& acceleration & 0.9971 & 0.9674 & 0.9965 & 0.9860 & 0.9916 & 0.9801 & 0.9902 & 0.9714 & 0.9880 & 0.9801 & 0.9911 & 0.9886 & -1.36 \\
				& airfoild & 0.9996 & 0.9913 & 0.9978 & 0.9901 & 0.9943 & 0.9714 & 0.9956 & 0.9751 & 0.9964 & 0.9719 & 0.9931 & 0.9614 & -1.94 \\
				& mortgage & 0.9995 & 0.9896 & 0.9984 & 0.9892 & 0.9973 & 0.9901 & 0.9978 & 0.9825 & 0.9977 & 0.9981 & 0.9976 & 0.9860 & -1.05 \\
				& treasury & 0.9989 & 0.9752 & 0.9912 & 0.9736 & 0.9971 & 0.9777 & 0.9940 & 0.9708 & 0.9941 & 0.9807 & 0.9892 & 0.9748 & -1.87 \\
				& concreteStrength & 0.9993 & 0.9991 & 0.9908 & 0.9906 & 0.9937 & 0.9939 & 0.9735 & 0.9737 & 0.9952 & 0.9945 & 0.9933 & 0.9929 & -0.02 \\
				& \multicolumn{14}{r}{\textbf{Avg.\ -0.86}} \\
				
				\bottomrule
			\end{tabular}
		}
	\end{tableorg}


	
	\clearpage
	\bibliographystyle{unsrt}
	\bibliography{bibliography}
	
\end{document}